\documentclass{article}

\usepackage[T1]{fontenc}
\usepackage{iclr2027_conference,times}

\usepackage{amsmath,amsfonts,bm}

\def\eqref#1{equation~\ref{#1}}

\def\1{\bm{1}}

\DeclareMathAlphabet{\mathsfit}{\encodingdefault}{\sfdefault}{m}{sl}
\SetMathAlphabet{\mathsfit}{bold}{\encodingdefault}{\sfdefault}{bx}{n}

\usepackage{amsmath,amssymb,bm}
\usepackage{booktabs}
\usepackage{multirow}
\usepackage{array,longtable}
\usepackage{enumitem}
\usepackage{graphicx}
\usepackage{microtype}
\usepackage{url}
\usepackage{xcolor}
\usepackage{colortbl}
\usepackage{tabularx}
\definecolor{oursbg}{gray}{0.9}
\newcommand{\ours}{\cellcolor{oursbg}}
\newcolumntype{R}{>{\raggedleft\arraybackslash}X}
\newcolumntype{C}{>{\centering\arraybackslash}X}
\usepackage{listings}
\usepackage{hyperref}
\hypersetup{hidelinks}

\usepackage{titlesec}
\titlespacing*{\paragraph}{0pt}{6pt}{1em}

\title{Dependency-Aware Reward Shaping\\for Agentic Reinforcement Learning}

\author{%
Ziyi Chen\textsuperscript{1,*}, Yan Zhang\textsuperscript{2,*},
Jianhui Wei\textsuperscript{3,*}, Daoan Zhang\textsuperscript{4},
Zuozhu Liu\textsuperscript{3}\\[4pt]
\normalfont\small\textsuperscript{1}University of Illinois Urbana-Champaign\\
\normalfont\small\textsuperscript{2}National University of Singapore\quad
\textsuperscript{3}Zhejiang University\\
\normalfont\small\textsuperscript{4}University of Rochester\\[3pt]
\normalfont\small\textsuperscript{*}Equal contribution.
}

\iclrfinalcopy
\hypersetup{
  pdftitle={Dependency-Aware Reward Shaping for Agentic Reinforcement Learning},
  pdfauthor={Ziyi Chen, Yan Zhang, Jianhui Wei, Daoan Zhang, Zuozhu Liu}
}

\begin{document}

\maketitle
\lhead{Preprint}

\begin{abstract}
When training large language models with reinforcement learning, terminal
rewards provide little guidance about which steps matter. Common methods for
assigning step credit overlook that work built on uncorrected mistakes is
wasted while independent work remains valid. With only a final
success/failure reward, every step in a failed episode has zero total
future reward, even when it made progress. We propose Dependency-Aware Reward
Shaping (DARS), which represents task progress as predicates linked by
prerequisite relations and assigns step-level credit over the dependency graph. An annotator marks which predicates each step verifies, invalidates, or repairs. Verified
predicates are discounted according to graph distance from the nearest
broken prerequisite, while independent predicates are unaffected. Repairs
update these weights based on any errors that remain; invalidated
predicates need re-verification to regain credit. A fixed potential converts these annotations into
signed per-step rewards. A common reward and
annotation interface allows DARS to integrate with a range of reasoning
and agentic training methods, such as GiGPO and ARPO/AEPO, without changing
their rollout strategies or optimizers. Across five task families and
models from 1.5B to 8B, DARS improves success by up to $10$ points over
GiGPO trained with the same budget and harness (ALFWorld), raises the
WebShop task score and Search-R1 QA accuracy, complements AEPO's
entropy-based training on AIME24/25 with a Python interpreter, and exceeds
OmniOPD in controlled tool-free reasoning comparisons at 1.7B and 4B. Ablations show that step-level credit,
dependency attenuation, and graph topology each contribute. On ALFWorld,
a distilled 8B annotator matches the API annotator, enabling DARS to run
efficiently without a frontier judge.
Code is available at \url{https://github.com/JianhuiWei7/DARS}.
\end{abstract}

\section{Introduction}
\label{sec:intro}

\begin{figure*}[t]
\centering
  \includegraphics[width=1\textwidth]{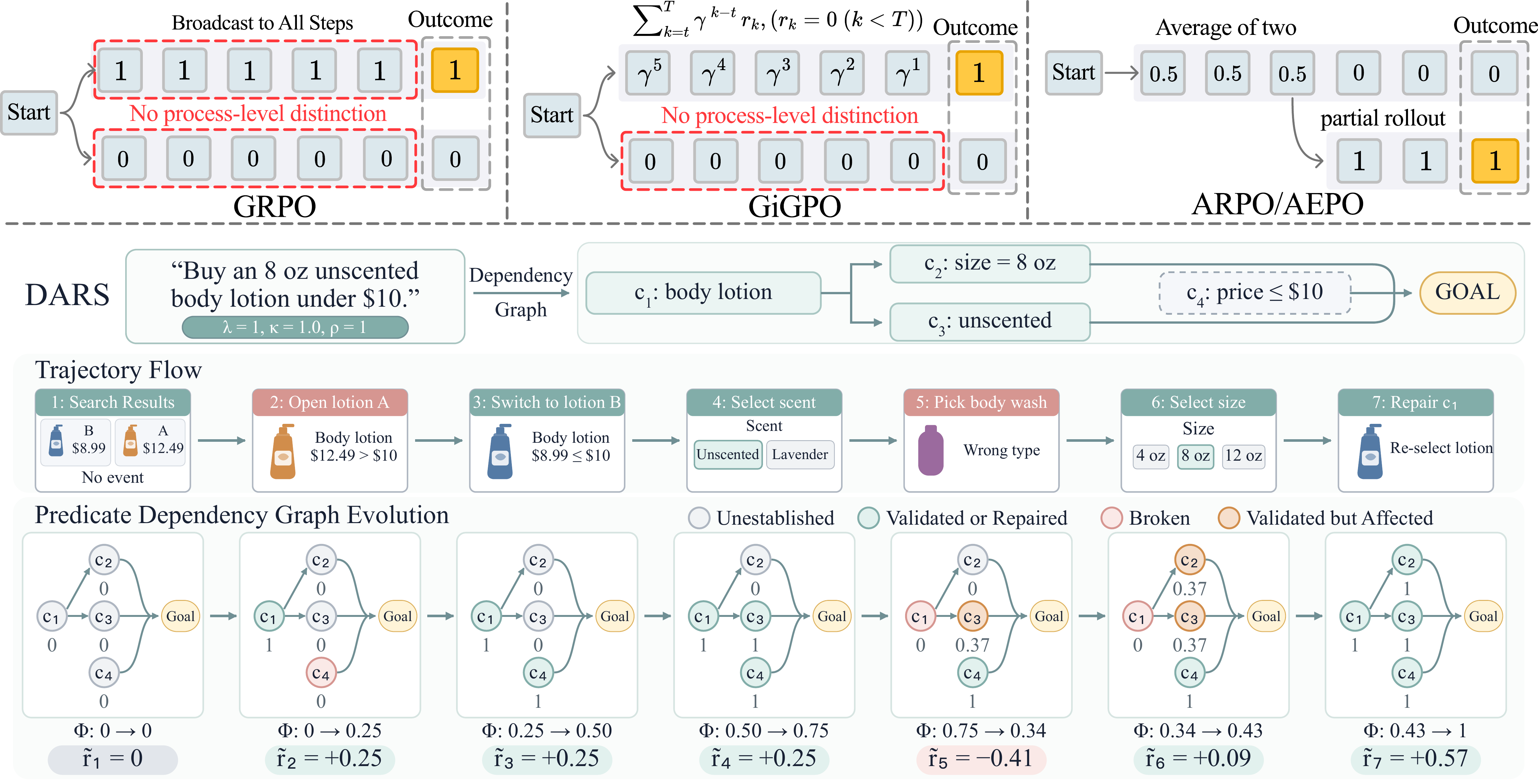}
    \caption{
    \textbf{Top:} Comparison of existing methods. GRPO shares an episode-level signal across all steps; GiGPO uses time-discounted future returns; and ARPO/AEPO allocates additional rollouts around uncertain positions. \textbf{Bottom:} A schematic DARS credit assignment example. Product type $c_1$ is a prerequisite for size $c_2$ and scent $c_3$, while price $c_4$ is independent. Picking body wash (Step 5) breaks $c_1$, attenuating the weights of $c_2,c_3$ while preserving $c_4$. Re-selecting lotion (Step 7) repairs and re-verifies $c_1$, restoring full weight to the retained local evidence for $c_2,c_3$. Step rewards $\tilde r_t$ are signed changes in the dependency-weighted potential $\Phi$, with $\lambda=\kappa=\rho=1$.
    }
  \label{fig:dars-motivation}

\end{figure*}

Language models increasingly perform tasks requiring sequences of
interdependent decisions: planning household tasks
\citep{shridhar2021alfworld}, navigating shopping websites
\citep{yao2022webshop}, and retrieving evidence across documents
\citep{jin2025searchr1,yao2023react}. Reinforcement learning with
verifiable outcome rewards is the standard way to train these agents
\citep{lambert2024tulu3,shao2024deepseekmath,deepseekai2025r1}, usually
using group-based objectives that compare rollouts without a learned value
function. Yet outcome rewards say little about each step's
contributions: useful intermediate decisions and irrelevant detours receive
the same credit, which makes policy optimization difficult
over long trajectories
\citep{minsky1961steps,sutton1988td,zhang2025agenticsurvey}.

Recent methods refine credit using uncertainty or temporal position. ARPO
\citep{dong2025arpo} and AEPO \citep{dong2026aepo} use token entropy
after tool calls to allocate additional rollouts
\citep{wang2025beyond}. This directs sampling toward uncertain
decisions, but uncertainty does not necessarily indicate importance:
alternative phrasings and different orders of independent subgoals can
also produce high entropy
\citep{wang2025hicra,li2026notalltokens}. GiGPO
\citep{feng2025gigpo} instead assigns finer credit by comparing actions
taken from recurring states using discounted future returns. However,
its discount is determined by the number of remaining turns rather than
by which subgoals a step establishes or which later decisions depend on
it \citep{shen2025carl,cheng2026graphcredit}. Moreover, under
outcome-only rewards, every step in a failed trajectory has zero return,
leaving useful progress indistinguishable from mistakes within the same
trajectory (Figure~\ref{fig:dars-motivation}, top).

To address these limitations, we propose
\textbf{Dependency-Aware Reward Shaping (DARS)}. We represent the
conditions for a successful task as a set of
predicates connected by prerequisite relations. Together, these
predicates and relations form the task's dependency graph. DARS assigns
step-level credit according to changes in valid progress over this
graph. Let $\pi^\star$ denote the target policy. When an error invalidates a predicate, credit assigned to its dependent predicates is attenuated,
while predicates on independent branches retain their credit.
Subsequent steps that rely on the invalidated predicate form the
\emph{dependent suffix}, which is off-distribution relative to
$\pi^\star$ until the error is repaired.
Figure~\ref{fig:dars-motivation} (bottom) illustrates this boundary:
picking body wash invalidates the product-type predicate,
attenuating credit for the dependent size and scent predicates while
preserving credit for the independent price predicate. Re-selecting
lotion repairs and re-verifies the prerequisite, restoring full weight
to the retained size and scent evidence. This perspective connects dependency-aware credit assignment to the
covariate shift underlying compounding errors in imitation learning
\citep{ross2011dagger} and autoregressive generation
\citep{bachmann2024pitfalls}, as well as to the propagation of root-cause
errors through agent trajectories
\citep{zhu2025agentfail,liang2026elpo}.

\begin{figure*}[t]
\centering
  \includegraphics[width=1\textwidth]{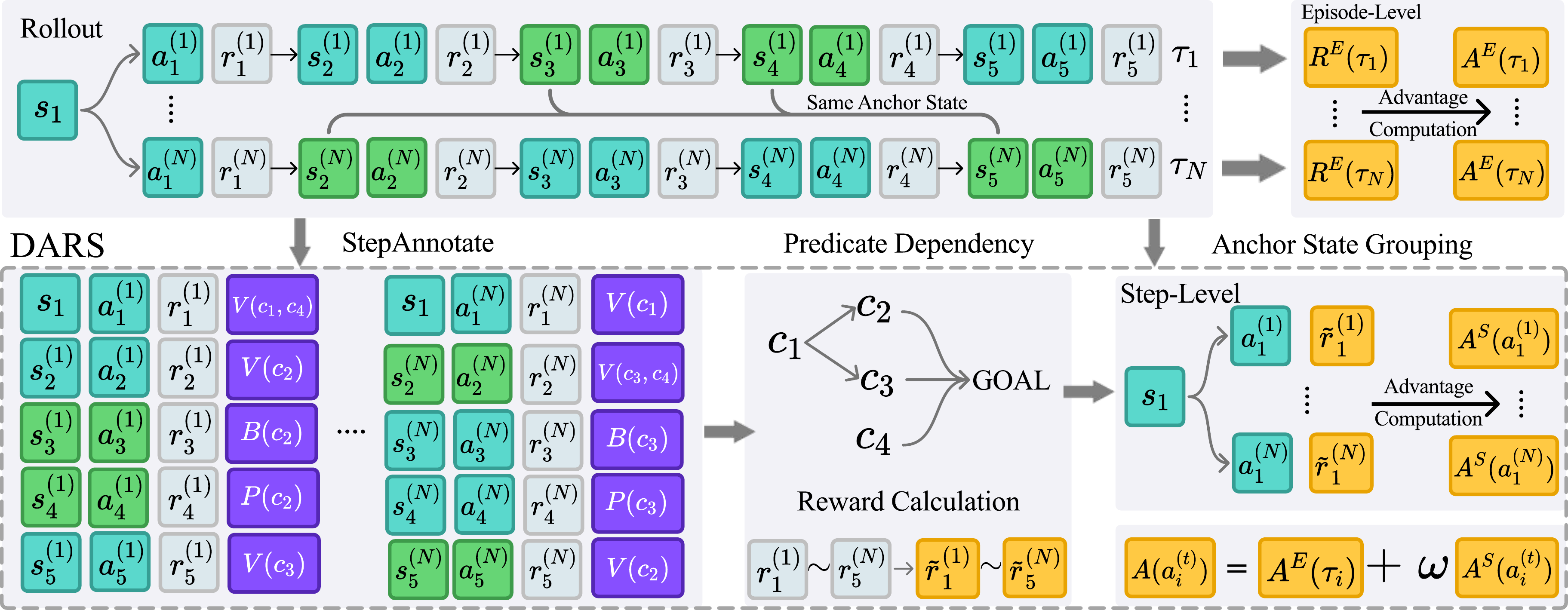}
  \caption{DARS on the GiGPO backbone. The learner samples $N$ rollouts
and retains its episode-level advantage $A^E$. An annotator identifies
task predicates and their per-step verification (V), invalidation (B),
and repair (P) events. A deterministic graph potential converts these
events into signed step rewards $\tilde r_t$, which replace the environment
rewards in the anchor-state channel producing $A^S$.}
  \label{fig:main_pipe}

\end{figure*}

DARS operationalizes this principle through a deterministic potential
over the dependency graph (Figure~\ref{fig:main_pipe}). An annotator
reads each completed trajectory once and reports which task predicates
each step verifies, invalidates, or repairs. The potential weights each
supported predicate according to its graph distance from the nearest
broken ancestor, and the signed change in this potential defines the
step reward. Consequently, credit is attenuated along dependent
branches, preserved on independent branches, and restored after repair.
Because DARS modifies only the reward signal, it can be integrated into
existing RL algorithms without changing their rollout strategies or
optimizers.

We evaluate one reward and annotation interface across five task
families. DARS improves ALFWorld and WebShop over published GiGPO
references, increases Search-R1 accuracy on shared-pool and held-out
splits, raises AEPO's best evaluated AIME24/25 score by $4.2$ points,
and exceeds OmniOPD at both tool-free reasoning scales
(Sections~\ref{sec:exp-main} and~\ref{sec:exp-reasoning}).

Our contributions are:
\begin{itemize}[leftmargin=*,nosep]
\item \textbf{Dependency-bounded invalidation (hypothesis).} Under
the task's predicate-state abstraction, what an unrepaired error
invalidates is bounded by the dependency structure rather than by step
position, so only the dependent suffix is off-distribution for the
target policy (Section~\ref{sec:intro});
consistent with this reading, removing the attenuation, serializing the
graph into a chain, or collapsing credit to a scalar each reduces
performance (Section~\ref{sec:exp-ablation}).
\item \textbf{Dependency-Aware Reward Shaping (DARS).} A fine-grained
step reward from a fixed dependency-graph potential over annotated task
predicates, signed and auditable, whose single decay parameter spans
milestone counting to hard masking below a break, and which integrates
with GiGPO, ARPO/AEPO, and a token-level interface without changing
their rollouts or optimizers (Section~\ref{sec:method}).
\item \textbf{Unified evidence on agentic and non-agentic tasks.} A
single reward and annotation interface improves on or matches strong
recent methods on household planning, web shopping, multi-hop retrieval,
and mathematical reasoning with and without tools, and learns faster
than its closest baseline.
\end{itemize}

\section{Related Work}
\label{sec:related}

\paragraph{Reinforcement learning for LLM agents.}
Group-relative RL with verifiable rewards is widely used to train
language-model agents
\citep{shao2024deepseekmath,lambert2024tulu3,yu2025dapo,zhang2025agenticsurvey}.
GiGPO \citep{feng2025gigpo} adds a step-level advantage over states
recurring across rollouts, a signal that vanishes when all anchor groups
are singletons. ARPO \citep{dong2025arpo} and AEPO \citep{dong2026aepo}
allocate rollout samples using token entropy after tool calls
\citep{wang2025beyond}; AEPO also modifies clipping and advantages using
entropy. Tree-structured rollouts share prefixes for sibling-contrastive
credit \citep{ji2026treegrpo}. DARS provides a reward signal that composes
with these mechanisms; we evaluate it atop GiGPO and AEPO.

\paragraph{Process rewards and LLM judges.}
Process reward models evaluate intermediate reasoning steps
\citep{uesato2022process,lightman2023letsverify,wang2024mathshepherd,setlur2024pavs,yuan2024implicitprm,cui2025prime}.
ProcessBench evaluates localization of the first incorrect step
\citep{zheng2024processbench}. LLMs also evaluate agent behavior
\citep{zhuge2024agentjudge} and attribute failures to individual steps
\citep{zhang2025failureattribution}, with known sensitivity to evaluator
bias and error \citep{zheng2023judging,gu2024judgesurvey}. DARS requests
only structural events (verification, invalidation, repair) from the
annotator and computes rewards deterministically from them, which keeps
the rule auditable and allows offline re-crediting of the same
annotations under different rules (Section~\ref{sec:exp-ablation}).
On-policy distillation \citep{agarwal2024gkd,gu2024minillm,omniopd2026}
and hidden-state intrinsic rewards \citep{starpo2026} supply process
signals without an environment; we compare with both in the
single-response setting.

\paragraph{Credit beyond temporal position.}
Classical RL studies contribution through return decomposition
\citep{arjonamedina2019rudder}, hindsight conditioning
\citep{harutyunyan2019hca}, and counterfactual analysis
\citep{meulemans2023cocoa}. Potential-based shaping preserves optimal
policies under appropriate conditions
\citep{ng1999policy,wiewiora2003principled,devlin2012dynamic}.
Procedural readings of proof structure originate in logic programming
\citep{kowalski1979algorithm,kazemi2023lambada}. Agentic credit methods
include stepwise progress attribution \citep{wang2025sparl}, implicit
step rewards \citep{liu2025istar}, hierarchical value decomposition
\citep{zhou2024archer,peng2026hiper}, error-localized optimization
\citep{liang2026elpo}, and graph-based attribution over visited states
\citep{cheng2026graphcredit}. DARS instead builds a task-predicate graph,
computing credit as a fixed function of dependency-weighted progress,
distinguishing dependent work, independent branches, and post-error
recovery.

\section{Preliminaries}
\label{sec:prelim}

\paragraph{Problem setup.}
A policy $\pi_\theta$ acts on a task $x\sim p(X)$ over segments
$t=1,\dots,T$. At segment $t$, it conditions on a state $s_t\in\mathcal{S}$,
emits $a_t\in\mathcal{V}^{\le n}$, and receives a reward
$r^{\mathrm{env}}_t\in\mathbb{R}$ and a next state $s_{t+1}$. In an
environment, each segment is a turn. Otherwise, a fixed segmentation
operator $\sigma$ splits the generation $y$ into
$(a_1,\dots,a_T)=\sigma(y)$ and $s_t=(x,a_{<t})$. An episode is a trajectory
$\bm{\tau}_i=\{(s^{(i)}_t,a^{(i)}_t,r^{(i)}_t)\}_{t=1}^{T_i}$ of length
$T_i$. Rewards are verifiable but sparse:
under a strict outcome reward, $r^{\mathrm{env}}_t$ is zero except at a
successful terminal segment (up to a small invalid-action penalty in the
interactive learners).
Failed episodes therefore have essentially all-zero reward sequences (the WebShop continuation runs of
Appendix~\ref{app:webshop-cont} instead use a graded terminal reward at
1.5B).

A segment is the unit for assessing what an episode has established:
an action whose effect the environment records or a reasoning span that
may fix an intermediate result. Segments that change nothing receive no
immediate shaping reward. Credit stays at this granularity: the
objective below gives every token of $a_t$ the same advantage.

\paragraph{Group-based RL.}
GRPO-style methods sample $N$ trajectories per task under
$\pi_{\theta_{\mathrm{old}}}$ and form advantages by within-group
comparison rather than against a learned value \citep{shao2024deepseekmath}.
Normalizing each trajectory's return
$R^{E}(\bm{\tau}_i)=\sum_{t=1}^{T_i} r^{(i)}_t$ across the group yields
an episode-level advantage $A^{E}(\bm{\tau}_i)$ shared by every segment.
GiGPO \citep{feng2025gigpo} adds a finer signal by comparing actions
from the same state. Let
$\mathcal{U}=\{\bar s_1,\dots,\bar s_U\}$ be the group's distinct states.
Each anchor state collects
\begin{equation}
G^{S}(\bar s)=\bigl\{(a^{(i)}_t,R^{S,(i)}_t)\;\big|\;s^{(i)}_t=\bar s\bigr\},
\qquad
R^{S,(i)}_t=\sum\nolimits_{k=t}^{T_i}\gamma^{\,k-t}\,r^{(i)}_k ,
\label{eq:anchor-group}
\end{equation}
with $\gamma\in(0,1]$. Normalizing within each anchor group gives
$A^{S}(a^{(i)}_t)$. The combined advantage
\begin{equation}
A(a^{(i)}_t)=A^{E}(\bm{\tau}_i)+\omega\,A^{S}(a^{(i)}_t),
\qquad \omega\ge 0,
\label{eq:gigpo-adv}
\end{equation}
enters the standard token-level clipped objective. Two properties matter
below. First, under terminal-only reward,
$R^{S,(i)}_t=\gamma^{\,T_i-t}\,r^{(i)}_{T_i}$ on success and $0$ on
failure. A failed episode's return does not distinguish useful from
incorrect actions. Its normalized advantage can still be nonzero in
an anchor group containing successful continuations. Second, anchor
comparisons require recurring states. If none recurs, every
$G^{S}(\bar s)$ is a singleton, $A^{S}\equiv 0$, and
Eq.~(\ref{eq:gigpo-adv}) reduces to $A=A^{E}$.

\paragraph{Two channels.}
We call the per-segment sequence $\{r_t\}$ entering
Eq.~(\ref{eq:anchor-group}) the \textbf{step channel} and
$R^{E}(\bm{\tau}_i)$ the \textbf{episode channel}. Unmodified, both carry
$r_t=r^{\mathrm{env}}_t$, so the step channel holds nothing before the
terminal segment. Section~\ref{sec:method} uses the task's dependency
structure to write a dense reward $\tilde r_t$ into the step channel.
The episode channel, anchor grouping, and optimizer are retained.
Learners without anchor comparisons use the token-level interface in
Section~\ref{sec:method-integration}.

\section{DARS}

\label{sec:method}

DARS assigns time-step credit from changes in dependency-weighted progress. It represents
a solution as a sequence of time steps, tracks the persistent state of each task
predicate, and attenuates verified progress that depends on unresolved errors.

\subsection{Solution Traces and Dependency Graphs}
\label{sec:method-proof}

For a task $x$, let $\bm\tau=(u_1,\ldots,u_T)$ be a solution trace of $T$ time
steps. DARS represents its task requirements with a dependency graph
$G_x=(\mathcal C_x,\mathcal E_x)$, where
$\mathcal C_x=\{c_1,\ldots,c_M\}$ is a set of predicates and
$c_j\rightarrow c_i$ means that $c_i$ presupposes $c_j$. Establishing a prerequisite
does not establish its child, which requires its own evidence. We use domain-specific
rules to instantiate ordered object chains, parallel requirement graphs, and hop or
derivation chains (Appendix~\ref{app:protocol}).

At each time step $t$, the annotator reports for each affected predicate $c_i$ a set
of events
\begin{equation}
  E_i(t)\subseteq\{V,B,P\},
  \label{eq:event}
\end{equation}
where $V$, $B$, and $P$ denote verification, invalidation, and repair.
Verification establishes $c_i$, invalidation marks it as broken, and repair lifts a
previous break. A repair does not by itself re-establish the predicate: a step that
corrects the error and re-establishes $c_i$ reports both $P$ and $V$. If the time step
does not affect $c_i$, $E_i(t)=\varnothing$ and its state remains unchanged.

\subsection{From Graph State to Step Credit}
\label{sec:method-reward}

\paragraph{Graph state.}
After each time step $t$, every predicate $c_i\in\mathcal C_x$ maintains a persistent
state
\begin{equation}
  S_i(t)\in\{0,1\},
  \label{eq:predicate-state}
\end{equation}
where $S_i(t)=1$ means that $c_i$ is verified and $S_i(t)=0$ means that $c_i$ is
broken or unestablished. A broken predicate has been invalidated and has not yet been
repaired; $\mathcal U(t)$ contains the indices of broken predicates. An unestablished
predicate has not yet received sufficient evidence to be verified. A verified predicate
has been established by the trajectory and remains valid after time step $t$. All
predicates are initially unestablished:
\begin{equation}
  S_i(0)=0,\quad i=1,\ldots,M,\qquad \mathcal U(0)=\varnothing.
  \label{eq:predicate-init}
\end{equation}

The annotation events $E_i(t)$ determine how the persistent state changes at time step
$t$. Invalidation adds $c_i$ to the broken set and repair removes it,
$\mathcal U(t)=\bigl(\mathcal U(t-1)\cup\{i: B\in E_i(t)\}\bigr)\setminus\{i: P\in E_i(t)\}$;
verification sets an unbroken predicate to the verified state, and the previous state
persists otherwise:
\begin{equation}
  S_i(t)=
  \begin{cases}
    0, & i\in\mathcal U(t),\\
    1, & i\notin\mathcal U(t)\ \text{and}\ V\in E_i(t),\\
    S_i(t-1), & \text{otherwise}.
  \end{cases}
  \label{eq:state-transition}
\end{equation}
Repair and verification are therefore separate: a repair lifts the break, and the
predicate regains support only through a verification in the same or a later step. Because the state is persistent, repeating an already verified predicate or taking a
time step unrelated to it does not change the graph state.

Local verification and prerequisite validity are distinct. Breaking an ancestor attenuates
a descendant's credit without automatically deleting its local evidence. If that
evidence is itself incorrect, the descendant must also be marked broken and require
its own repair. Repairing a prerequisite therefore restores full weight only to
descendants whose local evidence remains valid. Predicate definitions must likewise distinguish
legitimate progression from undoing established progress.

\paragraph{Dependency weighting.}

Let $d_G(c_j,c_i)$ denote the shortest directed-path distance from
$c_j$ to $c_i$. For each predicate $c_i$, let
$\mathcal B_i(t)=\{j\in\mathcal U(t)\mid c_j\text{ is a strict ancestor
of }c_i\}$ denote the set of its unresolved broken ancestors. The distance from $c_i$ to its nearest unresolved broken ancestor is
\begin{equation}
  d_i(t)
  =
  \begin{cases}
    \displaystyle
    \min_{j\in\mathcal B_i(t)}
    d_G(c_j,c_i),
      & \mathcal B_i(t)\neq\varnothing,\\
    \infty,
      & \mathcal B_i(t)=\varnothing.
  \end{cases}
  \label{eq:broken-distance}
\end{equation}
Its dependency weight is
\begin{equation}
  w_i(t)=
  \begin{cases}
    e^{-\lambda d_i(t)}, & d_i(t)<\infty,\\
    1, & d_i(t)=\infty,
  \end{cases}
  \qquad \lambda\geq0.
  \label{eq:weight}
\end{equation}
A predicate with no broken ancestor retains full weight. Otherwise, it receives the
attenuation determined by its nearest broken ancestor, while unrelated branches remain
unaffected. Because the nearest broken ancestor determines the weight, introducing a
second break closer to $c_i$ can increase its weight by replacing a more distant source
of invalidity. If the graph's maximum path length is $D_x$, every verified predicate
has weight at least $e^{-\lambda D_x}$.
Attenuation therefore depends on dependency depth, not the number of generated steps.
At $\lambda=0$, only broken predicates themselves lose credit.

\paragraph{Signed progress.}
The potential sums the contributions of verified predicates; unestablished and broken
predicates contribute zero:
\begin{align}
  \Phi_G(t)
  &=
  \frac{1}{M}
  \sum_{i=1}^{M} S_i(t)w_i(t),
  \qquad \Phi_G(0)=0,
  \label{eq:phi}\\
  \tilde r_t
  &=
  \rho\,\operatorname{clip}\!\left(
    \Phi_G(t)-\Phi_G(t-1),
    -\kappa,\kappa
  \right),
  \label{eq:reward}
\end{align}
where $\rho>0$ is the reward-scale coefficient and $\kappa\in(0,1]$. Time steps
receive positive credit for net progress and negative credit for net regression, while
an unchanged graph state receives zero credit. Because $\Phi_G(t)\in[0,1]$, setting
$\kappa=1$ leaves the potential
difference unchanged. The deployed configuration uses $\kappa=0.3$ except in the WebShop~1.5B
$\kappa=1$ configurations (Appendix~\ref{app:protocol}).

For example, consider $c_1\rightarrow c_3\leftarrow c_2$ and
$c_3\rightarrow c_4$. If all four predicates are verified, breaking $c_1$ changes
$\Phi_G$ from $1$ to $(1+e^{-\lambda}+e^{-2\lambda})/4$.
The independent prerequisite $c_2$ keeps its credit, but repeating it earns nothing.
A repair of $c_1$ removes its attenuation of $c_3$ and $c_4$; if the same step also
re-verifies $c_1$, the original potential is recovered, provided the remaining local
evidence is still valid (with $P$ alone the potential is $3/4$).

When $\kappa=1$, the undiscounted reward sum telescopes to
$\rho\Phi_G(T)$, so returning to the same potential cannot increase the total shaping
return. Clipping and learner-specific transformations, including discounting and group
normalization, need not preserve this identity. We therefore do not claim policy
invariance for the deployed signal.

\subsection{Annotation and Learning Interface}
\label{sec:method-integration}

Given the domain topology, one LLM call reads each completed trace together with the
task and its available evidence, instantiates the required nodes, and returns the
per-step verification, invalidation, and repair events. The edge rule completes the
instance graph, which is then held fixed throughout state replay. State updates and
numerical credit are deterministic conditional on this graph and its event
annotations. Structural checks validate formatting, node references, and step
alignment, not semantic correctness. Graph construction, segmentation, and evidence
requirements are summarized in Appendix~\ref{app:prompts}. When annotation uses blocks
containing multiple semantic steps, the potential is evaluated at block boundaries;
this coarsens credit placement without replacing the task DAG by a sequence graph.

DARS does not prescribe a rollout strategy or optimization loss. A learner with separate
outcome and step channels can use $\tilde r_t$ in the latter while retaining its original
outcome signal. For a direct token interface, let $t(\ell)$ identify the scored step or
block containing generated token $\ell$. Its credit is broadcast without division by
span length:
\begin{equation}
  A_\ell=A_\ell^{\mathrm{global}}+
    \eta\,\frac{\tilde r_{t(\ell)}}{s_x},
  \label{eq:cot-adv}
\end{equation}
where $s_x>0$ is the configured within-group scale and $\eta\geq0$ controls local
credit. The local term is not mean-centered, so zero remains neutral.
Using $A_\ell^{\mathrm{global}}=z_x(\Phi_G(T))$, with group-relative normalization
$z_x$, setting $\eta=0$ gives the annotation-matched trajectory-level control on fixed
traces. Training settings and reward configurations are recorded in
Appendix~\ref{app:protocol}, and annotation handling is summarized in
Appendix~\ref{app:prompts}. Optional commitment and completion-check terms are separate
experimental variants, not part of Eq.~\eqref{eq:reward}.

\section{Experiments}
\label{sec:exp}

\subsection{Setup}
\label{sec:exp-setup}
\paragraph{Tasks and learners.}
We evaluate DARS on five task families. ALFWorld \citep{shridhar2021alfworld} (Qwen2.5-1.5B/7B-Instruct;
unseen-game success), WebShop \citep{yao2022webshop} (Qwen2.5-1.5B/7B;
strict success and graded task score), and Search-R1
\citep{jin2025searchr1} (Qwen2.5-7B; exact-match QA accuracy) use GiGPO
with DARS in the step channel. Mathematics with a Python interpreter
(Qwen3-8B; mean AIME24/25) uses the ARPO/AEPO recipe. Tool-free reasoning
on DAPO-Math \citep{yu2025dapo} (Qwen3-1.7B/4B chat; average of
AMC, AIME24, and AIME25) uses the token interface of
Eq.~(\ref{eq:cot-adv}). Domain topologies and annotation prompts are in
Appendices~\ref{app:protocol} and~\ref{app:prompts}.

\paragraph{Baselines.}
ALFWorld and WebShop use published GRPO, RLOO, and GiGPO references
\citep{feng2025gigpo} and our local reproductions
(Appendix~\ref{app:local-baselines}). Search-R1 uses our GiGPO
reproduction for the chosen retriever and task pool.
Agentic mathematics uses ARPO and AEPO as outcome-only arms of the
$2\times2$ design. Tool-free baselines comprise vanilla on-policy
distillation, OmniOPD \citep{omniopd2026}, StaRPO \citep{starpo2026},
scalar process-feedback baselines, and a control collapsing DARS's
annotations into one trajectory-level scalar.

\paragraph{Evaluation protocol.}
ALFWorld and WebShop report performance on unseen tasks; Search-R1
includes a shared-pool diagnostic and a held-out split. WebShop's
headline policies continue from trained GiGPO policies and are compared
with their parents on fresh decoding seeds. Full protocols are in
Appendix~\ref{app:protocol}; the WebShop continuation protocol is in
Appendix~\ref{app:webshop-cont}.

\begin{table}[t]
\centering
\caption{
\textbf{Interactive agent tasks} (Qwen2.5-Instruct; scores in \%).
$^{\dagger}$Published result and uncertainty \citep{feng2025gigpo}.
Unmarked rows are our runs (DARS script, hyper-parameters, and harness;
mean $\pm$ SE over decoding draws); ours are shaded. $\Delta$: gain of
DARS over the matched GiGPO row in points (paired, before rounding; tests
in Appendix~\ref{app:local-baselines}). $^{\ddagger}$DARS continues its
GiGPO parent. \mbox{Search-R1}: ``shared'' reuses the training pool and
``held-out'' is disjoint from it; 662 GiGPO vs.\ 360 DARS steps (shared),
one epoch each (held-out); single-hop is NQ, TriviaQA, PopQA, multi-hop the
other four; DARS leads in all three decoding groups for every $\Delta$.
Protocols: Appendix~\ref{app:protocol}.
}
\label{tab:main-gigpo}
\vspace{3pt}
\small
\setlength{\tabcolsep}{3pt}
\renewcommand{\arraystretch}{0.95}
\begin{minipage}[t]{0.56\textwidth}
\centering
\textbf{(a)} ALFWorld: unseen-game success\\[2pt]
\begin{tabularx}{\linewidth}{lCC}
\toprule
Method & 1.5B & 7B \\
\midrule
GRPO & $72.8 \pm 3.6$\rlap{$^{\dagger}$} & $87.4 \pm 0.4$ \\
RLOO & $69.7 \pm 2.5$\rlap{$^{\dagger}$} & $85.9 \pm 0.4$ \\
GiGPO (w/ std)$^{\dagger}$ & $86.7 \pm 1.7$ & $90.8 \pm 1.3$ \\
GiGPO (w/o std)$^{\dagger}$ & $86.1 \pm 4.7$ & $90.2 \pm 2.3$ \\
GiGPO (same budget) & $86.9 \pm 0.8$ & $98.1 \pm 0.1$ \\
\rowcolor{oursbg} \textbf{DARS (ours)} & $\mathbf{96.9 \pm 0.0}$ & $\mathbf{98.6 \pm 0.2}$ \\
\rowcolor{oursbg} $\Delta$ vs.\ same budget & $\mathbf{+10.0}$ & $\mathbf{+0.5}$ \\
\bottomrule
\multicolumn{3}{p{\dimexpr\linewidth-2\tabcolsep}}{\footnotesize Steps: 150 for
published rows; 500 for all other rows.}
\end{tabularx}
\end{minipage}\hfill
\begin{minipage}[t]{0.41\textwidth}
\centering
\textbf{(b)} Search-R1 (7B): QA accuracy\\[2pt]
\begin{tabularx}{\linewidth}{lCC}
\toprule
Method & Shared & Held-out \\
\midrule
GiGPO & $38.7 \pm 0.4$ & $39.5 \pm 0.4$ \\
\rowcolor{oursbg} \textbf{DARS (ours)} & $\mathbf{43.1 \pm 0.5}$ & $\mathbf{41.6 \pm 0.6}$ \\
\rowcolor{oursbg} $\Delta$ vs.\ GiGPO & $\mathbf{+4.4}$ & $\mathbf{+2.1}$ \\
\midrule
\emph{Shared pool} & Single-hop & Multi-hop \\
GiGPO & $49.5 \pm 0.5$ & $30.7 \pm 0.7$ \\
\rowcolor{oursbg} \textbf{DARS (ours)} & $\mathbf{53.2 \pm 0.1}$ & $\mathbf{35.5 \pm 0.8}$ \\
\rowcolor{oursbg} $\Delta$ vs.\ GiGPO & $\mathbf{+3.7}$ & $\mathbf{+4.9}$ \\
\bottomrule
\end{tabularx}
\end{minipage}

\vspace{6pt}
\textbf{(c)} WebShop: success and task score\\[2pt]
\begin{tabularx}{\textwidth}{lCCCC}
\toprule
 & \multicolumn{2}{c}{1.5B} & \multicolumn{2}{c}{7B} \\
\cmidrule(lr){2-3}\cmidrule(lr){4-5}
Method & Success & Task score & Success & Task score \\
\midrule
GRPO & $61.9 \pm 1.0$ & $76.6 \pm 0.8$ & $66.1 \pm 3.7$\rlap{$^{\dagger}$} & $79.3 \pm 2.8$\rlap{$^{\dagger}$} \\
GiGPO (w/ std)$^{\dagger}$ & $65.0 \pm 3.2$ & $83.1 \pm 1.6$ & $72.8 \pm 3.2$ & $84.4 \pm 2.9$ \\
GiGPO (w/o std)$^{\dagger}$ & $67.4 \pm 4.5$ & $83.5 \pm 1.8$ & $75.2 \pm 3.8$ & $86.2 \pm 2.6$ \\
GiGPO (parent) & $81.5 \pm 0.8$ & $92.4 \pm 0.3$ & $84.3 \pm 0.7$ & $93.1 \pm 0.3$ \\
\rowcolor{oursbg} \textbf{DARS (ours)}$^{\ddagger}$ & $\mathbf{83.1 \pm 0.7}$ & $\mathbf{93.4 \pm 0.3}$ & $\mathbf{88.5 \pm 0.7}$ & $\mathbf{94.6 \pm 0.4}$ \\
\rowcolor{oursbg} $\Delta$ vs.\ parent & $\mathbf{+1.5}$ & $\mathbf{+1.1}$ & $\mathbf{+4.2}$ & $\mathbf{+1.5}$ \\
\bottomrule
\multicolumn{5}{p{\dimexpr\textwidth-2\tabcolsep}}{\footnotesize Steps: 150 for
published rows and 1.5B GRPO; parents 400 (1.5B) / 600 (7B), then 400 / 250 DARS steps.}
\end{tabularx}
\end{table}


\subsection{Interactive Agent Results}
\label{sec:exp-main}

\paragraph{ALFWorld.}
Dependency-aware credit supports both earlier learning and continued
improvement. At 1.5B, DARS leads GiGPO at every milestone; GiGPO peaks at
step~200 and stays below that peak, whereas DARS improves again after
step~300 (Table~\ref{tab:alfworld-milestones}) and finishes at $96.9\%$
versus $86.9\%$, narrowing the gap to the 7B policies
(Table~\ref{tab:main-gigpo}). At 7B, GiGPO already reaches $98.1\%$, so the
smaller margin reflects benchmark saturation rather than evidence that
dependency-aware credit becomes unnecessary for larger models. These
trajectories come from single training runs; repeated decoding does not
measure training-seed variability.

\paragraph{WebShop.}
Faster learning and higher eventual performance are different benefits.
In separate 1.5B runs from initialization with an additive reward, DARS
leads its matched GiGPO twin by five success points at step~150, but the
twin catches up by step~200 (Appendix~\ref{app:webshop-scratch}). The
continuations of Table~\ref{tab:main-gigpo} answer a different question:
additional GiGPO training explains the 1.5B gain over the parent, whereas
at 7B DARS retains a $1.5$-point success advantage over an equally extended
control ($p<10^{-5}$; Appendix~\ref{app:webshop-control}). The metrics also
differ: from 7B initialization, task score improves by $2.1$ points while
strict success is unresolved (Appendix~\ref{app:local-baselines}), so
satisfying more requirements need not mean completing more purchases.

\paragraph{Search-R1.}
Rewarding intermediate progress helps only while it stays tied to
completing the task. On the held-out split, the commitment terms alone
stop the policy from searching, graph credit alone eventually makes it
search repeatedly without answering, and only their combination exceeds
GiGPO ($41.6\%$ vs.\ $39.5\%$ in all three decoding groups;
Table~\ref{tab:search-heldout}). The policy must learn both to establish
an answer's prerequisites and to convert that progress into an answer.
Larger multi-hop gains support this reading but vary across datasets, and
2Wiki declines (Appendix~\ref{app:search-detail}).

\paragraph{Behavior.}
In paired ALFWorld~1.5B episodes, DARS makes fewer wrong-object pickups
than GiGPO (1 vs.\ 10) and GRPO (1 vs.\ 12); on 350 held-out Search-R1
questions from a later run, it searches before every answer, whereas GiGPO
answers without searching in $31\%$ of cases. These probes are descriptive
rather than causal (Appendix~\ref{app:qualitative}).

\subsection{Mathematical Reasoning}
\label{sec:exp-reasoning}

\begin{table}[t]
\centering
\footnotesize
\renewcommand{\arraystretch}{0.88}
\caption{
\textbf{Agentic mathematics with ARPO/AEPO} (Qwen3-8B, Python interpreter).
Mean AIME24/25 accuracy (\%); approximate decoding variation $\pm5$ points per cell; our methods are shaded.
All arms retain ARPO's rollouts. ``Best'' is the maximum across the displayed checkpoints
on the reported problems; $\Delta$ is its gain over initialization
($57.9$), in points. Protocols, per-benchmark results, and checkpoint-selection
rationale: Appendix~\ref{app:sanity}.
}
\label{tab:main-aepo}
\setlength{\tabcolsep}{3.5pt}
\begin{tabular}{@{}lcccccc@{}}
\toprule
Arm & Step 5 & Step 10 & Step 15 & Grid mean & Best & $\Delta$ vs.\ init.\ \\
\midrule
ARPO \citep{dong2025arpo}   & 61.7 & 61.3 & 62.9 & 62.0 & 62.9 & $+5.0$ \\
AEPO \citep{dong2026aepo}   & \textbf{63.3} & 61.7 & 60.0 & 61.7 & 63.3 & $+5.4$ \\
\rowcolor{oursbg}ARPO $+$ DARS               & 59.2 & 61.3 & \textbf{64.2} & 61.6 & 64.2 & $+6.3$ \\
\rowcolor{oursbg}AEPO $+$ DARS               & 60.4 & \textbf{67.5} & 62.9 & \textbf{63.6} & \textbf{67.5} & $\mathbf{+9.6}$ \\
\bottomrule
\end{tabular}
\end{table}

\paragraph{Agentic mathematics.}
Progress rewards can complement entropy-based training, but the benefit
depends on their interaction. Used as a dense trajectory-return
augmentation (Section~\ref{sec:method-integration}), DARS raises AEPO's
grid mean from $61.7\%$ to $63.6\%$, the highest in the $2\times2$ design,
but not ARPO's ($62.0\%$ to $61.6\%$), and it does not help at every
checkpoint (Table~\ref{tab:main-aepo}). Entropy mechanisms respond to
uncertainty while DARS assesses progress toward a solution; given
checkpoint variation and about $\pm5$ points of decoding variation per
cell, this suggests complementarity rather than established synergy. The
combined arm's best score of $67.5\%$ is consistent with this reading
(Appendix~\ref{app:sanity}).

\begin{table}[t]
\centering
\footnotesize
\renewcommand{\arraystretch}{0.88}
\setlength{\tabcolsep}{4.5pt}
\caption{
\textbf{Tool-free mathematical reasoning.} Percentage scores averaged
over each method's evaluation cells;
Avg.\ is the mean of AMC, AIME24, and AIME25.
Bold marks the best score within each model size. Our method is shaded.
Protocols, cell counts, and uncertainty: Appendix~\ref{app:reasoning}.
}
\label{tab:main-math}
\begin{tabular}{lcccc|cccc}
\toprule
 & \multicolumn{4}{c|}{Qwen3-1.7B} & \multicolumn{4}{c}{Qwen3-4B} \\
Method & AMC & AIME24 & AIME25 & Avg. & AMC & AIME24 & AIME25 & Avg. \\
\midrule
Untrained model          & 84.38 & 50.56 & 37.50 & 57.48 & 97.21 & 73.75 & 66.55 & 79.17 \\
\midrule
StaRPO                   & 84.37 & 50.42 & 37.50 & 57.43 & 97.03 & 72.40 & \textbf{67.40} & 78.94 \\
On-policy distillation   & 84.45 & \textbf{50.62} & 36.88 & 57.32 & 96.83 & 71.64 & 67.39 & 78.62 \\
OmniOPD                  & 85.52 & 49.86 & 36.34 & 57.24 & 96.33 & 72.43 & 66.56 & 78.44 \\
DARS w/o step credit     & 84.94 & 49.25 & 36.50 & 56.90 & \textbf{97.48} & 74.20 & 65.71 & 79.13 \\
\rowcolor{oursbg}\textbf{DARS (ours)}     & \textbf{86.53} & 49.92 & \textbf{38.25} & \textbf{58.23} & 96.72 & \textbf{74.64} & 67.12 & \textbf{79.50} \\
\bottomrule
\end{tabular}
\end{table}

\paragraph{Tool-free reasoning.}
Structured feedback remains useful inside a single response, and the
clearest evidence concerns how it is used. DARS has the highest mean at
both scales and exceeds OmniOPD at 1.7B ($p=0.031$) and 4B ($p=0.005$ on
the primary host; Appendix~\ref{app:reasoning}). The same-annotation
scalar control of Section~\ref{sec:exp-ablation} shows that placing credit
on steps matters, clearly at 1.7B but not resolved at 4B; because few
credited segments are negative, these runs mainly test the reinforcement
of useful steps rather than error repair.

\subsection{Ablations}
\label{sec:exp-ablation}

\begin{table}[t]
\centering
\footnotesize
\caption{\textbf{Ablations and annotator comparison.}
All scores are percentages; Avg.\ in (a) is the mean of AMC, AIME24,
and AIME25; DARS rows are shaded.
The sweep in (b) reports validation-selected peaks under a separate recipe ($\Delta$ before rounding).
Panel (d) reports peak success
(eight draws) and training-step time from the matched comparison.
Protocols, uncertainty, and tests: Appendices~\ref{app:reasoning},
\ref{app:dependency-ablation}, \ref{app:annot-quality}, and~\ref{app:cost}.}
\label{tab:dependency-ablation}
\setlength{\tabcolsep}{3pt}
\renewcommand{\arraystretch}{1.0}

\textbf{(a) Step credit: mathematical reasoning}\par\smallskip
\begin{tabularx}{\linewidth}{@{}l*{8}{R}@{}}
\toprule
& \multicolumn{4}{c}{Qwen3-1.7B} & \multicolumn{4}{c}{Qwen3-4B} \\
\cmidrule(lr){2-5}\cmidrule(l){6-9}
Variant & AMC & AIME24 & AIME25 & Avg. & AMC & AIME24 & AIME25 & Avg. \\
\midrule
Scalar only ($\eta=0$) & 84.94 & 49.25 & 36.50 & 56.90 & \textbf{97.48} & 74.20 & 65.71 & 79.13 \\
\rowcolor{oursbg}\textbf{DARS} ($\eta=1$) & \textbf{86.53} & \textbf{49.92} & \textbf{38.25} & \textbf{58.23} & 96.72 & \textbf{74.64} & \textbf{67.12} & \textbf{79.50} \\
\bottomrule
\end{tabularx}

\par\medskip
\begin{minipage}[t]{0.48\linewidth}
\vspace{0pt}
\textbf{(b) Attenuation: ALFWorld success}\par\smallskip
\begin{tabularx}{\linewidth}{@{}l*{2}{R}@{}}
\toprule
Variant & 1.5B & 7B \\
\midrule
No decay ($\lambda=0$) & 86.0 & 95.2 \\
\rowcolor{oursbg}\textbf{DARS} ($\lambda=1$) & \textbf{92.0} & \textbf{98.6} \\
\bottomrule
\end{tabularx}
\par\medskip
\textit{Separate 1.5B sweep: distilled annotator, 150 steps}\par\smallskip
\begin{tabular*}{\linewidth}{@{\extracolsep{\fill}}lrrrr@{}}
\toprule
$\lambda$ & $0$ & $0.5$ & $1$ & $\infty$ \\
\midrule
Success & 90.2 & \textbf{94.3} & 91.4 & 84.2 \\
$\Delta$ vs.\ $\lambda=0$ & -- & $+4.1$ & $+1.2$ & $-6.1$ \\
\bottomrule
\end{tabular*}
\end{minipage}
\hfill
\begin{minipage}[t]{0.48\linewidth}
\vspace{0pt}
\textbf{(c) Topology: WebShop 1.5B}\par\smallskip
\begin{tabularx}{\linewidth}{@{}l*{2}{R}@{}}
\toprule
Graph & Success & Task score \\
\midrule
Sequential chain & 49.0 & 61.1 \\
\rowcolor{oursbg}\textbf{DARS} (requirement graph) & \textbf{63.2} & \textbf{79.5} \\
\bottomrule
\end{tabularx}
\par\medskip
\textbf{(d) Annotator: ALFWorld 1.5B}\par\smallskip
\begin{tabular*}{\linewidth}{@{\extracolsep{\fill}}lrr@{}}
\toprule
Annotator & Success & Time/step (s) \\
\midrule
DeepSeek-V4-Flash & 89.1 & 512 \\
Distilled Qwen3-8B & 92.2 & 506 \\
\bottomrule
\end{tabular*}
\end{minipage}
\end{table}

\textbf{Credit placement matters even with the same annotations.}
The scalar control ($\eta=0$) uses the same graphs and events as DARS but
assigns one reward per response; restoring step-level credit raises the
average from $56.90$ to $58.23$ at 1.7B and from $79.13$ to $79.50$ at 4B
(Table~\ref{tab:dependency-ablation}(a)). Collapsing structured feedback into
a trajectory score thus discards information useful for learning.

\textbf{Discounting dependent progress works better than discarding it.}
Without attenuation ($\lambda=0$), ALFWorld success drops from $92.0\%$ to
$86.0\%$ at 1.5B and from $98.6\%$ to $95.2\%$ at 7B
(Table~\ref{tab:dependency-ablation}(b); DARS ahead on 4/4 paired draws at
1.5B and 15/15 at 7B, $p<0.001$), although the rule
changes only $2$--$12\%$ of reward vectors: those in which an unresolved
error is followed by dependent work. In a separate sweep, moderate
attenuation ($94.3\%$ at $\lambda=0.5$, $91.4\%$ at $\lambda=1$) beats both
no attenuation ($90.2\%$) and hard masking ($84.2\%$), so stronger
suppression is not better and $\lambda=1$ is a reasonable default rather
than an optimum (Appendix~\ref{app:dependency-ablation}).

\textbf{Representing independence matters as much as dependence.}
Replacing WebShop's requirement graph with a trajectory-ordered chain
lowers success by $14.2$ points and task score by $18.4$
(Table~\ref{tab:dependency-ablation}(c)). The chain lets an error reach later
predicates merely because they come later; offline, a quarter of the
affected trajectories lose more than half their credit. A graph-free
first-error prefix reward shows the same pattern: it helps ALFWorld, whose
goals are chains, but hurts WebShop and Search-R1
(Appendix~\ref{app:topology-diagnostics}).

\textbf{Structured annotation can be distilled.}
A Qwen3-8B annotator distilled from DeepSeek-V4-Flash reproduces its
teacher's step-level credit at $r=0.81$ (teacher self-agreement $0.86$) and
reaches $92.2\%$ peak success against $89.1\%$ for the teacher
(Table~\ref{tab:dependency-ablation}(d)), a difference within decoding noise.
This shows practical sufficiency, not superiority
(Appendix~\ref{app:annot-quality}).

\textbf{DARS adds little training cost.} The annotator is called once per
completed trajectory (on ALFWorld~1.5B only for failures, from about 116
to 14 calls per step as the policy improves), and policy generation
remains about $75\%$ of a step; a DARS step thus takes $340$--$420$\,s
against $270$--$400$\,s for GiGPO (Appendix~\ref{app:cost}). On
mathematics, DARS annotates each rollout once, whereas OmniOPD runs a 32B
teacher at every step.

\textbf{Different tasks exercise different parts of the mechanism.}
Only $3.5\%$ and $1.2\%$ of credited mathematics segments receive negative
credit at 1.7B and 4B, so those runs mainly test the placement of positive
credit, while WebShop isolates topology and ALFWorld attenuation
(Appendix~\ref{app:reasoning}).

\section{Conclusion and Limitations}

DARS represents task completion as a dependency graph,
where task predicates are connected by prerequisite relations. It converts
the verification, invalidation, and repair of predicates into signed step
rewards. Combined with the outcome signal, these rewards help the policy
learn which steps make valid progress, which mistakes invalidate dependent
work, and which actions repair earlier errors. Across five task families,
DARS improves matched baselines on ALFWorld and Search-R1, raises the
WebShop task score, works with both GiGPO and AEPO, and achieves the highest
average score among the compared tool-free reasoning methods. Ablations
further show that step-level credit, dependency attenuation, and graph
topology each contribute to performance. These results demonstrate that
explicitly modeling task dependencies provides a practical way to assign
credit beyond terminal rewards.

Two limitations remain. First, we do not systematically explore scaling: whether
larger annotators improve supervision quality or larger policies benefit
similarly from DARS remains an open question. Second, our evaluation
focuses on tasks with identifiable predicates and
prerequisite relations. In open-ended tasks, predicates may lack clear
logical dependencies, making graph construction and dependency-based
credit assignment less straightforward. The applicability of DARS to such
tasks remains to be established.

\subsection*{AI use statement}
Generative AI tools assisted with language editing, literature organization,
and LaTeX source restructuring. The authors reviewed the AI-assisted work
and remain responsible for the paper's text, citations, claims, code, and
experimental artifacts.

\subsection*{Reproducibility statement}
Section~\ref{sec:exp-setup} summarizes the tasks, models, and comparisons.
Appendix~\ref{app:protocol} records training settings, decoding parameters,
checkpoint selection, and reward configurations for every arm, including
the local baselines. Appendix~\ref{app:prompts} gives the verbatim
annotation prompts, the output contract, and the state-replay procedure,
which together fully determine the reward from an annotation.
Appendices~\ref{app:annot-quality}--\ref{app:qualitative} provide the
annotation audits, cost measurements, per-benchmark results, ablation
protocols, and recorded trajectories supporting the reported comparisons.
Code for the annotator, the reward kernel, and the learner integrations,
together with the persisted annotation graphs used by the fixed-annotation
ablations, will be released.

\bibliographystyle{iclr2027_conference}
\bibliography{paper_references}

@article{abdin2025phi4reasoning,
  title   = {{Phi-4-reasoning} Technical Report},
  author  = {Abdin, Marah and Agarwal, Sahaj and Awadallah, Ahmed and
             Balachandran, Vidhisha and Behl, Harkirat and Chen, Lingjiao and
             de Rosa, Gustavo and Gunasekar, Suriya and Javaheripi, Mojan and
             Joshi, Neel and Kauffmann, Piero and Lara, Yash and
             Mendes, Caio C{\'e}sar Teodoro and Mitra, Arindam and Nushi, Besmira and
             Papailiopoulos, Dimitris and Saarikivi, Olli and Shah, Shital and
             Shrivastava, Vaishnavi and Vineet, Vibhav and Wu, Yue and
             Yousefi, Safoora and Zheng, Guoqing},
  journal = {arXiv preprint arXiv:2504.21318},
  year    = {2025},
  url     = {https://arxiv.org/abs/2504.21318},
}

@inproceedings{petrenko2026entropypreserving,
  title     = {Entropy-Preserving Reinforcement Learning},
  author    = {Petrenko, Aleksei and Lipkin, Ben and Chen, Kevin and Wijmans, Erik and
               Cusumano-Towner, Marco and Giryes, Raja and Kr{\"a}henb{\"u}hl, Philipp},
  booktitle = {International Conference on Learning Representations (ICLR)},
  year      = {2026},
  note      = {arXiv:2603.11682},
  url       = {https://arxiv.org/abs/2603.11682},
}

@inproceedings{yao2023react,
  title     = {{ReAct}: Synergizing Reasoning and Acting in Language Models},
  author    = {Yao, Shunyu and Zhao, Jeffrey and Yu, Dian and Du, Nan and Shafran, Izhak and
               Narasimhan, Karthik and Cao, Yuan},
  booktitle = {International Conference on Learning Representations (ICLR)},
  year      = {2023},
  note      = {arXiv:2210.03629},
}

@inproceedings{yao2022webshop,
  title     = {{WebShop}: Towards Scalable Real-World Web Interaction with Grounded Language Agents},
  author    = {Yao, Shunyu and Chen, Howard and Yang, John and Narasimhan, Karthik},
  booktitle = {Advances in Neural Information Processing Systems (NeurIPS)},
  year      = {2022},
  note      = {arXiv:2207.01206},
}

@inproceedings{shridhar2021alfworld,
  title     = {{ALFWorld}: Aligning Text and Embodied Environments for Interactive Learning},
  author    = {Shridhar, Mohit and Yuan, Xingdi and C{\^o}t{\'e}, Marc-Alexandre and Bisk, Yonatan and
               Trischler, Adam and Hausknecht, Matthew},
  booktitle = {International Conference on Learning Representations (ICLR)},
  year      = {2021},
  note      = {arXiv:2010.03768},
}

@article{jin2025searchr1,
  title   = {{Search-R1}: Training {LLM}s to Reason and Leverage Search Engines with Reinforcement
             Learning},
  author  = {Jin, Bowen and Zeng, Hansi and Yue, Zhenrui and Yoon, Jinsung and Arik, Sercan and
             Wang, Dong and Zamani, Hamed and Han, Jiawei},
  journal = {arXiv preprint arXiv:2503.09516},
  year    = {2025},
}

@article{zhang2025agenticsurvey,
  title   = {The Landscape of Agentic Reinforcement Learning for {LLM}s: A Survey},
  author  = {Zhang, Guibin and Hu, Muxin and Wan, Guancheng and Yu, Hanjie and others},
  journal = {Transactions on Machine Learning Research (TMLR)},
  year    = {2025},
  note    = {arXiv:2509.02547},
}

@article{shao2024deepseekmath,
  title   = {{DeepSeekMath}: Pushing the Limits of Mathematical Reasoning in Open Language Models},
  author  = {Shao, Zhihong and Wang, Peiyi and Zhu, Qihao and Xu, Runxin and Song, Junxiao and
             Bi, Xiao and Zhang, Haowei and Zhang, Mingchuan and Li, Y. K. and Wu, Y. and
             Guo, Daya},
  journal = {arXiv preprint arXiv:2402.03300},
  year    = {2024},
}

@article{deepseekai2025r1,
  title   = {{DeepSeek-R1} Incentivizes Reasoning in {LLM}s through Reinforcement Learning},
  author  = {Guo, Daya and Yang, Dejian and Zhang, Haowei and Song, Junxiao and Wang, Peiyi and
             Zhu, Qihao and Xu, Runxin and Zhang, Ruoyu and Ma, Shirong and Bi, Xiao and others},
  journal = {Nature},
  volume  = {645},
  number  = {8081},
  pages   = {633--638},
  year    = {2025},
  doi     = {10.1038/s41586-025-09422-z},
  note    = {arXiv:2501.12948},
}

@article{lambert2024tulu3,
  title   = {{T\"ulu} 3: Pushing Frontiers in Open Language Model Post-Training},
  author  = {Lambert, Nathan and Morrison, Jacob and Pyatkin, Valentina and Huang, Shengyi and
             Ivison, Hamish and Brahman, Faeze and others},
  journal = {arXiv preprint arXiv:2411.15124},
  year    = {2024},
}

@inproceedings{yu2025dapo,
  title     = {{DAPO}: An Open-Source {LLM} Reinforcement Learning System at Scale},
  author    = {Yu, Qiying and Zhang, Zheng and Zhu, Ruofei and Yuan, Yufeng and Zuo, Xiaochen and
               Yue, Yu and others},
  booktitle = {Advances in Neural Information Processing Systems (NeurIPS)},
  year      = {2025},
  note      = {arXiv:2503.14476},
}

@article{shen2025carl,
  title   = {{CARL}: Criticality-Aware Agentic Reinforcement Learning},
  author  = {Shen, Leyang and Zhang, Yang and Ling, Chun Kai and Zhao, Xiaoyan and Chua, Tat-Seng},
  journal = {arXiv preprint arXiv:2512.04949},
  year    = {2025},
}

@inproceedings{cheng2026graphcredit,
  title     = {Beyond Trajectory-Level Attribution: Graph-Based Credit Assignment for Agentic
               Reinforcement Learning},
  author    = {Cheng, Xin and He, Shuo and Feng, Lang and Xu, Haiyang and Yan, Ming and
               Feng, Lei and An, Bo},
  booktitle = {International Conference on Machine Learning (ICML)},
  year      = {2026},
  note      = {arXiv:2605.26684},
}

@article{zhu2025agentfail,
  title   = {Where {LLM} Agents Fail and How They Can Learn From Failures},
  author  = {Zhu, Kunlun and Liu, Zijia and Li, Bingxuan and Tian, Muxin and Yang, Yingxuan and
             Zhang, Jiaxun and Han, Pengrui and Xie, Qipeng and Cui, Fuyang and Zhang, Weijia and
             Ma, Xiaoteng and Yu, Xiaodong and Ramesh, Gowtham and Wu, Jialian and Liu, Zicheng and
             Lu, Pan and Zou, James and You, Jiaxuan},
  journal = {arXiv preprint arXiv:2509.25370},
  year    = {2025},
}

@article{liang2026elpo,
  title   = {Learning from the Irrecoverable: Error-Localized Policy Optimization for
             Tool-Integrated {LLM} Reasoning},
  author  = {Liang, Qiao and Zhu, Yuke and Ge, Chao and Yang, Lei and Shen, Ying and Zheng, Bo and
             Guo, Sheng},
  journal = {arXiv preprint arXiv:2602.09598},
  year    = {2026},
}

@inproceedings{feng2025gigpo,
  title     = {Group-in-Group Policy Optimization for {LLM} Agent Training},
  author    = {Feng, Lang and Xue, Zhenghai and Liu, Tingcong and An, Bo},
  booktitle = {Advances in Neural Information Processing Systems (NeurIPS)},
  year      = {2025},
  note      = {arXiv:2505.10978},
}

@inproceedings{dong2025arpo,
  title     = {Agentic Reinforced Policy Optimization},
  author    = {Dong, Guanting and Mao, Hangyu and Ma, Kai and Bao, Licheng and Chen, Yifei and
               Wang, Zhongyuan and Chen, Zhongxia and Du, Jiazhen and Wang, Huiyang and
               Zhang, Fuzheng and Zhou, Guorui and Zhu, Yutao and Wen, Ji-Rong and Dou, Zhicheng},
  booktitle = {International Conference on Learning Representations (ICLR)},
  year      = {2026},
  note      = {arXiv:2507.19849},
}

@inproceedings{dong2026aepo,
  title     = {Toward Generalized Web Agent Training: A Deep Dive into Entropy-Balanced
               Reinforcement Learning},
  author    = {Dong, Guanting and Bao, Licheng and Wang, Zhongyuan and Zhao, Kangzhi and
               Li, Xiaoxi and Jin, Jiajie and Yang, Jinghan and Mao, Hangyu and Zhang, Fuzheng and
               Gai, Kun and Zhou, Guorui and Zhu, Yutao and Wen, Ji-Rong and Dou, Zhicheng},
  booktitle = {Proceedings of the ACM Web Conference (WWW)},
  year      = {2026},
  doi       = {10.1145/3774904.3792301},
  note      = {arXiv:2510.14545},
}

@inproceedings{wang2025beyond,
  title     = {Beyond the 80/20 Rule: High-Entropy Minority Tokens Drive Effective Reinforcement
               Learning for {LLM} Reasoning},
  author    = {Wang, Shenzhi and Yu, Le and Gao, Chang and Zheng, Chujie and Liu, Shixuan and
               Lu, Rui and Dang, Kai and Chen, Xionghui and Yang, Jianxin and Zhang, Zhenru and
               Liu, Yuqiong and Yang, An and Zhao, Andrew and Yue, Yang and Song, Shiji and
               Yu, Bowen and Huang, Gao and Lin, Junyang},
  booktitle = {Advances in Neural Information Processing Systems (NeurIPS)},
  year      = {2025},
  note      = {arXiv:2506.01939},
}

@article{wang2025hicra,
  title   = {Emergent Hierarchical Reasoning in {LLM}s through Reinforcement Learning},
  author  = {Wang, Haozhe and Xu, Qixin and Liu, Che and Wu, Junhong and Lin, Fangzhen and
             Chen, Wenhu},
  journal = {arXiv preprint arXiv:2509.03646},
  year    = {2025},
}

@article{li2026notalltokens,
  title   = {Not All Tokens Learn Alike: Attention Entropy Reveals Heterogeneous Signals in
             {RL} Reasoning},
  author  = {Li, Gengyang and Wu, Zheng-Fan and Bao, Siqi and Wu, Yunfang},
  journal = {arXiv preprint arXiv:2605.07660},
  year    = {2026},
}

@inproceedings{lightman2023letsverify,
  title     = {Let's Verify Step by Step},
  author    = {Lightman, Hunter and Kosaraju, Vineet and Burda, Yura and Edwards, Harri and
               Baker, Bowen and Lee, Teddy and Leike, Jan and Schulman, John and Sutskever, Ilya and
               Cobbe, Karl},
  booktitle = {International Conference on Learning Representations (ICLR)},
  year      = {2024},
  note      = {arXiv:2305.20050},
}

@inproceedings{zheng2023judging,
  title     = {Judging {LLM}-as-a-Judge with {MT-Bench} and {Chatbot Arena}},
  author    = {Zheng, Lianmin and Chiang, Wei-Lin and Sheng, Ying and Zhuang, Siyuan and
               Wu, Zhanghao and Zhuang, Yonghao and others},
  booktitle = {Advances in Neural Information Processing Systems (NeurIPS), Datasets and Benchmarks},
  year      = {2023},
  note      = {arXiv:2306.05685},
}

@article{gu2024judgesurvey,
  title   = {A Survey on {LLM}-as-a-Judge},
  author  = {Gu, Jiawei and Jiang, Xuhui and Shi, Zhichao and Tan, Hexiang and Zhai, Xuehao and
             Xu, Chengjin and others},
  journal = {arXiv preprint arXiv:2411.15594},
  year    = {2024},
}

@inproceedings{zhang2025failureattribution,
  title     = {Which Agent Causes Task Failures and When? On Automated Failure Attribution of
               {LLM} Multi-Agent Systems},
  author    = {Zhang, Shaokun and Yin, Ming and Zhang, Jieyu and Liu, Jiale and Han, Zhiguang and
               others},
  booktitle = {International Conference on Machine Learning (ICML)},
  year      = {2025},
  note      = {arXiv:2505.00212},
}

@article{minsky1961steps,
  title   = {Steps Toward Artificial Intelligence},
  author  = {Minsky, Marvin},
  journal = {Proceedings of the IRE},
  volume  = {49},
  number  = {1},
  pages   = {8--30},
  year    = {1961},
}

@article{sutton1988td,
  title   = {Learning to Predict by the Methods of Temporal Differences},
  author  = {Sutton, Richard S.},
  journal = {Machine Learning},
  volume  = {3},
  pages   = {9--44},
  year    = {1988},
}

@inproceedings{ng1999policy,
  title     = {Policy Invariance Under Reward Transformations: Theory and Application to Reward
               Shaping},
  author    = {Ng, Andrew Y. and Harada, Daishi and Russell, Stuart},
  booktitle = {International Conference on Machine Learning (ICML)},
  pages     = {278--287},
  year      = {1999},
}

@inproceedings{wiewiora2003principled,
  title     = {Principled Methods for Advising Reinforcement Learning Agents},
  author    = {Wiewiora, Eric and Cottrell, Garrison W. and Elkan, Charles},
  booktitle = {International Conference on Machine Learning (ICML)},
  pages     = {792--799},
  year      = {2003},
}

@inproceedings{ross2011dagger,
  title     = {A Reduction of Imitation Learning and Structured Prediction to No-Regret Online
               Learning},
  author    = {Ross, St{\'e}phane and Gordon, Geoffrey J. and Bagnell, J. Andrew},
  booktitle = {International Conference on Artificial Intelligence and Statistics (AISTATS)},
  series    = {PMLR},
  volume    = {15},
  pages     = {627--635},
  year      = {2011},
  note      = {arXiv:1011.0686},
}

@inproceedings{bachmann2024pitfalls,
  title     = {The Pitfalls of Next-Token Prediction},
  author    = {Bachmann, Gregor and Nagarajan, Vaishnavh},
  booktitle = {International Conference on Machine Learning (ICML)},
  year      = {2024},
  note      = {arXiv:2403.06963},
}

@article{kowalski1979algorithm,
  title   = {Algorithm = Logic + Control},
  author  = {Kowalski, Robert},
  journal = {Communications of the ACM},
  volume  = {22},
  number  = {7},
  pages   = {424--436},
  year    = {1979},
}

@article{uesato2022process,
  title   = {Solving Math Word Problems with Process- and Outcome-Based Feedback},
  author  = {Uesato, Jonathan and Kushman, Nate and Kumar, Ramana and Song, Francis and
             Siegel, Noah and Wang, Lisa and Creswell, Antonia and Irving, Geoffrey and
             Higgins, Irina},
  journal = {arXiv preprint arXiv:2211.14275},
  year    = {2022},
}

@inproceedings{wang2024mathshepherd,
  title     = {{Math-Shepherd}: Verify and Reinforce {LLM}s Step-by-step without Human Annotations},
  author    = {Wang, Peiyi and Li, Lei and Shao, Zhihong and Xu, R. X. and Dai, Damai and
               Li, Yifei and Chen, Deli and Wu, Y. and Sui, Zhifang},
  booktitle = {Proceedings of the 62nd Annual Meeting of the Association for Computational
               Linguistics (ACL)},
  pages     = {9426--9439},
  year      = {2024},
  doi       = {10.18653/v1/2024.acl-long.510},
  note      = {arXiv:2312.08935},
}

@inproceedings{setlur2024pavs,
  title     = {Rewarding Progress: Scaling Automated Process Verifiers for {LLM} Reasoning},
  author    = {Setlur, Amrith and Nagpal, Chirag and Fisch, Adam and Geng, Xinyang and
               Eisenstein, Jacob and Agarwal, Rishabh and Agarwal, Alekh and Berant, Jonathan and
               Kumar, Aviral},
  booktitle = {International Conference on Learning Representations (ICLR)},
  year      = {2025},
  note      = {arXiv:2410.08146},
}

@inproceedings{zheng2024processbench,
  title     = {{ProcessBench}: Identifying Process Errors in Mathematical Reasoning},
  author    = {Zheng, Chujie and Zhang, Zhenru and Zhang, Beichen and Lin, Runji and Lu, Keming and
               Yu, Bowen and Liu, Dayiheng and Zhou, Jingren and Lin, Junyang},
  booktitle = {Proceedings of the Annual Meeting of the Association for Computational Linguistics
               (ACL)},
  year      = {2025},
  note      = {arXiv:2412.06559},
}

@article{cui2025prime,
  title   = {Process Reinforcement through Implicit Rewards},
  author  = {Cui, Ganqu and Yuan, Lifan and Wang, Zefan and Wang, Hanbin and Li, Wendi and
             He, Bingxiang and Fan, Yuchen and Yu, Tianyu and Xu, Qixin and Chen, Weize and others},
  journal = {arXiv preprint arXiv:2502.01456},
  year    = {2025},
}

@article{yuan2024implicitprm,
  title   = {Free Process Rewards without Process Labels},
  author  = {Yuan, Lifan and Li, Wendi and Chen, Huayu and Cui, Ganqu and Ding, Ning and
             Zhang, Kaiyan and Zhou, Bowen and Liu, Zhiyuan and Peng, Hao},
  journal = {arXiv preprint arXiv:2412.01981},
  year    = {2024},
}

@inproceedings{harutyunyan2019hca,
  title     = {Hindsight Credit Assignment},
  author    = {Harutyunyan, Anna and Dabney, Will and Mesnard, Thomas and Azar, Mohammad and
               Piot, Bilal and Heess, Nicolas and van Hasselt, Hado and Wayne, Greg and
               Singh, Satinder and Precup, Doina and Munos, R{\'e}mi},
  booktitle = {Advances in Neural Information Processing Systems (NeurIPS)},
  year      = {2019},
  note      = {arXiv:1912.02503},
}

@inproceedings{meulemans2023cocoa,
  title     = {Would I Have Gotten That Reward? Long-term Credit Assignment by Counterfactual
               Contribution Analysis},
  author    = {Meulemans, Alexander and Schug, Simon and Kobayashi, Seijin and Daw, Nathaniel and
               Wayne, Gregory},
  booktitle = {Advances in Neural Information Processing Systems (NeurIPS)},
  year      = {2023},
  note      = {arXiv:2306.16803},
}

@inproceedings{arjonamedina2019rudder,
  title     = {{RUDDER}: Return Decomposition for Delayed Rewards},
  author    = {Arjona-Medina, Jose A. and Gillhofer, Michael and Widrich, Michael and
               Unterthiner, Thomas and Brandstetter, Johannes and Hochreiter, Sepp},
  booktitle = {Advances in Neural Information Processing Systems (NeurIPS)},
  pages     = {13544--13555},
  year      = {2019},
  note      = {arXiv:1806.07857},
}

@inproceedings{devlin2012dynamic,
  title     = {Dynamic Potential-Based Reward Shaping},
  author    = {Devlin, Sam and Kudenko, Daniel},
  booktitle = {Proceedings of the International Conference on Autonomous Agents and Multiagent
               Systems (AAMAS)},
  pages     = {433--440},
  year      = {2012},
}

@article{wang2025sparl,
  title   = {{SPA-RL}: Reinforcing {LLM} Agents via Stepwise Progress Attribution},
  author  = {Wang, Hanlin and Leong, Chak Tou and Wang, Jiashuo and Wang, Jian and Li, Wenjie},
  journal = {arXiv preprint arXiv:2505.20732},
  year    = {2025},
}

@article{liu2025istar,
  title   = {Agentic Reinforcement Learning with Implicit Step Rewards},
  author  = {Liu, Xiaoqian and Wang, Ke and Wu, Yuchuan and Huang, Fei and Li, Yongbin and
             Zhang, Junge and Jiao, Jianbin},
  journal = {arXiv preprint arXiv:2509.19199},
  year    = {2025},
}

@inproceedings{peng2026hiper,
  title     = {{HiPER}: Hierarchical Reinforcement Learning with Explicit Credit Assignment for
               Large Language Model Agents},
  author    = {Peng, Jiangweizhi and Liu, Yuanxin and Zhou, Ruida and Fleming, Charles and
               Wang, Zhaoran and Garcia, Alfredo and Hong, Mingyi},
  booktitle = {International Conference on Machine Learning (ICML)},
  year      = {2026},
  note      = {arXiv:2602.16165},
}

@inproceedings{ji2026treegrpo,
  title     = {Tree Search for {LLM} Agent Reinforcement Learning},
  author    = {Ji, Yuxiang and Ma, Ziyu and Wang, Yong and Chen, Guanhua and Chu, Xiangxiang and
               Wu, Liaoni},
  booktitle = {International Conference on Learning Representations (ICLR)},
  year      = {2026},
  note      = {arXiv:2509.21240},
}

@inproceedings{zhou2024archer,
  title     = {{ArCHer}: Training Language Model Agents via Hierarchical Multi-Turn {RL}},
  author    = {Zhou, Yifei and Zanette, Andrea and Pan, Jiayi and Levine, Sergey and Kumar, Aviral},
  booktitle = {Proceedings of the 41st International Conference on Machine Learning (ICML)},
  series    = {PMLR},
  volume    = {235},
  pages     = {62178--62209},
  year      = {2024},
  note      = {arXiv:2402.19446},
}

@article{zhuge2024agentjudge,
  title   = {{Agent-as-a-Judge}: Evaluate Agents with Agents},
  author  = {Zhuge, Mingchen and Zhao, Changsheng and Ashley, Dylan and Wang, Wenyi and
             Khizbullin, Dmitrii and Xiong, Yunyang and Liu, Zechun and Chang, Ernie and
             Krishnamoorthi, Raghuraman and Tian, Yuandong and Shi, Yangyang and Chandra, Vikas and
             Schmidhuber, J{\"u}rgen},
  journal = {arXiv preprint arXiv:2410.10934},
  year    = {2024},
}

@inproceedings{kazemi2023lambada,
  title     = {{LAMBADA}: Backward Chaining for Automated Reasoning in Natural Language},
  author    = {Kazemi, Mehran and Kim, Najoung and Bhatia, Deepti and Xu, Xin and
               Ramachandran, Deepak},
  booktitle = {Proceedings of the Annual Meeting of the Association for Computational Linguistics
               (ACL)},
  year      = {2023},
  note      = {arXiv:2212.13894},
}

@inproceedings{agarwal2024gkd,
  title     = {On-Policy Distillation of Language Models: Learning from Self-Generated Mistakes},
  author    = {Agarwal, Rishabh and Vieillard, Nino and Zhou, Yongchao and Stanczyk, Piotr and
               Ramos, Sabela and Geist, Matthieu and Bachem, Olivier},
  booktitle = {International Conference on Learning Representations (ICLR)},
  year      = {2024},
  note      = {arXiv:2306.13649},
}

@inproceedings{gu2024minillm,
  title     = {{MiniLLM}: Knowledge Distillation of Large Language Models},
  author    = {Gu, Yuxian and Dong, Li and Wei, Furu and Huang, Minlie},
  booktitle = {International Conference on Learning Representations (ICLR)},
  year      = {2024},
  note      = {arXiv:2306.08543},
}

@article{omniopd2026,
  title   = {{OmniOPD}: Logit-Free On-Policy Distillation via Speculative Verification},
  author  = {Zhou, Yuhang and Zhang, Lizhu and Wu, Yifan and Wang, Mingyi and Bo, Peng and
             Liu, Jiayi and Fan, Xiangjun and Zhao, Zhuokai},
  journal = {arXiv preprint arXiv:2606.01476},
  year    = {2026},
}

@article{starpo2026,
  title   = {{StaRPO}: Stability-Augmented Reinforcement Policy Optimization},
  author  = {Zhang, Jinghan and Mo, Fengran and Weerasooriya, Tharindu Cyril and Dai, Ruimin and
             Han, Xiaoyan and Fu, Yanjie and Wang, Dakuo and Liu, Kunpeng},
  journal = {arXiv preprint arXiv:2604.08905},
  year    = {2026},
}

\appendix
\section{Annotation Contract, Prompts, and State Replay}
\label{app:prompts}

\paragraph{Inputs and structural output.}
The annotator receives a domain system prompt (below), the task, and the
indexed trajectory: for each interactive turn the observation, the action,
and the environment's response to that action (the response is also the
next turn's observation; the 7B ALFWorld run and later runs attach it
explicitly to the acting turn as a \texttt{Result} field with the header
\emph{``A node becomes VERIFIED at the turn whose ACTION achieved it --
judge by that turn's Result, never by a later turn's State''}). A final
line states the episode outcome (success or failure) and, for Search and
mathematics, the reference answer. The annotator returns only structure: a \texttt{nodes}
list and one entry per step with \texttt{verified}, \texttt{errors}, and
\texttt{recovered} lists ($V_t$, $B_t$, $P_t$). It never emits a number.
A two-turn annotation for locating and picking up an object has the form
\begin{quote}
\small
\begin{verbatim}
{"nodes": ["locate", "pick", "place"],
 "turns": [
   {"turn_index": 1, "verified": ["locate"],
    "errors": [], "recovered": []},
   {"turn_index": 2, "verified": ["pick"],
    "errors": [], "recovered": []}
 ]}
\end{verbatim}
\end{quote}
The \texttt{place} node remains unestablished. Domain topologies determine
the edges: instance-suffixed chains in ALFWorld
(\texttt{locate:2}$\rightarrow$\texttt{pick:2}$\rightarrow$\texttt{place:2}),
\texttt{type}/\texttt{price} roots with \texttt{attr:*}/\texttt{opt:*}
children in WebShop, and \texttt{hop:1}$\rightarrow\cdots\rightarrow$
\texttt{answer} or \texttt{step:1}$\rightarrow\cdots\rightarrow$\texttt{answer}
chains in Search and mathematics.

\paragraph{State replay.}
Given the graph and events, the potential series is computed by the
following deterministic procedure, which implements
Eq.~(\ref{eq:state-transition}) and Eqs.~(\ref{eq:phi})--(\ref{eq:reward})
and is shared by all domains
(only the parent function differs); it is covered by unit tests for
independent branches, propagation along chains, recovery, persistence,
abandonment, and clipping:
\begin{quote}
\small
\begin{verbatim}
supported[c] = False; broken[c] = False   # for every node c
for t = 1..T:
    for c in B_t: broken[c] = True         # 1. invalidate
    for c in P_t: broken[c] = False        # 2. repair
    for c in nodes:                        # 3. verify (persistent)
        supported[c] = (supported[c] or c in V_t)
                       and not broken[c]
    Phi[t] = (1/M) * sum_{c: supported[c]} w(c)
    r[t] = sigma * clip(Phi[t] - Phi[t-1], -kappa, kappa)
\end{verbatim}
\end{quote}
The pseudocode is Eq.~(\ref{eq:state-transition}), with
\texttt{broken[}$c_i$\texttt{]} indicating $i\in\mathcal U(t)$ and
\texttt{supported[}$c_i$\texttt{]} indicating $S_i(t)=1$. On the persisted
annotation graphs, a $P$ report without $V$ on a broken predicate occurs
367 times in 120 of 2{,}327 ALFWorld graphs ($5.2\%$) and a $V$ report on
a predicate whose final break flag is set 351 times in 161 graphs
($6.9\%$); at least one of the two occurs in 264 ALFWorld graphs
($11.3\%$), 10 of 371 WebShop graphs ($2.7\%$), and 36 of 985
mathematics graphs ($3.7\%$), and a same-step $B$ and $P$ on one
predicate was never observed. A replay that instead set support on
either $V$ or $P$, giving those events precedence over $B$, never grants
less credit than the deployed rule; on the same graphs (with $\lambda=1$
and $\kappa=0.3$) it leaves the reward vector unchanged for $88.7\%$ of
ALFWorld, $97.6\%$ of WebShop, and $96.3\%$ of mathematics trajectories,
changes $1.0\%$, $0.4\%$, and $0.3\%$ of steps, and correlates with the
deployed per-step reward at Pearson $r=0.95$, $0.98$, and $0.99$. The
reported results are therefore insensitive to how a repair without a
same-step verification is resolved.
$w(c)$ is computed by a breadth-first search over ancestors that returns
$e^{-\lambda d}$ at the first (nearest) broken ancestor found at distance
$d$, and $1$ if none is found (Eq.~\ref{eq:weight}). The alternative
rule that applies the strongest attenuation over all broken ancestors
(the farthest one) is monotone under invalidation; the two rules differ
only when two broken ancestors lie on one path. On the persisted
annotation graphs available to us they give identical potentials on all
371 WebShop graphs (a depth-one topology), on $99.8\%$ of 2{,}327
ALFWorld graphs, and on $97\%$ of 985 mathematics graphs; among
3{,}849 ALFWorld steps and 1{,}744 mathematics steps that consist only of
invalidations, the deployed rule increased the potential once (the
strongest-attenuation rule never does). All reported runs use the
deployed rule. A node in both $P_t$ and $V_t$ is repaired and re-verified in
the same step (as in turn~4 of Table~\ref{tab:webshop-71074-recovery}); a
node in both $B_t$ and $V_t$ but not $P_t$ ends the step broken and
unsupported; a node in both $B_t$ and $P_t$ ends the step unbroken with
its previous support or a current $V$ report, so a previously broken node
becomes unestablished unless it is also in $V_t$.

\paragraph{Parsing and fallback.}
The parser normalizes node identifiers and domain ordering, drops event
references that are not in the node list, and aligns events with the
trajectory; omitted step entries receive empty event lists. An output with
no usable graph (malformed JSON, empty node list, or a multi-object
ALFWorld task without instance chains) is retried once with a corrective
message and otherwise rejected, in which case the trajectory keeps its
environment reward. In the 200-step 1.5B ALFWorld runs, transient API failures
left $5$--$12\%$ of trajectories with the environment reward; in the
reasoning-disabled configuration used for ALFWorld~7B and the
fixed-budget WebShop~7B runs (longer timeout) rejections were below
$10^{-4}$, and the 500-step 1.5B
runs (credit-aligned rendering, 16{,}000-token completion budget) had at
most one unusable annotation per run (Appendix~\ref{app:cost}).
Structural checks do not establish semantic correctness, which is
audited separately in Appendix~\ref{app:annot-quality}.

\paragraph{Segmentation.}
Interactive tasks use environment turns. Tool-free responses are packed
deterministically into eight contiguous paragraph-aligned blocks; credit
is assigned at block boundaries and broadcast over the tokens of the block.
The Python-interpreter mathematics track uses the ARPO turn structure
(reasoning, \texttt{<python>} call, \texttt{<result>}) as its steps.

\paragraph{Verbatim system prompts.}
The prompts below are the exact system prompts used in the reported runs.

\lstset{basicstyle=\ttfamily\scriptsize,breaklines=true,breakatwhitespace=true,columns=fullflexible,frame=single,framerule=0.3pt,xleftmargin=2pt,xrightmargin=2pt}

\noindent\textit{ALFWorld (v3, count- and identity-gated):}
\begin{lstlisting}
You analyze an ALFWorld household agent trajectory as ORDERED DEPENDENCY CHAINS of subgoal predicates. Vocabulary (base predicates): 'locate' (agent reached the target object's location), 'pick' (agent is HOLDING the correct target object), 'treat' (object correctly heated/cooled/cleaned as required, including the tool step: microwave/fridge/sinkbasin), 'place' (object put at the correct goal receptacle). STEP 1 - decompose the task into a chain per REQUIRED OBJECT. A plain 'put X in Y' object = [locate, pick, place]; a 'heat/cool/clean X and put in Y' object = [locate, pick, treat, place]. STEP 2 - COUNT: if the task says 'two', 'three', or 'both' (e.g. 'put two pillow in sofa'), emit ONE chain PER copy using instance suffixes: locate:1,pick:1,place:1 for the first object AND locate:2,pick:2,place:2 for the second, etc. A single-object task uses bare ids (no suffix). Each instance is an INDEPENDENT chain. STEP 3 - per turn, report chain state with these STRICT GATES: (a) IDENTITY GATE: only VERIFY locate/pick/treat if the object matches BOTH the task's object type AND its required property. Grabbing a MUG when the task says CUP, or an un-heated object when 'hot' is required, is NOT a verify -- mark 'pick' (or 'treat') in ERRORS. (b) COMPLETION GATE: only VERIFY 'place' when the object is confirmed AT THE CORRECT goal receptacle; a 'put' at any other receptacle is a 'place' ERROR, not a verify. (c) For multi-object tasks, attribute each verify/error to the correct instance (place:1 vs place:2). Removing an already-placed object re-marks that instance's 'place' (and 'pick') in ERRORS until re-satisfied. Mark RECOVERED at the turn a prior error on a node is repaired. Wandering/opening receptacles that establish no subgoal contributes nothing. Return ONLY JSON: {"nodes": [<all node ids across all instances, in order>], "turns": [{"turn_index": <1..N>, "verified": [..], "errors": [..], "recovered": [..]}]} with exactly N turn entries in order, using node ids drawn from the "nodes" list.
\end{lstlisting}

\noindent\textit{WebShop (v2, product-scoped requirements):}
\begin{lstlisting}
You analyze a WebShop agent trajectory as a DEPENDENCY GRAPH of requirement predicates. First decompose the INSTRUCTION into requirement nodes: exactly one 'type' node (the product category), zero+ 'attr:<name>' nodes (required attributes like waterproof, cotton), zero+ 'opt:<name>' nodes (selectable options like size, color), and a 'price' node if a price cap is given. Then for EACH turn report the graph state. A node is VERIFIED when the agent's actions/observations up to that turn establish it (opened a correct-type product => type; the product clearly has/plausibly has an attribute => that attr; the agent SELECTED the requested option value => that opt; within price cap => price). CRITICAL - option/attribute predicates are PRODUCT-SCOPED: an 'opt:' or 'attr:' node is only verified WHILE the agent remains on the product page where it was established, or after the agent clicks 'buy now' to lock it in. If the agent LEAVES that product WITHOUT buying it (clicks 'back to search', opens a DIFFERENT product, or the episode ends on a non-purchased product), the cart is ABANDONED: mark every opt:/attr: node that had been established on that product in ERRORS at the turn the agent leaves (those selections are lost). The 'type' and 'price' nodes may persist across navigation within the same product category. A purchase (buy now) permanently locks the predicates verified. Mark a node in ERRORS at the turn where an incorrect action CONTAMINATES it (opened a WRONG-type product => type error; selected a WRONG option value => that opt error; ABANDONED a ready product => its opt/attr errors as above). Mark a node in RECOVERED at the turn where a prior error on it is repaired (re-selected the option on a correct product; navigated to a correct-type product after a wrong one => type recovered). Aimless browsing that establishes no predicate contributes nothing (empty verified/errors/recovered). Return ONLY JSON: {"nodes": ["type", "attr:..", "opt:..", "price"], "turns": [{"turn_index": <1..N>, "verified": [..], "errors": [..], "recovered": [..]}]} with exactly N turn entries in order, using node ids drawn from the "nodes" list.
\end{lstlisting}

\noindent\textit{Search-R1 (hop chain):}
\begin{lstlisting}
You analyze a search-augmented QA agent trajectory as a DEPENDENCY CHAIN of reasoning hops. First decompose the QUESTION into the ordered sub-facts (hops) needed to answer it: a single-hop question has one hop; a multi-hop question has hop:1 (the bridge fact that must be found first), hop:2 (which builds on hop:1), and so on, in dependency order. ALWAYS include a final 'answer' node last. Then for EACH turn report the graph state. A hop is VERIFIED at the turn where the agent's retrieved observation actually contains that sub-fact. Mark a hop in ERRORS at the turn where a search CONTAMINATES it: the query drops or corrupts the entity/relation/time/constraint that hop hinges on, or retrieves evidence about the WRONG entity. Mark a hop in RECOVERED at the turn a prior error on it is repaired (a corrective search finally retrieves the right evidence). The 'answer' node is VERIFIED only at a turn that commits a final <answer> that is SUPPORTED by the retrieved evidence; put 'answer' in ERRORS at a turn that commits an <answer> the evidence does NOT support. A search that retrieves nothing on-topic, or aimless repetition, establishes no hop (empty verified/errors/recovered for that turn). Do NOT give a hop credit just because a search happened early. Return ONLY JSON: {"nodes": ["hop:1", "hop:2", ..., "answer"], "turns": [{"turn_index": <1..N>, "verified": [..], "errors": [..], "recovered": [..]}]} with exactly N turn entries in order, using node ids drawn from the "nodes" list.
\end{lstlisting}
The Search user message ends with a reference line: \emph{``REFERENCE: the
correct final answer is: \{gold\}. The outcome grader scored this trajectory
as \{SUCCEEDED/FAILED\} \ldots\ if the committed <answer> denotes a
DIFFERENT fact than the reference, put 'answer' in errors at the commit
turn and localize the FIRST contaminated hop; if it denotes the SAME fact
(case/punctuation/extra words do not matter), grade the chain on its merits
and do NOT invent errors.''}

\noindent\textit{Mathematics with a Python interpreter (ARPO/AEPO):}
\begin{lstlisting}
You analyze a math tool-integrated-reasoning agent trajectory as a DEPENDENCY CHAIN of reasoning steps. The agent solves a math problem by writing reasoning and, when useful, running Python code (a <python> block whose <result> it then sees) to compute or check intermediate quantities, before committing a final <answer> that contains \boxed{}. First decompose the SOLUTION into the ordered intermediate results (steps) the problem requires: step:1 is the first quantity or equation that must be established, step:2 builds on step:1, and so on in dependency order. ALWAYS include a final 'answer' node last. Then for EACH turn report the graph state. A step is VERIFIED at the turn where the agent correctly establishes that intermediate result -- either by sound reasoning or by a Python computation whose result actually computes the RIGHT quantity for that step (code that merely runs is NOT enough; it must compute the quantity the step needs). Mark a step in ERRORS at the turn where the agent CONTAMINATES it: an algebra or arithmetic mistake, a wrong formula or problem setup, misreading the question, or Python code that computes the wrong thing (a bug, a wrong expression, or the answer to a different sub-question). A later step that reuses a contaminated quantity is off-policy: its trust decays with dependency distance from the error. Mark a step in RECOVERED at the turn where a prior error on it is repaired (recomputed correctly, the bug fixed, or the approach corrected). The 'answer' node is VERIFIED only at a turn that commits a final <answer> whose boxed value is SUPPORTED by the correct chain; put 'answer' in ERRORS at a turn that commits a final <answer> the computations do NOT support (a wrong final value, or one contradicted by the interpreter). Reasoning or code that establishes no new intermediate result (restating, aimless exploration, a print with no bearing) contributes nothing (empty verified/errors/recovered for that turn). Do NOT give a step credit just because Python was called early. Return ONLY JSON: {"nodes": ["step:1", "step:2", ..., "answer"], "turns": [{"turn_index": <1..N>, "verified": [..], "errors": [..], "recovered": [..]}]} with exactly N turn entries in order, using node ids drawn from the "nodes" list.
\end{lstlisting}

\noindent\textit{Tool-free mathematics (segmented chain of thought):}
\begin{lstlisting}
You analyze a student model's chain-of-thought math solution as a DEPENDENCY CHAIN of reasoning steps. You are given the problem, a correct REFERENCE solution, and the student's solution split into numbered SEGMENTS in order.
First decompose what the PROBLEM requires into ordered intermediate results (steps), using the MINIMUM number of steps the problem truly needs: step:1 is the first quantity, equation or reduction that must be established, step:2 builds on step:1, and so on in dependency order. Derive these from the problem and the REFERENCE solution -- NOT from how the student happened to organize their writing. ALWAYS include a final 'answer' node last.
Then, for EACH segment, report the graph state. RULES:
(1) A step is VERIFIED at the segment where the student ACTUALLY establishes that intermediate result correctly (a correct derivation, algebraic manipulation, case analysis or computation that produces the right quantity). If in doubt, do NOT verify. Restating the problem, planning, aimless exploration, or checking without concluding establishes nothing.
(2) Mark a step in ERRORS at the segment where the student CONTAMINATES it: an algebra or arithmetic mistake, a wrong formula, a wrong problem setup, a misread constraint, an invalid case split, or an unjustified leap. A later step that reuses a contaminated quantity is off-policy; you do NOT need to re-mark it -- only mark where the error is INTRODUCED.
(3) Mark a step in RECOVERED at the segment where the student repairs a prior error on it (recomputes it correctly, catches the mistake, or corrects the approach). Self-correction is common in long chains-of-thought -- credit it when the repair actually lands.
(4) The 'answer' node goes in VERIFIED only at the segment that states a final answer matching the reference; it goes in ERRORS at the segment that states a final answer that does NOT match the reference. If the student never states a final answer, 'answer' must not appear anywhere.
Return ONLY JSON: {"nodes": ["step:1", "step:2", ..., "answer"], "segments": [{"segment_index": <1..N>, "verified": [..], "errors": [..], "recovered": [..]}]} with exactly N segment entries in order, using node ids drawn from the "nodes" list.
\end{lstlisting}

\section{Annotation Quality}
\label{app:annot-quality}

\paragraph{Offline replay bench.}
Table~\ref{tab:graphbench} replays 123 ALFWorld trajectories recorded
from an earlier training run through the production annotation path
without training.
Relative to a first prompt without count and identity gates, the deployed
prompt reduces the fraction of \emph{failed} trajectories whose graph
nonetheless saturates ($\Phi_G\ge0.99$) from $0.12$ to $0.02$, instantiates
instance chains on all 48 multi-object tasks, and leaves the mean
potential on successful trajectories nearly unchanged ($0.91\to0.89$).

\begin{table}[t]
\centering
\small
\caption{\textbf{Offline graph-induction bench.} Replay of $123$ recorded unseen-split ALFWorld
trajectories ($51$ failed, $72$ successful, including all $48$ multi-object tasks) through the
production message path, with no training involved. Gating reduces the fraction of \emph{failed}
trajectories whose induced graph nonetheless saturates; mean potential on successful trajectories
changes only slightly.}
\label{tab:graphbench}
\begin{tabular}{@{}p{0.66\linewidth}cc@{}}
\toprule
Measure & ungated & gated \\
\midrule
Failed trajectories with $\Phi_G \ge 0.99$ (over-credit) $\downarrow$ & 0.12 & \textbf{0.02} \\
Failed trajectories with $\Phi_G \ge 0.75$ $\downarrow$               & 0.14 & \textbf{0.04} \\
Count-aware graphs on multi-object tasks $\uparrow$                   & 0/48 & \textbf{48/48} \\
Mean $\Phi_G$ on successful trajectories (preserve)                   & 0.91 & 0.89 \\
Usable verdicts / median latency                                      & ---  & 122/123 \,/\, 23\,s \\
\bottomrule
\end{tabular}
\end{table}

\paragraph{Manual audit.}
Fifty-four production annotations from live 1.5B ALFWorld training were
sampled to exercise the known difficult cases (16 two-object tasks, 16
heat/cool/clean tasks, 12 plain failures, 6 parse failures, 4 successes)
and reviewed against the full trajectory by six independent reviewers,
blind to each other. Verdicts were $34$ correct ($63\%$), $12$ with a minor
error ($22\%$; typically a one-turn misalignment of a verification), and
$8$ with a major error ($15\%$; typically credit for sighting an object
without credit for the later pickup, or an over-credited failure). Across
roughly 15 identity traps (wrong object cleaned, heated, or placed), no
reviewer found positive credit for a wrong object, and a direct count over
60 multi-object trajectories found instance-suffixed chains in $46/46$
parsed graphs. The prompt's identity gate is applied to the object's
identity rather than to its not-yet-treated state: in the 128 annotated
heat/cool/clean trajectories of the production sample, \texttt{pick} is
verified at or before \texttt{treat} in $117$ of the $117$ trajectories
in which \texttt{treat} is verified.

\paragraph{Test--retest reliability.}
To measure the repeatability of the event labels on a random sample
rather than on selected hard cases, 1{,}000 held-out failed ALFWorld
trajectories were annotated twice by independent draws of the same
annotator (same-annotator agreement, not semantic accuracy). Agreement
between the two draws is Cohen's $\kappa=0.84$ for verification events
(2{,}211 and 2{,}132 events in the two draws; $83$--$86\%$ of one draw's
events are present in the other), $\kappa=0.68$ for invalidations
(2{,}402 and 2{,}185 events; $65$--$71\%$ overlap), and $\kappa=0.59$ for
repairs (379 and 314 events; $54$--$65\%$ overlap). Verification events
have the highest test--retest agreement; repairs are the rarest and
least repeatable event type.

\paragraph{Discrimination, density, and rendering.}
On a corpus of 128 trajectories from the 7B ALFWorld policy, the final
potential separates successful from failed episodes with
$\mathrm{AUC}=0.93$ and a mean gap of $0.40$, and false completions were
$0/9$ on failed episodes; because the annotator is told the outcome, this
is a consistency check on the induced graphs rather than independent
validation of intermediate events. Credit is sparse by design: in a
sample of 327 production ALFWorld~1.5B annotations, $22\%$ of turns carry
nonzero credit ($3.8$ credited turns per trajectory, $2.6\%$ of them
negative) and $4\%$ of trajectories carry none. Attaching the environment's response to the acting
turn matters for where credit lands: in a 200-transcript Search replay,
rendering each turn as State/Action/Result raised the share of hop
verifications assigned to the query that retrieved them from $0\%$ to
$54\%$ (with the observation/action rendering the verification is
typically recorded at the following turn, whose observation shows the
result), and on ALFWorld~7B it reduced credit recorded one turn after the
acting turn from $39.6\%$ to $0.7\%$ of credited events.

\paragraph{Distilled annotator.}
A Qwen3-8B model was fine-tuned on 23{,}498 API annotations of failed
trajectories (19{,}994 ALFWorld, 3{,}504 Search; held-out evaluation split
by task) to emit the same structural output. On held-out trajectories its
per-turn credit vector correlates with the API annotator's at $r=0.81$,
against an API-versus-API self-agreement ceiling of $0.86$. Trained end to
end on ALFWorld~1.5B in a run identical to its API-annotated twin in
policy, hyper-parameters, data, and prompt (differing only in which
endpoint answers the annotation call; 128 held-out games, eight draws),
it reaches peak success $92.2\%$ against $89.1\%$ for the API annotator, so
the local annotator loses nothing; we read the $+3.1$ points as within
decoding noise rather than as an advantage. This establishes practical sufficiency on ALFWorld; it does not show that
a stronger judge would add nothing.

\section{All Experiments at a Glance}
\label{app:all-experiments}

Table~\ref{tab:all-experiments} lists the principal trained arms,
variants, and controls behind the paper, with the checkpoint rule and
the number of decoding draws or evaluation cells behind each cell, so
that every number in the main text can be traced to one row.

\begingroup
\footnotesize
\setlength{\tabcolsep}{2.5pt}
\renewcommand{\arraystretch}{1.06}
\setlength{\extrarowheight}{0.4pt}
\setlength{\LTleft}{0pt}
\setlength{\LTright}{0pt}
\setlength{\LTcapwidth}{\textwidth}

\newlength{\allexpmetricwidth}
\settowidth{\allexpmetricwidth}{67.7 / 86.5 / 90.2 / 84.6 / 86.3}
\addtolength{\allexpmetricwidth}{2pt}

\newcommand{\AllExpCkpt}[1]{%
  \unskip\enspace\mbox{\textit{[#1]}}%
}
\newcommand{\AllExpMetric}[1]{\mbox{#1}}

\newcommand{\AllExpBlock}[2]{%
  \multicolumn{4}{p{\dimexpr\textwidth-2\tabcolsep\relax}}{%
    \textbf{#1}\newline
    #2%
  }\\*
  \addlinespace[2pt]
}
\newcommand{\AllExpSubblock}[1]{%
  \multicolumn{4}{p{\dimexpr\textwidth-2\tabcolsep\relax}}{%
    \textbf{#1}%
  }\\*
  \addlinespace[1pt]
}

\begin{longtable}{
  >{\raggedright\arraybackslash}
    p{\dimexpr\textwidth-\allexpmetricwidth-54pt-8\tabcolsep\relax}
  >{\centering\arraybackslash}p{27pt}
  >{\raggedright\arraybackslash}p{\allexpmetricwidth}
  >{\raggedright\arraybackslash}p{27pt}
}
\caption{\textbf{All experiments at a glance.} The principal trained
arms, variants, and controls reported in this paper, one row per (arm,
checkpoint) cell; detailed selection results appear in the cited tables.
``Draws'' is the number of independent decoding draws (interactive tasks
and agentic mathematics) or evaluation cells (tool-free reasoning) behind
the cell. Success / task score for ALFWorld and WebShop; exact-match QA
accuracy for Search-R1; mean AIME24/25 for agentic mathematics; the average
of AMC, AIME24, and AIME25 for tool-free reasoning, all in percent.
Checkpoints are shown in italic brackets in the first column.
DARS rows are shaded. Rows marked $^{\dagger}$ are published references
\citep{feng2025gigpo}.}
\label{tab:all-experiments}\\

\toprule
\textbf{Arm / configuration [checkpoint]} &
\textbf{Draws} &
\textbf{Metric} &
\textbf{Table} \\
\midrule
\endfirsthead

\toprule
\textbf{Arm / configuration [checkpoint]} &
\textbf{Draws} &
\textbf{Metric} &
\textbf{Table} \\
\midrule
\endhead

\midrule
\multicolumn{4}{r}{\emph{Continued on next page}}\\
\endfoot

\bottomrule
\endlastfoot

\AllExpBlock
  {ALFWorld --- Qwen2.5-1.5B-Instruct}
  {500 steps, 128 unseen games}

GiGPO (local) \AllExpCkpt{final 500}
  & 4 & 86.9 & \ref{tab:local-baselines} \\

GiGPO (local)
  \AllExpCkpt{milestones 100 / 150 / 200 / 300 / 400}
  & 4
  & \AllExpMetric{67.7 / 86.5 / 90.2 / 84.6 / 86.3}
  & \ref{tab:alfworld-milestones} \\

\rowcolor{oursbg}
\textbf{DARS} ($\lambda{=}1$, $\kappa{=}0.3$)
  \AllExpCkpt{final 500}
  & 4 & \textbf{96.9} & \ref{tab:main-gigpo} \\

\rowcolor{oursbg}
DARS \AllExpCkpt{milestones 100 / 150 / 200 / 300 / 400}
  & 4
  & \AllExpMetric{77.9 / 91.6 / 91.0 / 90.1 / 93.4}
  & \ref{tab:alfworld-milestones} \\

\midrule
\AllExpBlock
  {ALFWorld --- Qwen2.5-1.5B-Instruct}
  {200 steps, 128 unseen games}

GRPO (local) \AllExpCkpt{final 200}
  & 4 & 71.7 & \ref{tab:local-baselines} \\

GRPO (local) \AllExpCkpt{peak 170}
  & 4 & 70.9 & -- \\

RLOO (local) \AllExpCkpt{final 200}
  & 4 & 86.9 & \ref{tab:local-baselines} \\

RLOO (local) \AllExpCkpt{peak 150}
  & 4 & 75.8 & -- \\

PPO/GAE (local; collapses to invalid actions by step 200)
  \AllExpCkpt{peak 190}
  & 4 & 44.4 & -- \\

GiGPO (local) \AllExpCkpt{final 200}
  & 4 & 88.7 & \ref{tab:local-baselines} \\

\rowcolor{oursbg}
\textbf{DARS} ($\lambda{=}1$, $\kappa{=}0.3$)
  \AllExpCkpt{final 200}
  & 4 & \textbf{92.0} & \ref{tab:local-baselines} \\

\rowcolor{oursbg}
DARS $\lambda{=}0$ twin (same recipe)
  \AllExpCkpt{final 200}
  & 4 & 86.0 & \ref{tab:dependency-ablation} \\

\rowcolor{oursbg}
DARS $\lambda{=}0$ twin \AllExpCkpt{keep-best 180}
  & 4 & 87.9 & -- \\

GiGPO$^{\dagger}$ (w/ std) / GRPO$^{\dagger}$ / RLOO$^{\dagger}$
  \AllExpCkpt{150 it.}
  & --
  & \AllExpMetric{86.7 / 72.8 / 69.7}
  & \ref{tab:main-gigpo} \\

Sweep, distilled annotator, 150 steps:
  $\lambda{=}0$ / $0.5$ / $1$ / $\infty$
  \AllExpCkpt{peak}
  & 4
  & \AllExpMetric{90.2 / 94.3 / 91.4 / 84.2}
  & \ref{tab:lambda-sweep} \\

Annotator: API vs.\ distilled Qwen3-8B (matched pair)
  \AllExpCkpt{peak}
  & 8
  & \AllExpMetric{89.1 vs.\ 92.2}
  & \S\ref{app:annot-quality} \\

First-error prefix credit (graph-free), 150 steps
  \AllExpCkpt{peak}
  & --
  & \AllExpMetric{$+3.2$ vs.\ GiGPO}
  & \ref{tab:topology} \\

\midrule
\AllExpBlock
  {ALFWorld --- Qwen2.5-7B-Instruct}
  {500 steps, 128 unseen games}

GRPO (local) \AllExpCkpt{peak 460}
  & 16 & 87.4 & \ref{tab:alfworld7b-baselines} \\

GRPO (local) \AllExpCkpt{final 500}
  & 16 & 81.1 & \ref{tab:alfworld7b-baselines} \\

GRPO (local) \AllExpCkpt{step 280}
  & 8 & 83.4 & \ref{tab:alfworld7b-baselines} \\

GRPO (local) \AllExpCkpt{step 150}
  & 8 & 69.8 & \ref{tab:alfworld7b-baselines} \\

RLOO (local) \AllExpCkpt{peak 460}
  & 16 & 85.9 & \ref{tab:alfworld7b-baselines} \\

RLOO (local) \AllExpCkpt{final 500}
  & 16 & 87.6 & \ref{tab:alfworld7b-baselines} \\

RLOO (local) \AllExpCkpt{step 280}
  & 8 & 86.0 & \ref{tab:alfworld7b-baselines} \\

RLOO (local) \AllExpCkpt{step 150}
  & 8 & 79.2 & \ref{tab:alfworld7b-baselines} \\

GiGPO (local) \AllExpCkpt{train-val peak}
  & 16 & 98.1 & \ref{tab:local-baselines} \\

GiGPO (archived paired-test reference)
  \AllExpCkpt{peak 470}
  & 8 & 98.1 & \ref{tab:alfworld7b-baselines} \\

GiGPO (local) \AllExpCkpt{final 500}
  & 8 & 97.0 & \ref{tab:alfworld7b-baselines} \\

GiGPO (local) \AllExpCkpt{step 280}
  & 16 & 96.7 & \ref{tab:alfworld7b-baselines} \\

GiGPO (local) \AllExpCkpt{step 150}
  & 8 & 93.2 & \ref{tab:alfworld7b-baselines} \\

\rowcolor{oursbg}
\textbf{DARS} \AllExpCkpt{peak 445}
  & 16 & \textbf{98.6} & \ref{tab:main-gigpo} \\

\rowcolor{oursbg}
DARS $\lambda{=}0$ twin (from the shared step-300 checkpoint)
  \AllExpCkpt{peak 445}
  & 15 & 95.2 & \ref{tab:dependency-ablation} \\

\rowcolor{oursbg}
DARS $\lambda{=}0$ twin \AllExpCkpt{final 500}
  & 8 & 95.9 & -- \\

GiGPO$^{\dagger}$ (w/ std) \AllExpCkpt{150 it.}
  & -- & 90.8 & \ref{tab:main-gigpo} \\

\midrule
\AllExpBlock
  {WebShop --- Qwen2.5-1.5B-Instruct}
  {150 steps, 256 validation tasks (success / task score)}

GiGPO (local) \AllExpCkpt{final 150}
  & 4
  & \AllExpMetric{65.9 / 84.9}
  & \ref{tab:local-baselines} \\

GRPO (local) \AllExpCkpt{peak 145}
  & 8
  & \AllExpMetric{61.2 / 77.7}
  & \ref{tab:local-baselines} \\

GRPO (local) \AllExpCkpt{final 150}
  & 8
  & \AllExpMetric{61.9 / 76.6}
  & \ref{tab:local-baselines} \\

\rowcolor{oursbg}
DARS $\kappa{=}0.3$ \AllExpCkpt{final 150}
  & 4
  & \AllExpMetric{63.0 / 77.6}
  & \ref{tab:webshop-variants} \\

\rowcolor{oursbg}
\textbf{DARS} $\kappa{=}1$ \AllExpCkpt{final 150}
  & 4
  & \AllExpMetric{67.7 / \textbf{85.8}}
  & \ref{tab:webshop-variants} \\

\rowcolor{oursbg}
\textbf{DARS} $\kappa{=}1$ $+$ completion verifier
  \AllExpCkpt{final 150}
  & 4
  & \AllExpMetric{\textbf{69.6} / 85.6}
  & \ref{tab:webshop-variants} \\

Topology study: requirement graph ($\kappa{=}0.3$)
  \AllExpCkpt{final 150}
  & 4
  & \AllExpMetric{63.2 / 79.5}
  & \ref{tab:dependency-ablation} \\

Topology study: sequential chain ($\kappa{=}0.3$)
  \AllExpCkpt{final 150}
  & 4
  & \AllExpMetric{49.0 / 61.1}
  & \ref{tab:dependency-ablation} \\

First-error prefix credit (graph-free)
  & --
  & \AllExpMetric{$-42$ task score vs.\ GiGPO}
  & \ref{tab:topology} \\

GiGPO$^{\dagger}$ (w/o std) / (w/ std)
  \AllExpCkpt{150 it.}
  & --
  & \AllExpMetric{67.4 / 83.5; 65.0 / 83.1}
  & \ref{tab:main-gigpo} \\

\midrule
\AllExpBlock
  {WebShop continuation}
  {From our best saved stock-GiGPO checkpoint
   (Appendix~\ref{app:webshop-cont});
   evaluation seeds, not training seeds}

\AllExpSubblock{WebShop 1.5B cont.}

GiGPO parent (400 steps), fresh seeds 12--23
  \AllExpCkpt{400}
  & 12
  & \AllExpMetric{81.5 / 92.4}
  & \ref{tab:webshop-cont} \\

\rowcolor{oursbg}
\textbf{DARS continuation} (training seed 1; 800 total steps),
  fresh seeds 12--23 \AllExpCkpt{cont.\ step 400}
  & 12
  & \AllExpMetric{\textbf{83.1} / \textbf{93.4}}
  & \ref{tab:main-gigpo}, \ref{tab:webshop-cont} \\

\rowcolor{oursbg}
DARS continuation, training seed 0 / mixing weight 0.10,
  selection seeds 0--11 \AllExpCkpt{cont.\ step 400}
  & 12 / 11
  & \AllExpMetric{82.1 / 93.0; 81.6 / 93.1}
  & \ref{tab:webshop-cont} \\

\addlinespace[3pt]
\AllExpSubblock{WebShop 7B cont.}

GiGPO parent (600 steps), fresh seeds 12--23
  \AllExpCkpt{600}
  & 12
  & \AllExpMetric{84.3 / 93.1}
  & \ref{tab:webshop-cont} \\

\rowcolor{oursbg}
\textbf{DARS continuation}
  (strict reward, training seed 0; 850 total steps),
  fresh seeds 12--23 \AllExpCkpt{cont.\ step 250}
  & 12
  & \AllExpMetric{\textbf{88.5} / \textbf{94.6}}
  & \ref{tab:main-gigpo}, \ref{tab:webshop-cont} \\

\rowcolor{oursbg}
DARS continuation, strict reward, training seed 1,
  selection seeds 0--11 \AllExpCkpt{cont.\ step 250}
  & 12
  & \AllExpMetric{87.4 / 94.6}
  & \ref{tab:webshop-cont} \\

\midrule
\AllExpBlock
  {WebShop --- Qwen2.5-7B-Instruct}
  {250 steps, 256 validation tasks (success / task score)}

GiGPO (local) \AllExpCkpt{peak 245 (= final)}
  & 12
  & \AllExpMetric{77.3 / 85.7}
  & \ref{tab:local-baselines} \\

\rowcolor{oursbg}
\textbf{DARS} \AllExpCkpt{peak 245}
  & 12
  & \AllExpMetric{76.2 / \textbf{87.9}}
  & \ref{tab:local-baselines} \\

\rowcolor{oursbg}
DARS \AllExpCkpt{final 250}
  & 12
  & \AllExpMetric{74.8 / 84.2}
  & -- \\

GiGPO$^{\dagger}$ (w/o std) / GRPO$^{\dagger}$
  \AllExpCkpt{150 it.}
  & --
  & \AllExpMetric{75.2 / 86.2; 66.1 / 79.3}
  & \ref{tab:main-gigpo} \\

\midrule
\AllExpBlock
  {Search-R1 shared pool --- Qwen2.5-7B-Instruct}
  {Exact-match QA accuracy}

GiGPO (local) \AllExpCkpt{final 662}
  & $3{\times}4$ & 38.7 & \ref{tab:main-gigpo} \\

\rowcolor{oursbg}
\textbf{DARS} (hop chain $+$ commitment term)
  \AllExpCkpt{final 360}
  & $3{\times}4$ & \textbf{43.1} & \ref{tab:main-gigpo} \\

GiGPO (local) \AllExpCkpt{train-val peak}
  & $3{\times}4$ & 42.4 & \S\ref{app:search-detail} \\

\rowcolor{oursbg}
DARS \AllExpCkpt{train-val peak}
  & $3{\times}4$ & 44.7 & \S\ref{app:search-detail} \\

First-error prefix credit (graph-free)
  & -- & 0.0 success & \ref{tab:topology} \\

\midrule
\AllExpBlock
  {Search-R1 held-out --- Qwen2.5-7B-Instruct}
  {Exact-match QA accuracy}

GiGPO (local) \AllExpCkpt{\texttt{\upshape val350} peak 600}
  & $3{\times}4$ & 39.5 & \ref{tab:search-heldout} \\

\rowcolor{oursbg}
\textbf{DARS} ($\sigma{=}1$)
  \AllExpCkpt{\texttt{\upshape val350} peak 490}
  & $3{\times}4$ & \textbf{41.6} & \ref{tab:search-heldout} \\

\rowcolor{oursbg}
DARS ($\sigma{=}0.5$)
  \AllExpCkpt{\texttt{\upshape val350} peak 410}
  & $3{\times}4$ & 39.7 & \ref{tab:search-heldout} \\

Commitment terms only \AllExpCkpt{\texttt{\upshape val350} peak 100}
  & $3{\times}4$ & 24.6 & \ref{tab:search-heldout} \\

Graph credit only \AllExpCkpt{\texttt{\upshape val350} peak 80}
  & $3{\times}4$ & 36.9 & \ref{tab:search-heldout} \\

\midrule
\AllExpBlock
  {Agentic mathematics (Python)}
  {Qwen3-8B with a Python interpreter (mean AIME24/25)}

ARPO \AllExpCkpt{steps 5 / 10 / 15}
  & 4
  & \AllExpMetric{61.7 / 61.3 / 62.9}
  & \ref{tab:main-aepo} \\

ARPO \AllExpCkpt{13 milestones to step 78}
  & 4
  & \AllExpMetric{peak 62.9 @15, final 35.0}
  & \S\ref{app:sanity} \\

AEPO \AllExpCkpt{steps 5 / 10 / 15}
  & 4
  & \AllExpMetric{63.3 / 61.7 / 60.0}
  & \ref{tab:main-aepo} \\

\rowcolor{oursbg}
ARPO $+$ DARS \AllExpCkpt{steps 5 / 10 / 15}
  & 4
  & \AllExpMetric{59.2 / 61.3 / 64.2}
  & \ref{tab:main-aepo} \\

\rowcolor{oursbg}
\textbf{AEPO $+$ DARS} \AllExpCkpt{steps 5 / 10 / 15}
  & 4
  & \AllExpMetric{60.4 / \textbf{67.5} / 62.9}
  & \ref{tab:main-aepo} \\

Untrained policy
  & 4 & 57.9 & \ref{tab:main-aepo} \\

\midrule
\AllExpBlock
  {Tool-free reasoning --- Qwen3-1.7B chat}
  {60 steps on persisted rollouts
   (average of AMC, AIME24, AIME25; points)}

\rowcolor{oursbg}
\textbf{DARS} ($\eta{=}1$, $\lambda{=}1$)
  \AllExpCkpt{mean over cells}
  & 10 & \textbf{58.23} & \ref{tab:main-math} \\

\rowcolor{oursbg}
DARS $\lambda{=}0$ (same graphs)
  & 14 & 57.50 & \ref{tab:reasoning-arms} \\

\rowcolor{oursbg}
DARS $\eta{=}2$ / wider admission / broken-node penalty /
  answer-commit gates
  & 3 each
  & \AllExpMetric{57.27 / 57.14 / 57.10 / 57.08}
  & \ref{tab:reasoning-arms} \\

\rowcolor{oursbg}
DARS w/o step credit ($\eta{=}0$, same graphs)
  & 10 & 56.90 & \ref{tab:main-math} \\

Group-consensus decomposition variant
  & 10 & 57.00 & \S\ref{app:reasoning} \\

StaRPO
  & 3 & 57.43 & \ref{tab:reasoning-arms} \\

On-policy distillation (vanilla) / OmniOPD
  & 4 / 9
  & \AllExpMetric{57.32 / 57.24}
  & \ref{tab:main-math} \\

step-GRPO / FWTA-GRPO (front-weighted scalars)
  & 4 / 3
  & \AllExpMetric{57.40 / 56.81}
  & \ref{tab:reasoning-arms} \\

GRPO (task reward only) / brevity control / untrained model
  & 3 each
  & \AllExpMetric{57.69 / 55.83 / 57.48}
  & \ref{tab:reasoning-arms} \\

\midrule
\AllExpBlock
  {Tool-free reasoning --- Qwen3-4B chat}
  {200 steps (average of AMC, AIME24, AIME25; points)}

\rowcolor{oursbg}
\textbf{DARS} ($\eta{=}1$) \AllExpCkpt{mean over cells}
  & 20 & \textbf{79.50} & \ref{tab:main-math} \\

\rowcolor{oursbg}
DARS w/o step credit ($\eta{=}0$)
  & 13 & 79.13 & \ref{tab:main-math} \\

StaRPO
  & 4 & 78.94 & \ref{tab:main-math} \\

On-policy distillation (vanilla) / OmniOPD
  & 15 / 12
  & \AllExpMetric{78.62 / 78.44}
  & \ref{tab:main-math} \\

Rejection fine-tuning / untrained model
  & 4 / 14
  & \AllExpMetric{78.79 / 79.17}
  & \ref{tab:reasoning-4b} \\

\end{longtable}
\endgroup

\section{Budget-Matched Local Baselines and Learning Dynamics}
\label{app:local-baselines}

The published ALFWorld and WebShop references train for 150 iterations,
whereas the fixed-budget experiments below use 150--500 steps. To
compare methods at matched training budgets, we trained GiGPO ourselves
at every ALFWorld and WebShop scale, GRPO and RLOO on ALFWorld at both
scales, and GRPO on WebShop~1.5B, with the DARS launch script, hyper-parameters
(learning rate, KL coefficient, $\gamma$, group size, batch size, turn
limit), and policy-training budget of the DARS arm in the same block, and
evaluated all arms on one checksum-verified evaluation file with the same
decoding protocol (Table~\ref{tab:local-baselines}). The same PPO/GAE
script collapsed to invalid actions late in training and is omitted (its
peak checkpoint scores $44.4\%$ on ALFWorld~1.5B). DARS is ahead of the
matched GiGPO on ALFWorld at both scales and at both 1.5B budgets, and on
the WebShop graded score at both scales; on WebShop~7B strict success the
difference between the two fixed-budget arms is not statistically
significant ($-1.0$ points, $p=0.07$), so at that scale and budget the benefit
is in the graded score. ALFWorld~7B GRPO and RLOO remain below GiGPO
at their training-validation peaks, at the final step, and at matched
intermediate steps (Table~\ref{tab:alfworld7b-baselines}). For
WebShop~1.5B GRPO, we report both the training-validation peak and the
final checkpoint.

\paragraph{Matched references for Table~\ref{tab:main-gigpo}.}
Table~\ref{tab:main-matched} pairs each DARS row of
Table~\ref{tab:main-gigpo} with a GiGPO policy we trained with the same
script and evaluation harness. For ALFWorld, the reference is GiGPO trained
with the same policy-training budget. For the WebShop continuations, it
is the GiGPO parent the continuation starts from, scored on identical
evaluation draws, so these margins include the effect of additional
training, which the equally extended GiGPO control of
Appendix~\ref{app:webshop-control} separates. For Search-R1,
it is our GiGPO reproduction. Table~\ref{tab:main-gigpo} reports these
budget-controlled margins as $\Delta$; Table~\ref{tab:main-matched} adds
their paired tests and, in its last column, the margins over the stronger
published GiGPO variant, which additionally span different training budgets.

\begin{table}[h]
\centering
\small
\caption{\textbf{Matched references for the DARS rows of
Table~\ref{tab:main-gigpo}.} Success (success / task score for WebShop).
$\Delta$ against the reference is paired over identical evaluation draws
where available, with $p$ and draws won as in
Tables~\ref{tab:local-baselines} and~\ref{tab:webshop-cont}; the last
column repeats the margin over the stronger published GiGPO variant.
$^{\ddagger}$DARS continuation from the listed parent.}
\label{tab:main-matched}
\setlength{\tabcolsep}{4pt}
\resizebox{\textwidth}{!}{
\begin{tabular}{lllll}
\toprule
Setting & DARS & Local GiGPO reference & $\Delta$ vs.\ reference & $\Delta$ vs.\ published \\
\midrule
ALFWorld 1.5B & $96.9$ & $86.9$ (same 500-step budget) & $+10.0$ ($p=0.001$; 4/4) & $+10.2$ \\
ALFWorld 7B & $98.6$ & $98.1$ (same 500-step budget) & $+0.5$ ($p=0.014$; 9W/2L/5T) & $+7.8$ \\
WebShop 1.5B$^{\ddagger}$ & $83.1$ / $93.4$ & $81.5$ / $92.4$ (parent, 400 steps) & $+1.5$ / $+1.1$ ($p=0.001$; 11/12) & $+15.7$ \\
WebShop 7B$^{\ddagger}$ & $88.5$ / $94.6$ & $84.3$ / $93.1$ (parent, 600 steps) & $+4.2$ / $+1.5$ ($p=0.0005$; 12/12) & $+13.3$ \\
Search-R1 shared & $43.1$ & $38.7$ (662 steps; DARS 360) & $+4.4$ (3/3 groups) & -- \\
Search-R1 held-out & $41.6$ & $39.5$ (one epoch each) & $+2.1$ (3/3 groups) & -- \\
\bottomrule
\end{tabular}
}
\end{table}

\begin{table}[h]
\centering
\caption{
\textbf{Budget-matched local baselines.} GRPO, RLOO, and GiGPO trained
by us with the same script, hyper-parameters, policy-training budget,
and evaluation harness as the DARS arm of the same block, and evaluated
on the same checksum-verified evaluation file; only the advantage
estimator or the content of the step channel differs. ALFWorld~1.5B has
two blocks: the 500-step runs behind Table~\ref{tab:main-gigpo} (GiGPO
and DARS launched together on one host) and the earlier 200-step runs
that include GRPO and RLOO; ALFWorld~7B trains every arm for 500 steps,
WebShop for 150 steps at 1.5B and 250 at 7B. ALFWorld~7B rows use
training-validation peaks; final and intermediate checkpoints appear in
Table~\ref{tab:alfworld7b-baselines}. $\pm$ is the standard error
over independent decoding draws (4 at ALFWorld~1.5B; 16 at ALFWorld~7B;
12 paired at WebShop~7B; WebShop~1.5B cells are means of 4 draws, 8 for
GRPO). $\Delta$ is the difference from local GiGPO (success, or
success~/~task score), computed as the paired mean difference on shared
environment seeds where draws are paired (all 8 shared seeds for
ALFWorld~7B GRPO and RLOO) and as the difference of means otherwise; $p$
is an exact sign-flip test over paired draws where at least 8 exist,
otherwise a Welch $t$-test over draws. WebShop~1.5B GRPO was evaluated on
a different host from its references, so no $\Delta$ or paired test is
reported for those rows. The
ALFWorld~1.5B 500-step and ALFWorld~7B DARS rows repeat
Table~\ref{tab:main-gigpo}; the WebShop DARS rows are the fixed-budget
runs trained from the instruction-tuned policy, not the continuation
policies of Table~\ref{tab:main-gigpo}.
Scores are percentages and $\Delta$ is in percentage points; DARS rows are shaded.
}
\label{tab:local-baselines}
\footnotesize
\setlength{\tabcolsep}{4pt}
\renewcommand{\arraystretch}{0.95}
\resizebox{\textwidth}{!}{%
\begin{tabular}{llccl}
\toprule
Environment & Method & Success $\uparrow$ & Task score $\uparrow$ & $\Delta$ ($p$) \\
\midrule
\multirow{2}{*}{\shortstack[l]{ALFWorld\\1.5B, 500 steps}}
 & GiGPO (local) & $86.9 \pm 0.8$ & -- & -- \\
 & \ours \textbf{DARS (ours)} & \ours $\mathbf{96.9 \pm 0.0}$ & \ours -- & \ours $\mathbf{+10.0}$ ($0.001$; 4/4 draws) \\
\midrule
\multirow{4}{*}{\shortstack[l]{ALFWorld\\1.5B, 200 steps}}
 & GRPO (local) & $71.7 \pm 0.6$ & -- & $-17.0$ \\
 & RLOO (local) & $86.9 \pm 0.6$ & -- & $-1.8$ \\
 & GiGPO (local) & $88.7 \pm 0.8$ & -- & -- \\
 & \ours \textbf{DARS (ours)} & \ours $\mathbf{92.0 \pm 0.2}$ & \ours -- & \ours $\mathbf{+3.3}$ ($0.02$; 4/4 draws) \\
\midrule
\multirow{4}{*}{\shortstack[l]{ALFWorld\\7B, 500 steps}}
 & GRPO (local) & $87.4 \pm 0.4$ & -- & $-10.3$ ($0.0078$; 0/8) \\
 & RLOO (local) & $85.9 \pm 0.4$ & -- & $-12.2$ ($0.0078$; 0/8) \\
 & GiGPO (local) & $98.1 \pm 0.1$ & -- & -- \\
 & \ours \textbf{DARS (ours)} & \ours $\mathbf{98.6 \pm 0.2}$ & \ours -- & \ours $\mathbf{+0.5}$ ($0.014$; 9W/2L/5T) \\
\midrule
\multirow{5}{*}{\shortstack[l]{WebShop\\1.5B, 150 steps}}
 & GRPO (local, peak 145) & $61.2 \pm 1.0$ & $77.7 \pm 0.6$ & -- \\
 & GRPO (local, final 150) & $61.9 \pm 1.0$ & $76.6 \pm 0.8$ & -- \\
 & GiGPO (local) & $65.9$ & $84.9$ & -- \\
 & \ours \textbf{DARS (ours)} & \ours $67.7$ & \ours $\mathbf{85.8}$ & \ours $+1.8$~/~$+0.9$ \\
 & \ours \textbf{DARS $+$ completion verifier} & \ours $\mathbf{69.6}$ & \ours $85.6$ & \ours $\mathbf{+3.7}$~/~$+0.7$ \\
\midrule
\multirow{2}{*}{\shortstack[l]{WebShop\\7B, 250 steps}}
 & GiGPO (local) & $\mathbf{77.3 \pm 0.5}$ & $85.7 \pm 0.3$ & -- \\
 & \ours \textbf{DARS (ours)} & \ours $76.2 \pm 0.5$ & \ours $\mathbf{87.9 \pm 0.3}$ & \ours $-1.0$ ($0.07$, n.s.)~/~$\mathbf{+2.1}$ ($0.001$; 11/12) \\
\bottomrule
\end{tabular}%
}
\end{table}

\paragraph{ALFWorld 1.5B over 500 steps.}
The 500-step GiGPO and DARS arms of Table~\ref{tab:local-baselines}
were launched together with one script, one training seed, and one
host, and every milestone checkpoint was evaluated on the same 128
unseen games with the same four evaluation seeds
(Table~\ref{tab:alfworld-milestones}). DARS is ahead at every
milestone. Among the milestones, GiGPO is highest at step~200 ($90.2\%$)
and scores $84.6$--$86.9\%$ thereafter, whereas DARS improves again
after step~300 and reaches its highest score at step~500, where all four
draws score $96.9\%$ (124 of 128 games) and are perfect on
\texttt{pick\_and\_place}, \texttt{look}, \texttt{cool}, and two-object
tasks. The 500-step block of Table~\ref{tab:local-baselines} reports
the final checkpoints fixed in advance, and the draw-wise comparisons
use the same four environment seeds at every milestone.

\begin{table}[h]
\centering
\small
\caption{\textbf{ALFWorld 1.5B milestone curve of the 500-step runs}
(unseen-game success, mean $\pm$ SE over the same four evaluation seeds
per cell; the step-500 DARS entry is repeated in
Table~\ref{tab:main-gigpo}).}
\label{tab:alfworld-milestones}
\setlength{\tabcolsep}{5pt}
\resizebox{\textwidth}{!}{%
\begin{tabular}{lcccccc}
\toprule
Step & 100 & 150 & 200 & 300 & 400 & \textbf{500 (final)} \\
\midrule
GiGPO (local) & $67.7 \pm 1.9$ & $86.5 \pm 1.0$ & $90.2 \pm 0.2$ & $84.6 \pm 0.2$ & $86.3 \pm 0.4$ & $86.9 \pm 0.8$ \\
\rowcolor{oursbg}\textbf{DARS} & $\mathbf{77.9 \pm 1.6}$ & $\mathbf{91.6 \pm 0.6}$ & $\mathbf{91.0 \pm 0.4}$ & $\mathbf{90.1 \pm 0.7}$ & $\mathbf{93.4 \pm 0.4}$ & $\mathbf{96.9 \pm 0.0}$ \\
$\Delta$ (draws won) & $+10.2$ (4/4) & $+5.1$ (4/4) & $+0.8$ (3/4) & $+5.5$ (4/4) & $+7.1$ (4/4) & $\mathbf{+10.0}$ (4/4) \\
\bottomrule
\end{tabular}%
}
\end{table}

\paragraph{ALFWorld 1.5B per-class results (200-step runs).}
Table~\ref{tab:perclass} breaks the step-200 cells down by task class
(mean of the same four decoding draws). DARS gains most on
\texttt{pick\_and\_place} ($+15.7$ points) and \texttt{clean} ($+10.6$) and is
perfect on two-object tasks, whose graphs carry two independent chains;
GiGPO remains stronger on \texttt{look}, \texttt{heat}, and \texttt{cool}.

\begin{table}[t]
\centering
\small
\caption{\textbf{ALFWorld per-class success (unseen split, 200-step runs of Table~\ref{tab:local-baselines}, step-200 checkpoint, four draws).}
Classes: Pick $=$ \texttt{pick\_and\_place}, Look $=$ \texttt{look\_at\_obj\_in\_light},
Clean / Heat / Cool $=$ \texttt{pick\_X\_then\_place}, Pick2 $=$ \texttt{pick\_two\_obj\_and\_place}.
Success in percent. In these runs DARS improves overall success from $88.7\%$ to $92.0\%$ ($+3.3$ points), with the largest gains
on Pick ($+15.7$) and Clean ($+10.6$), and reaches perfect success on Pick2.
GiGPO remains stronger on Look, Heat, and Cool.}
\label{tab:perclass}
\resizebox{\textwidth}{!}{%
\begin{tabular}{lccccccc}
\toprule
Method & Pick & Look & Clean & Heat & Cool & Pick2 & All \\
\midrule
GiGPO
& 72.6
& \textbf{100.0}
& 82.6
& \textbf{100.0}
& \textbf{93.5}
& 97.5
& 88.7 \\
\rowcolor{oursbg}\quad + DARS
& \textbf{88.3}
& 91.9
& \textbf{93.2}
& 90.3
& 92.4
& \textbf{100.0}
& \textbf{92.0} \\
\bottomrule
\end{tabular}
}
\end{table}

\begin{figure}[h]
\centering
\includegraphics[width=0.98\textwidth]{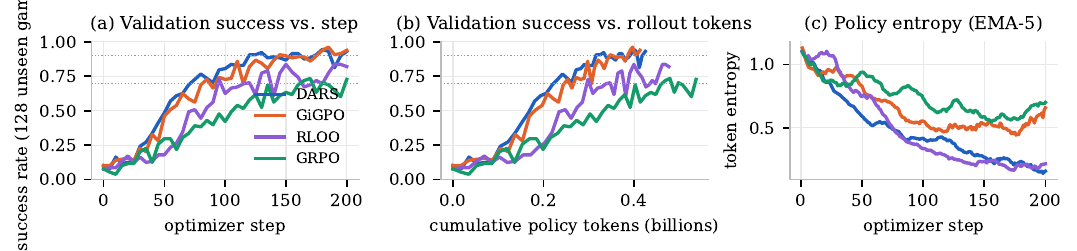}
\caption{\textbf{ALFWorld 1.5B learning dynamics} of the 200-step runs in
Table~\ref{tab:local-baselines} (validation on the 128 unseen games every five
steps, one draw at temperature $0.4$). DARS reaches each success level
earlier in (a) optimizer steps and (b) rollout tokens, and (c) ends
with lower token entropy.}
\label{fig:dynamics}
\end{figure}

\paragraph{Learning dynamics.}
\label{sec:exp-dynamics}
Figure~\ref{fig:dynamics} compares the validation curves of the 200-step
runs on ALFWorld~1.5B.
DARS first reaches $70\%$ validation success after $75$ optimizer steps
versus $85$ for GiGPO, $95$ for RLOO, and $165$ for GRPO, and $90\%$ after
$120$ steps versus $145$ for GiGPO (RLOO and GRPO never reach $90\%$); the
DARS$-$GiGPO gap over steps 65--130, where most rollouts still fail,
averages $+6.7$ points (DARS higher at 12 of 14 validation points), and the
normalized area under the curve over steps 0--180 is $0.643$ versus
$0.605$, $0.479$, and $0.409$. In rollout tokens, DARS needs $0.23$B
versus $0.25$B to reach $70\%$ and $0.32$B versus $0.35$B to reach $90\%$,
and it ends with lower token entropy ($0.24$ vs.\ $0.48$) at similar
reference-KL and gradient norm. This is policy-sample efficiency;
the wall-clock cost of annotation depends on the annotation endpoint
(Appendix~\ref{app:cost}).

\paragraph{ALFWorld 7B: algorithm baselines over 500 steps.}
GRPO and RLOO each completed 500 training steps with the GiGPO script
and hyper-parameters, changing only the advantage estimator. All
checkpoints in Table~\ref{tab:alfworld7b-baselines} were evaluated on
the same host, pinned harness, and checksum-verified file as the GiGPO
and DARS references: 128 unseen games, temperature $0.4$, top-$p$ $1$,
and top-$k$ disabled. Each arm has one training seed and no separate
hyper-parameter search. Both baselines select step~460 on training
validation; their selected peaks and step-500 finals have 16 decoding
draws each, and the step-150 and step-280 cells have eight.

GiGPO leads both baselines under every tested selection rule and matched
step. Across the eight comparisons, the paired gaps are $9.7$--$23.4$ points,
with GiGPO ahead on all eight shared evaluation seeds in each comparison
(two-sided exact sign-flip $p=0.0078$ each). These paired differences use
the shared seeds rather than differences between the full-cell means.
The GRPO--RLOO ordering depends on selection: at step~500, RLOO leads
GRPO by $6.5\pm0.5$ points (paired SE; 16/16 draws, $p<0.0001$), while at
their training-validation peaks RLOO trails by $1.5\pm0.6$ points
($p=0.021$; three wins and four ties in 16 draws). Thus a
training-validation peak need not be the strongest checkpoint on the
reported unseen set. The reference evaluations preceded the new
baseline evaluations by about two weeks; a contemporaneous drift
re-evaluation was not run.

\begin{table}[htbp]
\centering
\caption{\textbf{ALFWorld 7B baselines across checkpoint selections.}
Success is mean $\pm$ SE over the listed decoding draws on 128 unseen
games. ``Peak'' is selected on training validation, not on these
evaluation draws. $\Delta$ is baseline minus GiGPO, mean $\pm$ SE over
eight shared evaluation seeds; W counts baseline wins. All eight
baseline--GiGPO tests have two-sided exact sign-flip $p=0.0078$.
GiGPO rows use the archived reference pools for this comparison; the
peak row here has eight draws, while Table~\ref{tab:local-baselines}
reports its fuller 16-draw summary. Paired differences need not equal
differences of the displayed full-cell means. Success and differences are in percent.
}
\label{tab:alfworld7b-baselines}
\small
\setlength{\tabcolsep}{4pt}
\begin{tabular}{llccccc}
\toprule
Selection & Method & Step & Draws & Success $\uparrow$ & $\Delta$ & W/8 \\
\midrule
\multirow{3}{*}{Peak}
 & GiGPO & 470 & 8 & $98.1\pm 0.1$ & -- & -- \\
 & GRPO & 460 & 16 & $87.4\pm 0.4$ & $-10.3\pm 0.4$ & 0/8 \\
 & RLOO & 460 & 16 & $85.9\pm 0.4$ & $-12.2\pm 0.5$ & 0/8 \\
\midrule
\multirow{3}{*}{Final}
 & GiGPO & 500 & 8 & $97.0\pm 0.2$ & -- & -- \\
 & GRPO & 500 & 16 & $81.1\pm 0.4$ & $-15.8\pm 0.6$ & 0/8 \\
 & RLOO & 500 & 16 & $87.6\pm 0.4$ & $-9.7\pm 0.6$ & 0/8 \\
\midrule
\multirow{3}{*}{Matched step}
 & GiGPO & 280 & 16 & $96.7\pm 0.2$ & -- & -- \\
 & GRPO & 280 & 8 & $83.4\pm 1.0$ & $-13.5\pm 1.1$ & 0/8 \\
 & RLOO & 280 & 8 & $86.0\pm 0.8$ & $-11.0\pm 0.8$ & 0/8 \\
\midrule
\multirow{3}{*}{Matched step}
 & GiGPO & 150 & 8 & $93.2\pm 0.5$ & -- & -- \\
 & GRPO & 150 & 8 & $69.8\pm 1.0$ & $-23.4\pm 1.3$ & 0/8 \\
 & RLOO & 150 & 8 & $79.2\pm 0.8$ & $-14.0\pm 1.0$ & 0/8 \\
\bottomrule
\end{tabular}
\end{table}

\paragraph{WebShop 1.5B: GRPO over 150 steps.}
The GRPO baseline completed 150 steps and was evaluated on 256 tasks
with eight decoding draws per checkpoint. Its training-validation peak
at step~145 scores $61.2\pm1.0\%$ success and $77.7\pm0.6\%$ task score;
the final step~150 scores $61.9\pm1.0\%$ and $76.6\pm0.8\%$
(mean $\pm$ SE; Table~\ref{tab:local-baselines}). The two selections
favor different metrics, so both are reported. GRPO uses the same
checksum-verified evaluation file and decoding settings as the local
GiGPO and fixed-budget DARS references, but was evaluated on a different
host. We therefore report its scores without a paired comparison to
those references.

\section{WebShop: DARS Continuation from a Trained GiGPO Policy}
\label{app:webshop-cont}

The fixed-budget WebShop rows of Table~\ref{tab:local-baselines} train
DARS from the instruction-tuned policy for 150 (1.5B) or 250 (7B) steps.
The WebShop rows of Table~\ref{tab:main-gigpo} answer a different
question: does continuing our \emph{best saved} stock-GiGPO checkpoint
with the DARS reward raise it further? The comparison is therefore
between a continued policy and the policy it started from, scored on
identical evaluation draws. Because any additional training can raise
the parent, Appendix~\ref{app:webshop-control} compares each DARS
continuation with an equally extended GiGPO continuation of the same
parent.

\textbf{Setup.} All continuation arms of a given size start from the
same saved GiGPO parent (1.5B: 400 GiGPO steps from
Qwen2.5-1.5B-Instruct; 7B: 600 steps from Qwen2.5-7B-Instruct) with a
fresh optimizer and the parent as KL reference, and use stock GiGPO
grouping, 16 tasks $\times$ 8 rollouts of at most 15 turns, learning
rate $5\times10^{-7}$, KL coefficient $0.01$, discount $0.95$, and an
invalid-action penalty of $1.0$ (the shipped $0.1$ is scale-relative and
fails under a graded terminal reward). Unlike the single
completed-trajectory call of the fixed-budget experiments, the annotator here
is a local Qwen3-32B model that reads trajectory prefixes turn by turn
(goal schema, re-verification of recovered predicates, product scope 2)
with the WebShop predicate vocabulary of Appendix~\ref{app:prompts};
the one-call-per-trajectory cost figures in the main text refer to the
fixed-budget experiments. The confirmed 1.5B checkpoint uses a graded
terminal reward ($10\times$ task score) and the confirmed 7B checkpoint
a strict terminal reward ($10$ on exact success); both mix the graph
credit into the environment return with weight $0.25$, settling the
discounted cost at the last active turn so that the discounted
per-episode sum of shaped rewards equals the environment return.
Evaluation uses the stock code path with every treatment knob disabled.
The 1.5B search covered graded and strict terminal rewards, mixing
weight $0.10$, an additive potential-based variant, and two training
seeds (six arms, 400 continuation steps each); the 7B search used the
same variants for 250 steps (seven arms).

\textbf{Evaluation and selection.} Every checkpoint and its parent are
scored on the same harness, checksum-matched 256-task validation file,
temperature $0.4$, and environment seeds; differences are paired by seed
and tested with an exact two-sided sign-flip test. We screened every arm's
latest, keep-best, and final checkpoints at 1.5B, and keep-best and
final checkpoints at 7B, with two paired draws, advanced the highest
final checkpoints (three at 1.5B, two at 7B)
to 12 selection draws (environment seeds 0--11; the 1.5B
mixing-weight-$0.10$ candidate and the 7B locked candidate completed
11), locked the one with the highest mean success (ties broken by task
score), and then evaluated only the locked checkpoint and its parent on
12 \emph{unused} seeds (12--23) from the same validation pool. All four
confirmation $p$-values (two sizes $\times$ two metrics; $0.001$,
$0.001$, $0.0005$, $0.0005$) lie below even the conservative four-claim
Bonferroni threshold ($0.05/4=0.0125$), and all four claims hold under
Holm's step-down procedure.

\begin{table}[h]
\centering
\scriptsize
\caption{\textbf{WebShop continuation from our best saved stock-GiGPO
checkpoint.} Mean $\pm$ standard deviation over decoding draws;
$\Delta$ is the paired difference on identical environment seeds,
computed before rounding (draws won / total; raw two-sided sign-flip
$p$, leading zero omitted). The confirmation rows are the headline and the main-table cells:
fresh evaluation seeds on the same validation task pool. At 1.5B, success
wins on confirmation are 11 wins and one tie and task-score wins are 11
wins and one loss; at 7B every confirmation draw favors DARS on both
metrics. Scores are percentages; $\Delta$ in points; DARS rows are shaded.}
\label{tab:webshop-cont}
\setlength{\tabcolsep}{3pt}
\resizebox{\textwidth}{!}{%
\begin{tabular}{@{}llccll@{}}
\toprule
Size & Checkpoint & Success & Task score & $\Delta$ success (won; $p$) & $\Delta$ task score (won; $p$) \\
\midrule
\multicolumn{6}{@{}l}{\emph{1.5B, selection draws (seeds 0--11)}} \\
1.5B & GiGPO parent (400 steps) & $80.4 \pm 2.4$ & $91.7 \pm 1.0$ & -- & -- \\
 \rowcolor{oursbg}& DARS, graded, seed 1, step 400 (locked) & $82.1 \pm 2.2$ & $93.0 \pm 1.0$ & $+1.7$ (11/12; .001) & $+1.3$ (12/12; .0005) \\
 \rowcolor{oursbg}& DARS, graded, seed 0, step 400 & $82.1 \pm 2.2$ & $93.0 \pm 1.2$ & $+1.7$ (9/12; .004) & $+1.3$ (10/12; .003) \\
 \rowcolor{oursbg}& DARS, graded, mixing weight $0.10$, step 400 ($n{=}11$) & $81.6 \pm 2.4$ & $93.1 \pm 1.1$ & $+1.3$ (10/11; .002) & $+1.4$ (11/11; .001) \\
\midrule
\multicolumn{6}{@{}l}{\emph{1.5B, confirmation draws (fresh seeds 12--23)}} \\
1.5B & GiGPO parent (400 steps) & $81.5 \pm 2.9$ & $92.4 \pm 1.0$ & -- & -- \\
 \rowcolor{oursbg}& \textbf{DARS, locked (800 total steps)} & $\mathbf{83.1 \pm 2.4}$ & $\mathbf{93.4 \pm 0.9}$ & $\mathbf{+1.5}$ (11/12; .001) & $\mathbf{+1.1}$ (11/12; .001) \\
\midrule
\multicolumn{6}{@{}l}{\emph{7B, selection draws (seeds 0--11)}} \\
7B & GiGPO parent (600 steps) & $84.0 \pm 2.5$ & $92.5 \pm 1.3$ & -- & -- \\
 \rowcolor{oursbg}& DARS, strict, seed 0, step 250 (locked; $n{=}11$) & $88.1 \pm 2.1$ & $94.7 \pm 0.7$ & $+4.3$ (11/11; .001) & $+2.3$ (11/11; .001) \\
 \rowcolor{oursbg}& DARS, strict, seed 1, step 250 & $87.4 \pm 2.0$ & $94.6$ & $+3.5$ (12/12; .0005) & $+2.1$ (11/12; .001) \\
\midrule
\multicolumn{6}{@{}l}{\emph{7B, confirmation draws (fresh seeds 12--23)}} \\
7B & GiGPO parent (600 steps) & $84.3 \pm 2.5$ & $93.1 \pm 1.1$ & -- & -- \\
 \rowcolor{oursbg}& \textbf{DARS, locked (850 total steps)} & $\mathbf{88.5 \pm 2.4}$ & $\mathbf{94.6 \pm 1.4}$ & $\mathbf{+4.2}$ (12/12; .0005) & $\mathbf{+1.5}$ (12/12; .0005) \\
\bottomrule
\end{tabular}%
}
\end{table}

\textbf{Results.} On the fresh confirmation seeds the locked 1.5B
checkpoint scores $83.1\%$ / $93.4\%$ against its parent's $81.5\%$ /
$92.4\%$, and the locked 7B checkpoint $88.5\%$ / $94.6\%$ against its
parent's $84.3\%$ / $93.1\%$ (Table~\ref{tab:webshop-cont}); the paired
gains are similar on the selection and confirmation seed blocks
($+1.7$ versus $+1.5$ points of success at 1.5B, $+4.3$ versus $+4.2$ at
7B). At 7B every one of the 12
confirmation draws favors DARS on both metrics, with per-draw success
gains between $+2.7$ and $+5.1$ points. In the two-draw screens, all six
1.5B final checkpoints scored above the parent on both metrics ($+0.8$
to $+2.0$ points of success, $+0.7$ to $+1.7$ of task score), and the three
graded-reward candidates have similar selection means (within $0.6$
points of success and $0.1$ of task score). At 7B the two strict-reward arms were
the candidates advanced to selection. Training-validation keep-best
copies were not better than the final checkpoints on this harness at
either size, so single-draw training-validation peaks are used only as
selection signals.

\subsection{Equally extended GiGPO control}
\label{app:webshop-control}
To separate the effect of the DARS reward from the effect of additional
training, we continued each parent for the same number of steps with the
identical continuation recipe and the annotator disabled (stock GiGPO:
400 steps at 1.5B, 250 at 7B; two provider settings $\times$ two training
seeds per size). The design was registered before any control checkpoint
was scored. The control checkpoint was locked by the same rule as the
DARS checkpoint (highest mean success on evaluation seeds 0--11 among the
final and keep-best checkpoints of all control runs, ties broken by task
score, without consulting any DARS result), and the locked DARS and
control checkpoints and their parent were then scored on fresh evaluation
seeds 12--35 ($n=24$), paired by seed, with exact two-sided sign-flip
tests and Holm's correction over the four DARS-versus-control claims
(two sizes $\times$ two metrics).

\begin{table}[h]
\centering
\small
\caption{\textbf{WebShop continuation versus an equally extended GiGPO
continuation of the same parent} (fresh evaluation seeds 12--35, $n=24$,
paired by seed; draws won / tied / lost; exact sign-flip $p$). The 7B
success contrast passes the Holm correction over the four
DARS-versus-control claims. Differences are in percentage points.}
\label{tab:webshop-control}
\setlength{\tabcolsep}{4pt}
\resizebox{\textwidth}{!}{
\begin{tabular}{llll}
\toprule
Size (added steps) & GiGPO continuation $-$ parent & DARS continuation $-$ parent & DARS $-$ GiGPO continuation \\
 & success / task score & success / task score & success; task score \\
\midrule
7B (+250) & $+2.8$ / $+2.3$ & $+4.2$ / $+2.1$ & $\mathbf{+1.5}$ (20/2/2; $p<10^{-5}$); $-0.2$ ($p=0.16$) \\
1.5B (+400) & $+2.0$ / $+1.0$ & $+1.7$ / $+1.2$ & $-0.3$ ($p=0.12$); $+0.3$ ($p=0.13$) \\
\bottomrule
\end{tabular}
}
\end{table}

At 7B, on these draws the parent scores $84.0\%$ / $92.7\%$, the GiGPO
continuation $86.8\%$ / $95.0\%$, and the DARS continuation $88.3\%$ /
$94.7\%$: additional GiGPO training accounts for about two-thirds of the
DARS continuation's success gain over the parent, and DARS adds
$1.5$ points of success beyond it. At 1.5B, additional GiGPO training accounts
for the DARS continuation's gain over the parent.

\subsection{From scratch at the published 150-step budget}
\label{app:webshop-scratch}
To compare with GiGPO at its own training budget without a continuation
confound, we trained DARS and GiGPO from Qwen2.5-1.5B-Instruct with an
identical recipe, host, and environment seed, differing only in the
reward: strict terminal reward ($10$ on exact success), stock GiGPO
grouping, invalid-action penalty $1.0$, learning rate $10^{-6}$, KL
coefficient $0.01$, $\gamma=0.95$, and 16 tasks $\times$ 8 rollouts of at
most 15 turns. The DARS arm uses an additive variant of the step reward,
$r_t=r^{\mathrm{env}}_t+\alpha\,\tilde r_t$ with $\alpha=1$, so graph
progress enters the discounted return of the action that made it; the
prefix-reading Qwen3-32B annotator of the continuation runs supplies the
events ($\lambda=1$, $\kappa=0.3$). The GiGPO twin is the same run with the
annotator disabled. We selected $\alpha$ among $\{0.5,1,2.5,5\}$ on
evaluation seeds 0--11 ($\alpha=5$ did not learn and was stopped; every
scored $\alpha\in\{0.5,1,2.5\}$ leads the twin in mean success at every
scored step from 50 to 150), and confirmed the selected step-50 and
step-100 pairs on fresh evaluation seeds 12--23. Both arms are scored on
the matched harness of Appendix~\ref{app:webshop-cont} (256 tasks,
temperature $0.4$), paired by evaluation seed, with exact two-sided
sign-flip tests. At 150 steps the GiGPO twin scores $65.6\%$ / $83.7\%$,
within the published band ($67.4\pm4.5$ / $83.5\pm1.8$), and DARS
scores $70.6\%$ / $86.1\%$. The advantage is one of speed: it is largest at
step 100 and has closed by step 200, where the twin catches up.
The comparison uses one training seed.

\begin{table}[h]
\centering
\small
\caption{\textbf{WebShop 1.5B from scratch: DARS (additive, $\alpha=1$)
versus an identically configured GiGPO twin at matched steps} (training
seed 0). Mean $\pm$ standard error over evaluation seeds 0--11 (12 draws);
$^{a}$evaluation seeds 0--7 (8 draws); $^{b}$fresh evaluation seeds
12--23 (12 draws). $\Delta$ is the difference paired by evaluation seed, with
its standard error, draws won / lost, and the exact two-sided sign-flip
$p$. The standard errors reflect evaluation variability, not
training-seed variability. Scores are percentages; $\Delta$ in points.}
\label{tab:webshop-scratch}
\setlength{\tabcolsep}{3pt}
\resizebox{\textwidth}{!}{
\begin{tabular}{lcccccc}
\toprule
 & \multicolumn{3}{c}{Success} & \multicolumn{3}{c}{Task score} \\
\cmidrule(lr){2-4}\cmidrule(lr){5-7}
Step & DARS & GiGPO twin & $\Delta$ (W/L; $p$) & DARS & GiGPO twin & $\Delta$ (W/L; $p$) \\
\midrule
50 & $36.8\pm 1.1$ & $34.0\pm 0.9$ & $+2.8\pm 1.1$ (7/4; $0.042$) & $73.6\pm 0.4$ & $68.5\pm 0.4$ & $+5.1\pm 0.5$ (12/0; $0.0005$) \\
100 & $61.3\pm 0.9$ & $55.7\pm 0.9$ & $+5.6\pm 0.8$ (12/0; $0.0005$) & $82.3\pm 0.3$ & $79.2\pm 0.4$ & $+3.1\pm 0.5$ (12/0; $0.0005$) \\
150 & $\mathbf{70.6\pm 0.8}$ & $65.6\pm 0.8$ & $\mathbf{+5.0\pm 0.8}$ (11/0; $0.001$) & $\mathbf{86.1\pm 0.5}$ & $83.7\pm 0.3$ & $\mathbf{+2.4\pm 0.4}$ (11/1; $0.001$) \\
200$^{a}$ & $74.5\pm 0.9$ & $74.3\pm 0.9$ & $+0.1\pm 0.7$ (3/4; $0.89$) & $87.7\pm 0.3$ & $88.1\pm 0.6$ & $-0.4\pm 0.4$ (3/5; $0.40$) \\
\midrule
50$^{b}$ & $37.0\pm 1.1$ & $33.8\pm 0.9$ & $+3.2\pm 1.0$ (9/3; $0.009$) & $73.8\pm 0.8$ & $67.6\pm 0.6$ & $+6.2\pm 0.7$ (12/0; $0.0005$) \\
100$^{b}$ & $62.2\pm 1.1$ & $55.1\pm 1.2$ & $+7.1\pm 0.8$ (12/0; $0.0005$) & $82.5\pm 0.6$ & $79.4\pm 0.6$ & $+3.1\pm 0.4$ (12/0; $0.0005$) \\
\bottomrule
\end{tabular}
}
\end{table}

\section{Training Protocol and Hyper-parameters}
\label{app:protocol}

\paragraph{Domain topologies.}
ALFWorld uses one ordered chain
$\texttt{locate}\rightarrow\texttt{pick}\rightarrow\texttt{treat}\rightarrow\texttt{place}$
per required object instance (\texttt{treat} only for heat/cool/clean
tasks; instances carry suffixes \texttt{:1}, \texttt{:2}). WebShop uses a
parallel-AND graph whose roots are \texttt{type} and \texttt{price} and
whose children are the instruction's \texttt{attr:*} and \texttt{opt:*}
requirements. Search-R1 uses a chain of required retrieval hops ending in
\texttt{answer}; mathematics uses a chain of required derivation steps
ending in \texttt{answer}. The verbatim prompts below fix the vocabulary
and the connection rule for each family.

\paragraph{Interactive tasks and optimization.}
ALFWorld uses a $50$-action limit and reports success on the unseen
validation games; WebShop uses a $15$-action limit and reports strict
success and graded task score on 256 validation tasks; Search-R1 uses a
$4$-turn limit with a Wikipedia (wiki-18, e5) retriever over seven QA
subsets (NQ, TriviaQA, PopQA, HotpotQA, 2WikiMultihopQA, MuSiQue,
Bamboogle; 100 questions each). Group-relative training uses $16$ tasks
with $8$ rollouts per step (5 on Search-R1), $\gamma=0.95$, $\omega=1$
with GiGPO's standard-deviation normalization, learning rate $10^{-6}$, a
low-variance KL penalty with coefficient $0.01$ ($0.001$ on Search), and
an invalid-action penalty of $0.1$ ($0.01$ on Search), which is the only
non-terminal environment reward. In the ALFWorld~1.5B runs successful
trajectories keep their terminal reward and are not annotated (their
potential is saturated by definition); the 7B and WebShop runs annotate
every trajectory. The GRPO, RLOO, and PPO baselines use the
same script with only the advantage estimator changed (PPO adds a critic
at learning rate $10^{-5}$, $\lambda=0.95$; it collapsed to invalid
actions late in training and is omitted from Table~\ref{tab:main-gigpo};
its peak checkpoint scores $44.4\%$). Budgets: ALFWorld 500 steps at
1.5B (Table~\ref{tab:main-gigpo}; the earlier 200-step runs and the
150-step comparison are reported in Appendix~\ref{app:local-baselines})
and 500 at 7B, WebShop 150 steps at 1.5B and 250 at 7B for the
fixed-budget runs (the continuation runs of Table~\ref{tab:main-gigpo}
add 400 and 250 steps to trained GiGPO policies,
Appendix~\ref{app:webshop-cont}), Search-R1 one epoch (662 steps) for the
held-out protocol. The 7B policies are fully
fine-tuned. The ALFWorld~7B DARS run and its $\lambda=0$ twin share their first
300 steps (a common DARS checkpoint) and were trained 200 further steps
each with the State/Action/Result rendering; the GiGPO~7B baseline was
trained for the same 500 steps. All budgets are policy-training budgets
(optimizer steps and rollouts); annotation is additional
(Appendix~\ref{app:cost}).

\paragraph{Published references.}
The ALFWorld and WebShop rows marked $^{\dagger}$ in
Table~\ref{tab:main-gigpo} are from
Table~1 of \citet{feng2025gigpo}, with the source's uncertainty
estimates; margins over the better standard GiGPO variant at the same
scale are listed in Table~\ref{tab:main-matched}. The published models train for 150
iterations; Appendix~\ref{app:local-baselines} provides separate
fixed-budget comparisons.

\paragraph{Interactive evaluation.}
Our rows in Table~\ref{tab:main-gigpo} report means $\pm$ standard errors
over decoding draws (over decoding-group means for Search-R1);
published rows retain the source's uncertainty estimates.
All evaluations use top-$p$ $1.0$ and temperature $0.4$ ($1.0$ on
Search-R1), with independent environment seeds as decoding draws: four at ALFWorld~1.5B and in the
fixed-budget WebShop~1.5B runs (eight for each GRPO checkpoint), 16
(environment seeds 0--15) for the headline ALFWorld~7B cells, and
8--16 for the supplementary ALFWorld~7B checkpoints
(Table~\ref{tab:alfworld7b-baselines}; eight shared seeds for each GRPO
and RLOO comparison with GiGPO), 12 in the fixed-budget WebShop~7B runs and in each WebShop
confirmation cell (screening and selection counts in
Appendix~\ref{app:webshop-cont}), and three decoding groups (distinct sampler
seeds) of four draws each on Search-R1. Within a fixed evaluation configuration, the harness is deterministic
for a given checkpoint and environment seed, so same-seed pairing is
exact and permutation tests flip the sign of paired differences; with
fewer than 8 paired draws (ALFWorld~1.5B) we report the number of draws
on which DARS is ahead and a Welch $t$-test over draws. These tests are
conditional on the trained checkpoints and quantify evaluation
variability, not training-run variability.
The headline ALFWorld~1.5B and fixed-budget WebShop~1.5B rows read the final
checkpoint; ALFWorld~7B and the fixed-budget WebShop~7B runs read each
arm's keep-best checkpoint under training validation (single-draw
success on the seen split), and we verified that WebShop~7B's GiGPO
baseline scores identically at its peak (step 245) and its final step
(250). Appendix~\ref{app:local-baselines} additionally reports the
ALFWorld~7B baselines at final and matched intermediate steps and
WebShop~1.5B GRPO at its training-validation peak. The WebShop continuation runs follow the screening, selection,
and fresh-seed confirmation protocol of Appendix~\ref{app:webshop-cont}
(12 draws per confirmation cell).

\paragraph{Search-R1 configurations.}
The shared-pool diagnostic (Table~\ref{tab:main-gigpo}) uses the hop-chain
reward with an answer-commitment term read off the same graph: $+0.3$ on
the first step at which \texttt{answer} enters $V_t$ and $-0.5$ on the
final step if it never does. The held-out protocol trains all arms for
one epoch and selects on \texttt{val350}; its DARS arm uses commitment
weights $+0.5$ / $-0.8$ and a $-0.3$ term on a committed unsupported
answer, and a second DARS arm with the dense-channel scale halved
($\sigma=0.5$) scores $39.7\pm0.2\%$ on \texttt{test350}
(Appendix~\ref{app:search-detail}). The commitment terms address a
failure mode specific to this task: under the graph reward alone a
policy that keeps searching can retain most of the potential without
ever answering, as the graph-only arm of Table~\ref{tab:search-heldout}
does from step~330.

\paragraph{Agentic mathematics.}
The four arms of Table~\ref{tab:main-aepo} share one launch script and
differ in two flags: ARPO is the entropy-branched rollout with dual-clip
PPO on group-relative advantages and the outcome reward; AEPO adds
entropy-balanced clipping and the entropy-aware advantage. ARPO $+$ DARS
and AEPO $+$ DARS add the step reward of Eq.~(\ref{eq:reward}) on the last token of
every turn (the dynamic entropy-balanced rollout is off in every arm, as
in the authors' reasoning recipe). Both DARS arms use the same LLM
annotator to produce graph events; the annotator is part of DARS, while
LLM-equality grading is shared by all arms at evaluation.
All arms use batch $128$, mini-batch
$16$, $16$ rollouts, $8$ initial rollouts, beam size $2$, branch
probability $0.5$, learning rate $10^{-6}$, no KL term, and the same data
order. Evaluation: temperature $0.6$, top-$p$ $0.95$, top-$k$ $20$,
repetition penalty $1.1$, $4096$ response tokens, four draws, LLM-equality
grading with an audit that no reported cell contains a decoding error or
truncation. The seed-0 comparison uses steps 5, 10, and 15, the common
retained checkpoints across all four arms; the ARPO arm was additionally
evaluated at 13 milestones up to its final step 78. The checkpoint
comparison is detailed in Appendix~\ref{app:sanity}.

\paragraph{Tool-free reasoning (ablation setting).}
Qwen3-1.7B (chat) samples four rollouts per DAPO-Math prompt once; the
775 usable annotation graphs over these persisted rollouts define 142
groups (551 rollouts) that every arm trains on for 60 steps with learning
rate $10^{-6}$, group size $4$, a reference-KL coefficient of $0.1$, and
sequence length $8192$. This is a fixed-rollout (off-policy after the
first step) design chosen so that all arms see identical data and
identical annotations. Evaluation uses the chat template, temperature
$1.0$, top-$p$ $1.0$, eight draws per problem, a $32{,}768$-token budget,
and symbolic-equivalence grading; an evaluation cell is a (checkpoint,
host, repeat) triple and an arm's score is the mean over its cells.

\paragraph{Reward configuration per track.}
Table~\ref{tab:reward-config} summarizes the default reward
configurations of the fixed-budget comparisons (ablation settings are
specified with each ablation); the WebShop continuation runs of
Table~\ref{tab:main-gigpo} use the configuration of
Appendix~\ref{app:webshop-cont}. All tracks use persistent support, explicit
invalidation, and repair followed by re-verification
(Section~\ref{sec:method-reward}), with $\lambda=1$ and $\sigma=1$. The
clip bound is $\kappa=0.3$ except in the fixed-budget WebShop~1.5B
configuration, which uses $\kappa=1$ with and without the completion
verifier (sensitivity study below).

\begin{table}[h]
\centering
\small
\caption{\textbf{Reward configurations behind the main DARS results.} Topology is the fixed domain
rule; node instantiation is performed jointly with event annotation for each rollout. The WebShop
continuation study uses its own configuration (Appendix~\ref{app:webshop-cont}).}
\label{tab:reward-config}
\resizebox{\textwidth}{!}{%
\begin{tabular}{@{}llllccl@{}}
\toprule
Track & Topology & Node instantiation (per rollout) & $\kappa$ & $\lambda$ & $\sigma$ & Optional terms \\
\midrule
ALFWorld 1.5B/7B   & Object chains     & objects, treatment      & $0.3$ & $1.0$ & $1.0$ & none \\
WebShop 1.5B (fixed budget) & Parallel-AND & attributes, options & $0.3$ / $1.0$ & $1.0$ & $1.0$ & optional verifier \\
WebShop 7B (fixed budget) & Parallel-AND & attributes, options & $0.3$ & $1.0$ & $1.0$ & none \\
Search-R1 7B       & Hop chain         & required hops           & $0.3$ & $1.0$ & $1.0$ & answer-commitment term \\
Math with Python   & Derivation chain  & required steps          & $0.3$ & $1.0$ & $1.0$ & none \\
Math tool-free     & Derivation chain  & required steps          & $0.3$ & $1.0$ & $1.0$ & none \\
\bottomrule
\end{tabular}%
}
\end{table}

\paragraph{WebShop 1.5B clip-bound sensitivity.}
Table~\ref{tab:webshop-variants} compares reward configurations at
WebShop~1.5B on the same optimizer, rollout budget, and step-150
checkpoint (four draws), together with the local GiGPO baseline. Because
these variants were compared on the same 256-task validation set that
they are reported on, the fixed-budget WebShop~1.5B cells ($\kappa=1$,
with and without the verifier; $67.7\%$ / $85.8\%$ and $69.6\%$ / $85.6\%$)
are a configuration comparison rather than a confirmatory evaluation of
a frozen recipe. The
clip bound
matters: $\kappa=0.3$ truncates the large potential changes that
WebShop's 8-node graphs produce when a product is opened or abandoned and
scores $4.7$ points lower success than $\kappa=1$, which is exactly
telescoping. An optional completion verifier makes a second annotation
call on failed trajectories whose graph is near-saturated
($\Phi_G(T)\ge0.75$), names the unsatisfied requirement, and injects it
as a last-step invalidation; it raises success by a further $1.9$ points at
unchanged task score. On ALFWorld the verifier fires on $0.1\%$ of failed
trajectories and has no measurable effect, so it is not part of the main
recipe. A local GRPO baseline trained with the same script at this scale
scores $61.9\%$ success / $76.6\%$ task score at its final checkpoint and
$61.2\%$ / $77.7\%$ at its training-validation peak (eight draws each;
Table~\ref{tab:local-baselines}). These GRPO cells were evaluated on a
different host from the variant rows, so we do not report paired
differences against them.

\begin{table}[h]
\centering
\small
\caption{\textbf{WebShop 1.5B clip-bound sensitivity.} Same optimizer, rollout budget and
step-$150$ checkpoint; four decoding draws at temperature $0.4$. Scores in percent; DARS rows are shaded.}
\label{tab:webshop-variants}
\begin{tabular}{lcc}
\toprule
Variant & Success $\uparrow$ & Task score $\uparrow$ \\
\midrule
GiGPO (local)                            & $65.9$ & $84.9$ \\
\rowcolor{oursbg}DARS, $\kappa=0.3$                       & $63.0$ & $77.6$ \\
\rowcolor{oursbg}DARS, $\kappa=1$                         & $67.7$ & $\mathbf{85.8}$ \\
\rowcolor{oursbg}DARS, $\kappa=1$ $+$ completion verifier & $\mathbf{69.6}$ & $85.6$ \\
\bottomrule
\end{tabular}
\end{table}

\paragraph{Clipping.}
The clip bound $\kappa=0.3$ truncates only steps that change the
potential by more than $0.3$ (events touching several predicates at
once, or a break whose attenuation reaches several supported
descendants) and exists to keep the step channel's scale comparable to
the terminal reward under GiGPO's group normalization. Clipping breaks
the exact telescoping identity asymmetrically: a large loss is truncated
to $-\kappa$ while its later recovery is paid in several smaller steps,
so a break, repair, and re-verification of a root predicate in a
three-node chain nets $+0.3$ under $\kappa=0.3$ although the graph
returns to its original state. We therefore do not claim invariance for
the clipped reward. Two empirical observations bound the concern: negative-credit
turns are rare in the recorded training corpora ($0.4\%$ of ALFWorld~7B
turns and $3.3$--$5\%$ of WebShop~7B turns), so a policy would have to
provoke repeated annotated breaks to farm credit, which we did not
observe (under nearest-ancestor weighting a repair of a nearer break can
also expose a farther one and lower the potential, which is likewise
rare); and the unclipped $\kappa=1$
configuration, which is exactly telescoping, performs best in the
WebShop~1.5B sensitivity study. On the persisted annotation graphs, the deployed reward (nearest
broken ancestor, $\kappa=0.3$) and a monotone, unclipped variant
(strongest attenuation, $\kappa=1$) agree per turn at Pearson
$r=0.91$ on WebShop, $0.99$ on ALFWorld, and $0.98$ on mathematics, with
identical reward vectors on $64$--$94\%$ of trajectories and clipping
binding on $0.4$--$5.2\%$ of turns; the residual differences are the
multi-predicate events described above. Because the two ancestor
rules give identical potentials on every WebShop requirement graph (a
depth-one topology) and on $99.8\%$ of ALFWorld graphs, the WebShop
requirement-graph results are, at the same clip bound, also results for
the monotone strongest-attenuation rule, and the ALFWorld runs differ
from it on $0.2\%$ of graphs; the unclipped $\kappa=1$ rule was trained
online at WebShop~1.5B (Table~\ref{tab:webshop-variants}). A variant that carries the
truncated residual forward (so that the clipped sum still equals the
final potential once the episode is long enough) is implemented and
unit-tested but was not used for any reported run.

\paragraph{Clipping excess over training.}
Whether optimization learns to exploit the clip asymmetry can be measured
directly on the training stream, since the graph is persisted for every
on-policy trajectory. For each trajectory we compare the sum of the
deployed clipped rewards ($\kappa=0.3$) with the unclipped telescoping
sum (which equals the final potential) and count break--repair cycles on
a predicate. Over three on-policy ALFWorld~1.5B DARS training runs (150
steps $\times$ 128 trajectories each, 57{,}600 trajectories), a cycle
occurs in $1.2$--$1.5\%$ of trajectories ($2.9$--$3.3\%$ of failed
ones), positive clipping excess amounts to $1.1$--$1.4\%$ of all
positive credit, and the mean excess is negative ($-0.003$ to $-0.004$;
clipping removes more credit than it adds, with excess above $+0.05$ on
$0.8$--$1.0\%$ of trajectories and below $-0.05$ on $2.9$--$3.3\%$).
Neither quantity trends upward over training: by 25-step block, the
share of trajectories with a cycle is $0.6$--$0.7\%$ in the first block,
peaks at $1.6$--$2.3\%$ mid-run, and is $1.0$--$1.6\%$ in the last, and
the positive-excess share of positive credit stays between $0.5\%$ and
$2.1\%$ in every block. Successful trajectories, which are not annotated
at 1.5B, contribute no excess by construction.

\paragraph{Signed versus one-sided credit.}
\label{app:credit-checks}
An earlier ALFWorld~1.5B experiment with an environment-state milestone
potential in GiGPO's step channel compared signed credit with two
one-sided variants. In the progress-only arm, positive-credit events per
episode rose from $8.6$ to $17.3$ between the early and late halves of the
training stream while success fell from $17\%$ to $14\%$; in the
penalty-only arm success stayed near $3\%$ and the potential remained
low. The signed variant avoided both collapses. These diagnostics motivate
retaining both signs; they do not establish reward-farming immunity for
the clipped, learner-transformed signal.

\section{Annotation Cost}
\label{app:cost}

The default annotator is \texttt{deepseek-v4-flash} at temperature $0$.
In the fixed-budget runs, the primary annotation of a trajectory is one
call after it completes (all trajectories in the 7B and WebShop runs; failed
trajectories only in the ALFWorld~1.5B runs); the WebShop continuation
runs annotate prefixes turn by turn with a local annotator
(Appendix~\ref{app:webshop-cont}). An ALFWorld prompt
is about $7$k tokens (median $6{,}984$, 90th percentile $8{,}452$) and a
completion is $110$--$1{,}200$ tokens with the annotator's reasoning
disabled (median latency $16$\,s); a
training step of $128$ trajectories therefore costs about $0.9$M input
and $0.2$M output tokens, roughly \$0.2 at list price, or \$40 for the
200-step 1.5B run. In the 500-step ALFWorld~1.5B runs, which were launched
together on one host while the API endpoint also served about nine other
runs, a DARS step took about $340$--$420$\,s against $270$--$400$\,s for
GiGPO. In a pair of DARS runs that differ only in which endpoint answers
the annotation call, the API annotator took $512$\,s per step and the
distilled 8B annotator (Appendix~\ref{app:annot-quality}), served with
vLLM on one H100, took $506$\,s; policy generation is about $75\%$ of the
step, so annotation is not the bottleneck. These are step times observed
on shared hosts and a shared annotation endpoint, not a controlled speed
benchmark. The local annotator removes the
per-account API rate limit and the per-call cost: one H100 serves about
$627$ calls per minute on short prompts and $61$--$109$ per minute on the
$7$k-token ALFWorld prompts, depending on the output format. An earlier
measurement of $1{,}259$\,s per step, in the 150-step runs with 8
concurrent calls inside a synchronous rollout loop, was taken while the
shared API endpoint was congested and unstable; it reflects queuing at the
endpoint rather than annotation work. Overlapping annotation of one batch
with the next rollout would reduce the remaining overhead further. Enabling the annotator's reasoning mode is not
recommended: with a $4$k completion budget it silently exhausted the
budget on $17\%$ of trajectories (up to $47\%$ of the longest) in an early
7B run, and all reported ALFWorld~7B and fixed-budget WebShop~7B cells use the reasoning-off configuration, on
which the rejection rate was $0/128$ in a full replay and zero over the
WebShop~7B run.
On ALFWorld~1.5B, where only failed trajectories are annotated, the
number of annotation calls per step falls from about $116$ to $14$ over
training, so the load and the influence of the annotator both shrink as
the policy improves.

\section{Search-R1 Detail}
\label{app:search-detail}

The shared-pool diagnostic trains and evaluates on the same 700
questions. Table~\ref{tab:main-gigpo} compares final checkpoints after
662 GiGPO steps and 360 DARS steps; DARS is higher in all three decoding
groups. Accuracies pool all 700 questions and four draws per decoding
group. The harness logs the unweighted mean of its two validation
batches (512 and 188 questions), so we recompute the pooled accuracy,
and that of MuSiQue, the one subset split across them, from the logged
rates, which fix the per-batch counts of correct draws exactly. The held-out protocol uses disjoint training, selection, and
reporting pools as detailed below.

\begin{table}[h]
\centering
\small
\setlength{\tabcolsep}{4pt}
\caption{\textbf{Search-R1 shared-pool diagnostic by QA subset} (first decoding
group, four draws, final checkpoints; each subset has 100 questions, and
``All'' pools them). DARS is higher on six of seven subsets, with the
largest gains on the multi-hop HotpotQA and Bamboogle subsets. Accuracy in percent; $\Delta$ in points; DARS is shaded.}
\label{tab:search-subsets}
\begin{tabular}{lcccccccc}
\toprule
Arm & NQ & TriviaQA & PopQA & HotpotQA & 2Wiki & MuSiQue & Bamboogle & All \\
\midrule
GiGPO (step 662) & 33.3 & 66.0 & 49.5 & 37.5 & \textbf{46.8} & 15.0 & 28.5 & 39.5 \\
\rowcolor{oursbg}DARS (step 360)  & \textbf{39.0} & \textbf{66.2} & \textbf{54.7} & \textbf{50.2} & 42.7 & \textbf{17.7} & \textbf{37.2} & \textbf{44.0} \\
$\Delta$ & $+5.7$ & $+0.2$ & $+5.2$ & $+12.7$ & $-4.1$ & $+2.7$ & $+8.7$ & $+4.5$ \\
\bottomrule
\end{tabular}
\end{table}

On the shared pool, the peak-vs-peak comparison (each arm at its
training-validation maximum) gives $44.7\pm0.4\%$ for DARS against
$42.4\pm0.3\%$ for GiGPO, positive under all three decoding groups; the
final-checkpoint comparison of Table~\ref{tab:main-gigpo} is the one not
selected on the reported set. Under the held-out protocol
(Table~\ref{tab:search-heldout}) every arm trains for one epoch and is
read at its \texttt{val350} peak; the \texttt{val350} and \texttt{test350}
halves are stratified by subset, disjoint from each other and from the
training questions, and checksum-identical across hosts.

Removing either term of the DARS reward loses the held-out gain
(Table~\ref{tab:search-heldout}). With the answer-commitment terms alone,
the policy stops searching ($0.0$ search calls per question) and scores
$24.6\pm0.5\%$. With graph credit alone, training-validation accuracy
peaks at step~80 ($36.9\pm0.3\%$ on \texttt{test350}), and from step~330
the policy issues three searches on every validation question and scores
$0\%$. The commitment terms therefore prevent this collapse, while graph
credit supplies the retrieval behavior that the commitment terms alone
do not.

\begin{table}[h]
\centering
\small
\caption{\textbf{Search-R1 held-out protocol.} Exact-match accuracy on
\texttt{test350} at each arm's \texttt{val350}-selected checkpoint;
mean $\pm$ SE over three decoding groups of four draws each. The
commitment-only and graph-only arms each keep one of the two terms of the
DARS reward (a $2\times2$ factorial with GiGPO and DARS; same launcher,
rendering, and selection rule). Accuracy in percent; DARS rows are shaded.}
\label{tab:search-heldout}
\begin{tabular}{llcc}
\toprule
Arm & Selected step & \texttt{test350} & $\Delta$ vs.\ GiGPO (per group) \\
\midrule
GiGPO (our reproduction) & 600 & $39.5 \pm 0.4$ & -- \\
Commitment terms only & 100 & $24.6 \pm 0.5$ & $-15.0$ ($-15.7$, $-13.5$, $-15.7$) \\
Graph credit only & 80 & $36.9 \pm 0.3$ & $-2.6$ ($-2.9$, $-2.0$, $-3.0$) \\
\rowcolor{oursbg}DARS, $\sigma=0.5$ & 410 & $39.7 \pm 0.2$ & $+0.2$ \\
\rowcolor{oursbg}\textbf{DARS} (Table~\ref{tab:main-gigpo}) & 490 & $\mathbf{41.6 \pm 0.6}$ & $+2.1$ ($+1.8$, $+3.5$, $+1.0$) \\
\bottomrule
\end{tabular}
\end{table}

\section{Agentic Mathematics: Full Detail}
\label{app:sanity}

All cells use the protocol of Appendix~\ref{app:protocol}: Qwen3-8B,
Python interpreter with search disabled, four draws, LLM-equality
grading, no invalid draws. The untrained policy scores AIME24 $65.8\%$ /
AIME25 $50.0\%$ ($57.9\%$ mean). Table~\ref{tab:main-aepo} summarizes the
$2\times2$ design discussed in Section~\ref{sec:exp-reasoning}, and
Table~\ref{tab:aepo-grid} gives the per-benchmark values behind it; MATH500, GSM8K,
and MATH are at ceiling for this model and differ across arms by less
than the $\pm1$-point decoding variability (Table~\ref{tab:aepo-saturated}),
which is why the primary metric is the AIME mean.

\paragraph{Checkpoint comparison.}
We use the common retained seed-0 checkpoints at steps 5, 10, and 15,
matching optimizer updates, data order, and the number of checkpoint
choices across all four arms. Each arm's best score summarizes the
performance it attains on this grid while allowing the arms to peak at
different milestones. Peak-checkpoint reporting has precedent in
reasoning RL: Phi-4-reasoning selects its RL checkpoint by the best
observed AIME24 score \citep{abdin2025phi4reasoning}, and
\citet{petrenko2026entropypreserving} select by the highest mean
AIME24/25 score when a separate validation split is unavailable.
We additionally report all 12 matched-step cells and their grid means.
ARPO's maximum across its extended 13-milestone evaluation through
step~78 is also $62.9\%$ at step~15, so using the common grid preserves
its strongest observed checkpoint. ``Best'' is selected on the reported
AIME problems and denotes peak performance within the grid; it is not
an independently selected test estimate.

\paragraph{Evaluation variability.}
Four decoding draws on each of 60 problems yield 240 responses per
cell. The approximate $\pm5$-point variation describes decoding, not a
formal confidence interval or variation across training seeds. We
use the same evaluation protocol and draw count for every arm.

\begin{table}[h]
\centering
\small
\caption{\textbf{Matched-step results by benchmark.} Accuracy in percent; DARS arms are shaded.}
\label{tab:aepo-grid}
\begin{tabular}{llcccc}
\toprule
Step & Arm & MATH500 & AIME24 & AIME25 & Mean AIME \\
\midrule
\multirow{4}{*}{5}
 & ARPO              & ---   & 68.3 & 55.0 & 61.7 \\
 & AEPO              & 93.3 & \textbf{72.5} & 54.2 & \textbf{63.3} \\
 & \ours ARPO $+$ DARS & \ours 95.8 & \ours 64.2 & \ours 54.2 & \ours 59.2 \\
 & \ours AEPO $+$ DARS & \ours 93.3 & \ours 65.8 & \ours \textbf{55.0} & \ours 60.4 \\
\midrule
\multirow{4}{*}{10}
 & ARPO              & ---   & 65.8 & 56.7 & 61.3 \\
 & AEPO              & 90.0 & 65.8 & 57.5 & 61.7 \\
 & \ours ARPO $+$ DARS & \ours 91.7 & \ours 67.5 & \ours 55.0 & \ours 61.3 \\
 & \ours AEPO $+$ DARS & \ours 93.3 & \ours \textbf{70.0} & \ours \textbf{65.0} & \ours \textbf{67.5} \\
\midrule
\multirow{4}{*}{15}
 & ARPO              & ---   & \textbf{68.3} & 57.5 & 62.9 \\
 & AEPO              & 93.3 & 65.0 & 55.0 & 60.0 \\
 & \ours ARPO $+$ DARS & \ours 91.7 & \ours \textbf{68.3} & \ours \textbf{60.0} & \ours \textbf{64.2} \\
 & \ours AEPO $+$ DARS & \ours 95.0 & \ours 66.7 & \ours 59.2 & \ours 62.9 \\
\bottomrule
\end{tabular}
\end{table}

\begin{table}[h]
\centering
\small
\caption{\textbf{Saturated benchmarks} at each arm's best grid checkpoint. Accuracy in percent; DARS arms are shaded.}
\label{tab:aepo-saturated}
\begin{tabular}{lccc}
\toprule
Arm & MATH500 & GSM8K & MATH \\
\midrule
AEPO                & 94.8 & 95.6 & \textbf{97.3} \\
\rowcolor{oursbg}AEPO $+$ DARS       & \textbf{94.9} & \textbf{96.1} & 96.9 \\
\rowcolor{oursbg}ARPO $+$ DARS       & 94.6 & 95.8 & 96.2 \\
\bottomrule
\end{tabular}
\end{table}

\section{Tool-Free Reasoning: Full Arm Table and Diagnostics}
\label{app:reasoning}

This appendix expands Section~\ref{sec:exp-reasoning} and the step-credit
comparisons in Table~\ref{tab:dependency-ablation}(a). All cells use the
protocol of Appendix~\ref{app:protocol}: DAPO-Math prompt set, no tools,
one chain of thought per problem, decoding at temperature $1.0$ / top-$p$
$1.0$ with a $32{,}768$-token budget and $n=8$ draws,
symbolic-equivalence grading, and the average of AMC, AIME24, and
AIME25 in points as the reported score. An evaluation cell is a (checkpoint, host, repeat) triple and an
arm's score is the mean over its cells. All 1.7B arms train on the same
persisted rollouts; the DARS variants, the $\lambda=0$ arm, and the scalar
control additionally share the same 775 annotation graphs, re-credited
offline. In the 60-step 1.7B experiment, the main method means span
$56.90$--$58.23$ around the initialization ($57.48$); the brevity control
scores $55.83$. The within-checkpoint
spread of a single evaluation cell is about $0.9$ points, which is why
arm means over cells and exact permutation tests are used rather than
single cells or maxima.

\paragraph{Full arm table.}
Table~\ref{tab:reasoning-arms} reports the 1.7B arms, variants, and
controls (78 evaluation cells in total); the teacher-based arms distill from
a Qwen3-32B teacher. Arms marked $^{\ddagger}$ were evaluated on
a second host; cross-host comparisons use the average over these integer-answer
benchmarks, where grader disagreement is limited to a small number of
items, and a same-checkpoint cross-host check gave an offset of $-0.002$.

\begin{table}[h]
\centering
\footnotesize
\caption{\textbf{Reported arms, tool-free reasoning, Qwen3-1.7B chat (78 cells).}
Average of AMC, AIME24, and AIME25 (Avg.), as the mean over evaluation
cells (standard deviation across cells in parentheses); $p$ is an exact permutation test against DARS
($\eta{=}1$) where the cell composition permits one; for the
$\lambda{=}0$ arm the entry is the number of (checkpoint, host)-matched
cells on which DARS is higher.
$^{\ddagger}$ evaluated on the second host.}
\label{tab:reasoning-arms}
\resizebox{\textwidth}{!}{%
\begin{tabular}{@{}llccc@{}}
\toprule
Signal source & Arm & cells & Avg. & $\Delta$ vs.\ DARS ($p$/wins) \\
\midrule
\multirow{7}{*}{Dependency graph}
 & \textbf{DARS} ($\eta{=}1$, $\lambda{=}1$)     & 10 & \textbf{58.23} (1.09) & -- \\
 & DARS, $\lambda{=}0$ (same graphs)             & 14 & 57.50 (1.13) & $-0.73$ (7/10) \\
 & DARS ($\eta{=}2$)                             &  3 & 57.27 (1.75) & $-0.96$ \\
 & DARS, wider group admission                   &  3 & 57.14 (0.44) & $-1.09$ \\
 & DARS, broken-node penalty ($\rho_{\mathrm b}{=}0.25$) &  3 & 57.10 (0.15) & $-1.13$ \\
 & DARS, answer-commit gates                     &  3 & 57.08 (0.33) & $-1.15$ \\
 & DARS w/o step credit ($\eta{=}0$, same graphs) & 10 & 56.90 (0.76) & $-1.34$ ($0.002$) \\
\midrule
\multirow{2}{*}{Teacher}
 & On-policy distillation (vanilla)              &  4 & 57.32 (1.28) & $-0.92$ \\
 & OmniOPD                                       &  9 & 57.24 (0.70) & $-0.99$ ($0.031$) \\
\midrule
\multirow{2}{*}{Front-weighted scalar}
 & step-GRPO ($\alpha{=}0.02$)                   &  4 & 57.40 (1.54) & $-0.84$ ($0.27$) \\
 & FWTA-GRPO                                     &  3 & 56.81 (0.83) & $-1.43$ ($0.056$) \\
\midrule
Hidden states
 & StaRPO$^{\ddagger}$ (main table)              &  3 & 57.43 (0.88) & $-0.80$ \\
\midrule
\multirow{3}{*}{Controls}
 & GRPO, task reward only$^{\ddagger}$           &  3 & 57.69 (0.05) & $-0.54$ \\
 & Untrained model                               &  3 & 57.48 (0.77) & $-0.76$ ($0.30$) \\
 & Brevity control$^{\ddagger}$                  &  3 & 55.83 (1.54) & $-2.40$ \\
\bottomrule
\end{tabular}%
}
\end{table}

\paragraph{StaRPO configuration and controls.}
StaRPO is run with its paper-specified segmentation and intrinsic
reward. StaRPO's intrinsic reward admits
groups with zero task-reward variance, which account for $73.5\%$ of
prompts in this dataset; the brevity control uses the same
group-admission rule with a length penalty, helping distinguish reward
content from training-data volume, and scores below StaRPO. In groups with no task-reward variance, the intrinsic signal
determines the ranking for any positive coefficient.

\paragraph{Annotation gates.}
Before training, the annotation pass was checked on the persisted
rollouts (Table~\ref{tab:reasoning-gates}): $96.9\%$ of 1.7B rollouts
received a usable graph (against $61.4\%$ for an earlier step-level
judge); $142$ of $200$ groups have nonzero within-group variance of
$\Phi_G(T)$; $28.6\%$ of segments carry nonzero credit, of which $3.5\%$
is negative. The low share of negative credit means this setting mainly
tests the placement of positive progress credit; the broken-node-penalty
variant, which charges every currently broken predicate and raises the
negative share to $19$--$22\%$, did not improve on the default within its
three cells. The domains thus exercise different parts of the mechanism: mathematics
mainly the placement of positive credit, ALFWorld the attenuation
(Appendix~\ref{app:dependency-ablation}), and WebShop the topology
(Appendix~\ref{app:topology-diagnostics}); the cross-domain gains do not
imply that error propagation and repair dominate everywhere.

\begin{table}[h]
\centering
\small
\caption{\textbf{Annotation gates, tool-free reasoning.} Coverage is compared with the reference step judge.}
\label{tab:reasoning-gates}
\begin{tabular}{lcc}
\toprule
Gate & 1.7B & 4B \\
\midrule
Usable induced chains & 96.9\% (prior judge 61.4\%) & 91.2\% \\
$\rho(\Phi_G(T), \text{outcome})$ & $+0.468$ & $+0.218$ \\
Answer-node agreement with the outcome verifier & 79\% & 80.2\% \\
Groups with nonzero $\Phi$ variance & 142/200 & 263 groups \\
Segments carrying credit & 28.6\% & 28.3\% \\
Negative credit share & 3.5\% & 1.2\% \\
\bottomrule
\end{tabular}
\end{table}

The answer-node agreement figures are conditional on the annotation
providing an \texttt{answer}-node verdict; in the 4B audit, $29.0\%$ of
the 1{,}469 usable annotations provide no such verdict (an omission), and
the $80.2\%$ agreement is measured over the remaining annotations.

\paragraph{4B training configuration and results.}
The 4B experiments use $400$ prompts with $8$ rollouts each, a
$32{,}768$-token generation cap, and $91.2\%$ annotation coverage,
yielding $263$ usable groups. DARS and its scalar control train for
$200$ steps at learning rate $10^{-5}$ and sequence length $12{,}288$;
their relative weight changes from initialization are
$1.22\times10^{-3}$ and $1.39\times10^{-3}$, and weight changes are also
verified for the distillation and StaRPO baselines before evaluation.
Table~\ref{tab:reasoning-4b} reports the pooled means behind the 4B rows
of Table~\ref{tab:main-math} and the primary-host contrasts. The
primary-host DARS--OmniOPD comparison uses 16 and 9 cells, respectively,
with average scores of $79.49$ and $78.29$ ($p=0.005$).
This test uses primary-host cells, whereas the main table reports
pooled means across hosts.

\begin{table}[h]
\centering
\small
\caption{\textbf{Tool-free reasoning at 4B (Qwen3-4B chat).}
Avg.\ is the mean of AMC, AIME24, and AIME25, and AIME is the mean of
AIME24 and AIME25. The upper panel pools
evaluation cells across hosts, with counts $n$ and standard deviations
in parentheses. The lower panel reports the primary-host analysis,
using $16$, $15$, $13$, and $9$ cells for DARS, initialization, the scalar
control, and OmniOPD, respectively, with permutation $p$-values; this analysis
uses a different evaluation-cell set from the pooled summary, so its
counts are not a subset of the upper panel.}
\label{tab:reasoning-4b}
\begin{tabular}{lccc}
\toprule
Arm & $n$ & Avg. & AIME \\
\midrule
\textbf{DARS} ($\eta{=}1$)        & 20 & \textbf{79.50} (1.06) & \textbf{70.88} \\
Untrained model                   & 14 & 79.17 (0.77) & 70.15 \\
DARS w/o step credit              & 13 & 79.13 (0.81) & 69.95 \\
StaRPO (paper segmentation)       &  4 & 78.94 (1.17) & 69.90 \\
Rejection fine-tuning             &  4 & 78.79 (1.16) & 69.64 \\
On-policy distillation            & 15 & 78.62 (0.96) & 69.51 \\
OmniOPD                          & 12 & 78.44 (0.57) & 69.50 \\
\midrule
\multicolumn{4}{l}{\emph{Contrasts (primary-host cells)}} \\
DARS $-$ OmniOPD          & & $+1.20$ ($p = 0.005$) & $+1.59$ ($p = 0.008$) \\
DARS $-$ w/o step credit  & & $+0.36$ ($p = 0.32$) & $+0.89$ ($p = 0.090$) \\
DARS $-$ untrained        & & $+0.35$ ($p = 0.30$) & $+0.65$ ($p = 0.16$) \\
w/o step credit $-$ untrained & & $-0.01$ ($p = 0.98$) & $-0.24$ ($p = 0.58$) \\
\bottomrule
\end{tabular}
\end{table}

\paragraph{Annotation diagnostics.}
An audit of $1469$ annotations on the 4B training data
found internal contradictions in about $1\%$ of graphs; the annotator
leaves the answer verdict unspecified on $29.0\%$ of rollouts, and the
first fifth of the response receives $50.8\%$ of absolute credit mass. A
variant that fixes one group-consensus decomposition per problem raises
the potential's discrimination of the outcome ($\mathrm{AUC}$
$0.86\to0.97$) but concentrates credit on the final segment of failed
rollouts (mid-trajectory credited segments $1.8\to1.3$ per failed rollout)
and did not improve training within its ten cells ($57.00$), so we retain
per-rollout instantiation. The matched scalar control uses the same
graphs, events, and rollout groups as DARS, so the step-credit comparison
holds this annotation basis fixed.

\paragraph{Broken-node-penalty variant.}
This variant replaces Eq.~(\ref{eq:phi}) with
$\widetilde{\Phi}_G(t)=\big(\sum_i S_i(t)\,w_i(t)-\rho_{\mathrm b}\,|\mathcal U(t)|\big)/M$,
where $S_i(t)$ and $\mathcal U(t)$ are the predicate states and broken set of
Eq.~(\ref{eq:state-transition}) and $\rho_{\mathrm b}=0.25$; the penalty charges every currently broken predicate,
including ones never supported, and is refunded on repair.

\section{Dependency and Topology Ablations: Protocols and Offline Census}
\label{app:dependency-ablation}

\subsection{What each ablation removes}

Setting $\lambda=0$ makes the dependency-distance factor equal to one:
supported predicates keep unit weight even when they depend on an
unrepaired error, while unestablished or broken predicates still
contribute zero. Graph instantiation, verification and repair events, and
stepwise potential differences are unchanged, so this removes attenuation
along dependency edges and nothing else. The scalar control ($\eta=0$)
instead removes the step channel and keeps the final potential. The
sequential-chain variant keeps every annotation event but replaces the
domain topology by a chain in trajectory order, so that an error
attenuates every later predicate whether or not it depends on the error.

\subsection{Protocols}

\paragraph{Tool-free mathematics, 1.7B.}
The $\lambda=0$ arm re-credits the 775 persisted annotation graphs of the
DARS arm at $\lambda=0$ and trains on the same 142 groups and 551 rollouts
with matched initialization, optimizer, and 60-step budget; the scalar
control uses the same graphs with $\eta=0$. Evaluation cells are
(checkpoint, host, repeat) triples; the $\lambda=0$ arm has ten cells on
the first host and four on the second, the DARS arm six and four. The
$\lambda=1$ arm is higher on 7 of the 10 (checkpoint, host)-matched
pairs, with a mean paired advantage of $0.51$ points (the pooled
14-versus-10 arm-mean difference is $0.73$ points); the direction is
consistent, and the DARS-versus-scalar difference has exact permutation
$p=0.002$.

\paragraph{ALFWorld 1.5B (200-step recipe).}
The $\lambda=0$ twin of the 200-step DARS run of
Table~\ref{tab:local-baselines} uses the identical launcher,
initialization, annotator, and 200-step budget with only
$\lambda$ changed, trained on the same host and evaluated with the same
harness and the same 128-game parquet as the 200-step cells (four
paired environment seeds). At the final step 200 it scores $86.0\%$
(draws $86.7$, $85.2$, $85.2$, $86.7$) against $92.0\%$ for
$\lambda=1$ ($92.2$, $92.2$, $92.2$, $91.4$), a paired difference of
$-6.1$ points with $\lambda=1$ ahead on all four draws; at step 195 the
difference is $-5.1$ points (4/4), and the $\lambda=0$ arm's best checkpoint
(keep-best, step 180, $87.9\%$) remains $4.1$ points below the 200-step DARS
cell on all four draws. A same-day re-evaluation of the $\lambda=1$ checkpoint
reproduced its four draws exactly, so the comparison is not affected by
harness drift.

\paragraph{ALFWorld 7B.}
The $\lambda=0$ twin uses the full 7B recipe (annotator reasoning off,
State/Action/Result rendering) and resumes from the same step-300
checkpoint as the DARS arm to step 500 with only
$\lambda$ changed; its own annotation dump recomputes bit-exactly at
$\lambda=0$. Both arms are read at their keep-best checkpoint (both at
step 445) on the pinned harness with paired decoding seeds: $95.2\%$ over
15 draws for $\lambda=0$ against $98.6\%$ for $\lambda=1$ (paired
difference $-3.3$ points, standard deviation $1.3$, $\lambda=1$ higher on
15/15 draws, sign-flip $p<0.001$).

\paragraph{WebShop 1.5B topology.}
Two arms with the $\kappa=0.3$ recipe and the WebShop annotator were
trained for 150 steps from the same initialization: the
parallel-AND requirement graph and the sequential chain. Both were
evaluated at the final checkpoint on one host with four paired decoding
draws and both metrics.
The requirement graph scores $63.2\%$ success / $79.5\%$ task score and the
sequential chain $49.0\%$ / $61.1\%$; the chain is lower on all four paired
draws for both metrics ($-14.2$ and $-18.4$ points on average). The
requirement-graph arm is an independent training run of the same
$\kappa=0.3$ configuration as the $\kappa=0.3$ row of
Table~\ref{tab:webshop-variants} ($63.0\%$ / $77.6\%$); the two runs agree
to within $0.2$ points of success and $1.9$ of task score.

\paragraph{ALFWorld 1.5B decay sweep from initialization.}
An earlier sweep trained DARS from initialization at four values of
$\lambda$ with the distilled local annotator (Appendix~\ref{app:annot-quality}),
$\kappa=0.3$, and a 150-step budget, and evaluated each arm's
training-validation peak checkpoint on the unseen games with four draws
(Table~\ref{tab:lambda-sweep}). $\lambda=\infty$ zeroes every predicate
below a broken ancestor (hard masking, a dependency-ordered analogue of first-error credit). The sweep orders moderate
decay ($\lambda=0.5$: $94.3\%$; $\lambda=1$: $91.4\%$) above milestone
counting ($\lambda=0$: $90.2\%$) above hard masking
($\lambda=\infty$: $84.2\%$); the same ordering holds for the normalized
area under the validation curve over steps 0--150 relative to the
$\lambda=0$ arm ($+0.030$ for $\lambda=0.5$, $+0.026$ for $\lambda=1$).
Because these arms use a different annotator and budget from the
200-step recipe, the sweep is reported as supporting the direction of
the matched twin above rather than as a second matched estimate.
Soft attenuation's advantage over hard masking could reflect useful local
content in dependent steps or tolerance to annotation errors; the sweep
does not distinguish the two, but it argues against treating stronger
suppression as automatically better.

\begin{table}[h]
\centering
\small
\caption{\textbf{Decay-strength sweep on ALFWorld 1.5B from initialization.}
Unseen-game success at the training-validation peak checkpoint, four
draws per cell; distilled annotator, $\kappa=0.3$, 150 steps. Success in percent; $\Delta$ in points.}
\label{tab:lambda-sweep}
\begin{tabular}{lcc}
\toprule
$\lambda$ & Success & $\Delta$ vs.\ $\lambda=0$ \\
\midrule
$0$ (milestone counting) & 90.2 & -- \\
$0.5$ & \textbf{94.3} & $+4.1$ \\
$1$ (default) & 91.4 & $+1.2$ \\
$\infty$ (hard masking) & 84.2 & $-6.1$ \\
\bottomrule
\end{tabular}
\end{table}

\subsection{Offline census: which trajectories an ablation touches}
\label{app:census}

Because credit is a deterministic function of the persisted annotations,
each ablation's footprint can be measured offline on real annotated
graphs before any training (Table~\ref{tab:census}). Two observations
explain the end-to-end results. First, the dependency attenuation
touches few trajectories ($2$--$12\%$ across environments) because it
requires an error \emph{followed by} dependent work; the effect of
$\lambda$ is therefore concentrated on exactly those trajectories (the
census locates where the rule acts, not which of these trajectories
produced the online gains).
Second, serializing the graph is nearly inert on ALFWorld, whose
per-object chains are already sequential (worst-decile credit change
$-2.8\%$), but severe on WebShop, whose graphs fan out from
\texttt{type} (a quarter of affected trajectories lose more than half of
their credit). This is why the topology ablation, rather than the attenuation
ablation, is run on WebShop: on the persisted depth-one WebShop graphs
the attenuation changes $8.6\%$ of reward vectors and never removes more
than half of a trajectory's credit, whereas serializing the graph costs
$14.2$ points of success online. The online attenuation comparisons are run on
the deeper ALFWorld chains and mathematics derivations, where they favor
$\lambda=1$ in direction in all three cases (4/4 paired draws at
ALFWorld~1.5B, 7/10 matched cells on mathematics) and significantly at
ALFWorld~7B (15/15, $p<0.001$).

\begin{table}[h]
\centering
\small
\setlength{\tabcolsep}{4pt}
\caption{\textbf{Offline census of the ablations on persisted annotation graphs.}
``Changed'' is the fraction of trajectories whose reward vector differs
from the DARS reward; the remaining columns describe the distribution of
the relative change in total credit over the changed trajectories.}
\label{tab:census}
\begin{tabular}{@{}llrrrr@{}}
\toprule
Corpus & Variant & Changed & Mean change & 10th pct. & $>50\%$ lost \\
\midrule
\multirow{2}{*}{WebShop 1.5B (371 graphs)}
 & sequential chain & 12.7\% & $-20.4\%$ & $-99.5\%$ & 25.2\% \\
 & $\lambda=0$ & 8.6\% & $+27.6\%$ & $+0.0\%$ & 0.0\% \\
\midrule
\multirow{2}{*}{ALFWorld 1.5B (2000 graphs)}
 & sequential chain & 10.6\% & $-1.0\%$ & $-2.8\%$ & 0.5\% \\
 & $\lambda=0$ & 11.7\% & $-1.0\%$ & $-9.5\%$ & 1.2\% \\
\midrule
ALFWorld 7B (27{,}520 graphs) & $\lambda=0$ & 2.0\% & -- & -- & -- \\
Math 1.7B (775 graphs) & $\lambda=0$ & 10.6\% & -- & -- & -- \\
\bottomrule
\end{tabular}
\end{table}

\subsection{A graph-free process reward: first-error prefix credit}
\label{app:topology-diagnostics}

Before the graph reward, we trained a simpler process reward with the
same annotator model: it localizes the first erroneous turn of a failed
trajectory (as in ProcessBench-style first-error localization) and
credits the turns before it with front-weighted credit in GiGPO's step
channel, with no predicates and no graph. Table~\ref{tab:topology}
collects the outcomes. On ALFWorld, whose goals are chains, this reward
improves over GiGPO by $+3.2$ points (training-validation peaks, 150 steps),
consistent with the census above that chain credit and
graph credit nearly coincide there. On WebShop it lowered the graded task
score by $42$ points relative to GiGPO because a correct browsing prefix earns
credit without a purchase, and on Search-R1 the policy collapsed to
$0\%$ success because searching earns prefix credit while committing an
answer does not ($96\%$ of trajectories never answered). A prefix is not a
model of what the task requires; the predicate graph is, which is why the
WebShop and Search-R1 rewards define progress through requirement and
hop predicates and why Search-R1 additionally uses the answer-commitment
term of Appendix~\ref{app:protocol}. These earlier studies used earlier
protocols and are reported as motivation, not as matched entries of
Table~\ref{tab:main-gigpo}.

\begin{table}[!htbp]
\centering
\small
\caption{\textbf{First-error prefix credit as a graph-free process reward.} The same annotator
model localizes the first erroneous turn and the turns before it receive front-weighted credit
(no predicates, no dependency graph). Outcomes relative to the GiGPO baseline in each
environment's earlier local study (training-validation peaks for ALFWorld).}
\label{tab:topology}
\resizebox{\textwidth}{!}{%
\begin{tabular}{llll}
\toprule
Environment & Goal structure & First-error prefix credit vs.\ GiGPO & Observed behavior \\
\midrule
ALFWorld  & ordered chain      & $+3.2$ points success & chain credit and graph credit nearly coincide \\
WebShop   & parallel-AND       & $-42$ points task score       & credits 7 browsing turns without a purchase \\
Search-R1 & instrumental prefix & success falls to $0\%$       & $96\%$ of trajectories never commit an answer \\
\bottomrule
\end{tabular}
}
\end{table}

\section{Recorded Credit and Paired Policy Episodes}
\label{app:qualitative}

We examine which intermediate decisions the local signal distinguishes and
how the trained policies handle corresponding decisions. The training
examples below contain recorded graph events and shaping credits. The
paired evaluation examples contain actions and environment observations,
but \emph{no recorded DARS shaping credits}. We therefore describe their
connection to the training rule without assigning retrospective rewards.
All action sequences are complete; repeated page text and model reasoning
are omitted, and the observations needed to interpret the decisions are
reported in the accompanying text. Turn indices start at zero.

\subsection{Recorded credit within complete training trajectories}
\label{app:qual-credit}

The two WebShop~7B examples illustrate completion of individual
requirements and loss and recovery of established progress. The displayed credit is the recorded
$\operatorname{clip}(\Delta\Phi_G,-0.3,0.3)$ with scale~1, not the final
policy advantage. Each graph and its annotations are kept as recorded.
Clipping means the paid credits need not sum to the endpoint potential.

\paragraph{Completing the requested options.}
The task requests Blu-ray-compatible, gold-plated, high-speed, heavy-duty
HDMI cables, in blue braided color and a 20-pack, for less than \$40.
The agent opens product \texttt{b08qsnm69h}, whose page lists both requested
options and a price of \$15.49. It selects the size and color before buying
(Table~\ref{tab:webshop-71441-success}). The receipt records
\texttt{size: 20-pack} and \texttt{color: blue braided}; the task score is~1.
Both option clicks receive $+0.125$ before purchase. Buying then receives
zero shaping credit because the annotated requirements already saturate
the potential. The local feedback identifies the contributions of the
option selections within the successful episode.

In the HDMI example, T denotes product type, P price, S size, and C
color, and A1--A4 denote Blu-ray compatibility, gold plating, high speed,
and heavy-duty construction. ``Verify,'' ``Error,'' and
``Repair'' denote the recorded annotation events.

\begingroup\small
\setlength{\tabcolsep}{4pt}
\begin{longtable}{@{}r >{\raggedright\arraybackslash}p{0.40\linewidth} >{\raggedright\arraybackslash}p{0.28\linewidth} rr@{}}
\caption{Completing both requested options (record 71441). All five actions are shown.}\label{tab:webshop-71441-success}\\
\toprule
$t$ & Executed action & Recorded events & $\Phi_G$ & Credit \\
\midrule
\endfirsthead
\multicolumn{5}{l}{\small Table~\thetable\ (continued)}\\
\toprule
$t$ & Executed action & Recorded events & $\Phi_G$ & Credit \\
\midrule
\endhead
\bottomrule
\endfoot
0 & \texttt{search[blu ray gold plated high speed heavy duty HDMI cable blue braided 20-pack price \textless{} 40.00]} & \textemdash{} & $0.000$ & $+0.000$ \\*
1 & \texttt{click[b08qsnm69h]} & Verify: T, A1, A2, A3, A4, P & $0.750$ & $+0.300$ \\*
2 & \texttt{click[20-pack]} & Verify: S & $0.875$ & $+0.125$ \\*
3 & \texttt{click[blue braided]} & Verify: C & $1.000$ & $+0.125$ \\*
4 & \texttt{click[buy now]} & \textemdash{} & $1.000$ & $+0.000$ \\
\end{longtable}
\endgroup

\paragraph{Distinguishing abandonment from recovery within a success.}
The third task requests a bronze-finish vanity light below \$110.
The agent opens a matching \$77.99 product, leaves it, searches again,
returns to the same product, and buys successfully
(Table~\ref{tab:webshop-71074-recovery}). Leaving loses two verified
attributes and receives $-0.3$; reopening restores them and receives
$+0.3$. Type and price remain supported during the detour. A final
success label cannot express this difference between losing and restoring
progress. Here the detour earns zero net shaping credit; this is a
property of this recorded sequence, not a general guarantee after clipping.
In this table, B denotes bronze finish and L vanity light.

\begingroup\small
\setlength{\tabcolsep}{4pt}
\begin{longtable}{@{}r >{\raggedright\arraybackslash}p{0.40\linewidth} >{\raggedright\arraybackslash}p{0.28\linewidth} rr@{}}
\caption{Abandoning and recovering a suitable product (record 71074). All six actions are shown.}\label{tab:webshop-71074-recovery}\\
\toprule
$t$ & Executed action & Recorded events & $\Phi_G$ & Credit \\
\midrule
\endfirsthead
\multicolumn{5}{l}{\small Table~\thetable\ (continued)}\\
\toprule
$t$ & Executed action & Recorded events & $\Phi_G$ & Credit \\
\midrule
\endhead
\bottomrule
\endfoot
0 & \texttt{search[Vanity bronze finish vanity light price \textless{} 110.00]} & \textemdash{} & $0.000$ & $+0.000$ \\*
1 & \texttt{click[b09j9y9h95]} & Verify: T, B, L, P & $1.000$ & $+0.300$ \\*
2 & \texttt{click[back to search]} & Error: B, L & $0.500$ & $-0.300$ \\*
3 & \texttt{search[ Kira Home Ainsley 21.5" 3-Light Farmhouse Vanity/Bathroom Light + Clear Cylinder Glass Shades, Oil-Rubbed Bronze Finish \$77.99]} & \textemdash{} & $0.500$ & $+0.000$ \\*
4 & \texttt{click[b09j9y9h95]} & Verify: T, B, L, P; Repair: B, L & $1.000$ & $+0.300$ \\*
5 & \texttt{click[buy now]} & \textemdash{} & $1.000$ & $+0.000$ \\
\end{longtable}
\endgroup

\subsection{Using a retrieved fact in the next query}
\label{app:qual-search}

Search question~178 asks for the birthdate of the Uruguayan former
footballer whose management team includes Charlie Oatway.
\emph{The first retrieval results are byte-identical for both policies.}
They state that Oatway continues ``his integral role within Gus Poyet's
management team.'' DARS then queries Poyet; the returned biography gives
15~November~1967, which it answers correctly. GiGPO instead queries
Gast\'on Pereyra, a name absent from the question and first results.
The new results concern Dar\'io Pereyra; GiGPO repeats its query and
answers 30~May~1984, incorrectly (Table~\ref{tab:search-q178}).

The decisive behavior is using the established person to resolve the
remaining fact. DARS's search annotation rule verifies a hop when the
retrieval contains its sub-fact, and marks a query that corrupts the
required entity or relation as an error. This lets training retain the
useful first retrieval while distinguishing a continuation that changes
the subject. An off-topic or repeated query does not automatically incur
a negative reward: negative shaping requires a loss of credited potential.
The selected policy contrast shows the practical benefit of the distinction:
the correct answer with two searches rather than an incorrect answer
after three.

\begingroup\small
\setlength{\tabcolsep}{4pt}
\begin{longtable}{@{}r >{\raggedright\arraybackslash}p{0.44\linewidth} >{\raggedright\arraybackslash}p{0.44\linewidth}@{}}
\caption{Search: using the retrieved intermediate entity. Complete action sequences; no shaping credits were recorded at evaluation.}\label{tab:search-q178}\\
\toprule
$t$ & DARS & GiGPO \\
\midrule
\endfirsthead
\multicolumn{3}{l}{\small Table~\thetable\ (continued)}\\
\toprule
$t$ & DARS & GiGPO \\
\midrule
\endhead
\bottomrule
\endfoot
0 & \textbf{Search:} \texttt{Uruguayan former footballer who managed team with Charlie Oatway} & \textbf{Search:} \texttt{What is the birthdate of this Uruguayan former footballer, who management team consists of Charlie Oatway} \\*
1 & \textbf{Search:} \texttt{Gus Poyet Uruguayan football manager} & \textbf{Search:} \texttt{Gastón Pereyra birthdate uruguayan former footballer} \\*
2 & \textbf{Answer:} \texttt{15 November 1967} & \textbf{Search:} \texttt{Gastón Pereyra birthdate uruguayan former footballer} \\*
3 & \textemdash{} & \textbf{Answer:} \texttt{30 May 1984} \\
\end{longtable}
\endgroup

\subsection{Avoiding irrelevant work within a successful episode}
\label{app:qual-object}

In ALFWorld (1.5B, environment~42), the task is to put two
soapbars in the garbage can. Both policies first visit countertop~1 and
observe candles, a soap bottle, and a spray bottle. DARS leaves to look
for soapbars, finds two on toilet~1, and delivers them. GiGPO first takes
the soap bottle and delivers it, then returns to the search and eventually
delivers the soapbars as well (Table~\ref{tab:alf-soapbars}). Both succeed,
in nine and fourteen actions, respectively.

The identity gate makes this distinction explicit: picking up an available
object is insufficient to verify the requested object's pickup predicate.
Likewise, the correct receptacle does not make a wrong-object delivery
useful. The final success label does not identify the irrelevant bottle
delivery, whereas local task-progress credit can withhold reward from it.
The five-action difference also includes a later navigation detour; it
is not assigned entirely to the initial pickup.

\begingroup\small
\setlength{\tabcolsep}{4pt}
\begin{longtable}{@{}r >{\raggedright\arraybackslash}p{0.44\linewidth} >{\raggedright\arraybackslash}p{0.44\linewidth}@{}}
\caption{ALFWorld: avoiding a soap-bottle detour on a soapbar task. Complete action sequences; both policies succeed.}\label{tab:alf-soapbars}\\
\toprule
$t$ & DARS & GiGPO \\
\midrule
\endfirsthead
\multicolumn{3}{l}{\small Table~\thetable\ (continued)}\\
\toprule
$t$ & DARS & GiGPO \\
\midrule
\endhead
\bottomrule
\endfoot
0 & \texttt{go to countertop 1} & \texttt{go to countertop 1} \\*
1 & \texttt{go to toilet 1} & \texttt{take soapbottle 1 from countertop 1} \\*
2 & \texttt{take soapbar 1 from toilet 1} & \texttt{go to garbagecan 1} \\*
3 & \texttt{go to garbagecan 1} & \texttt{move soapbottle 1 to garbagecan 1} \\*
4 & \texttt{move soapbar 1 to garbagecan 1} & \texttt{go to countertop 1} \\*
5 & \texttt{go to toilet 1} & \texttt{go to toilet 1} \\*
6 & \texttt{take soapbar 2 from toilet 1} & \texttt{take soapbar 1 from toilet 1} \\*
7 & \texttt{go to garbagecan 1} & \texttt{go to garbagecan 1} \\*
8 & \texttt{move soapbar 2 to garbagecan 1} & \texttt{move soapbar 1 to garbagecan 1} \\*
9 & \textemdash{} & \texttt{go to countertop 1} \\*
10 & \textemdash{} & \texttt{go to toilet 1} \\*
11 & \textemdash{} & \texttt{take soapbar 2 from toilet 1} \\*
12 & \textemdash{} & \texttt{go to garbagecan 1} \\*
13 & \textemdash{} & \texttt{move soapbar 2 to garbagecan 1} \\
\end{longtable}
\endgroup

\subsection{Completing the second required object}
\label{app:qual-count}

In the two-pillow task (environment~119), both policies observe
pillow~1 and pillow~2 on armchair~1, then place pillow~1 on sofa~1.
DARS returns to the armchair, takes pillow~2, and completes the task in
13~actions. GiGPO repeatedly tries to take pillow~1 from sidetable~1,
receives ``Nothing happens,'' and exhausts the 50-action budget
(Table~\ref{tab:alf-second-pillow}). Both trajectories had already exposed
the second object's location.

The count-aware graph assigns separate chains to the required copies.
Completing one therefore leaves another available for new progress
credit; repeating the first does not establish the second. This case
illustrates acting on the outstanding instance after partial success.
It does not imply that DARS always preserves completed work: other
archived trajectories rehandle previously placed objects.

\begingroup\small
\setlength{\tabcolsep}{4pt}
\begin{longtable}{@{}r >{\raggedright\arraybackslash}p{0.44\linewidth} >{\raggedright\arraybackslash}p{0.44\linewidth}@{}}
\caption{ALFWorld: completing the second required pillow. Complete action sequences; DARS succeeds at action 13, GiGPO reaches the 50-action limit.}\label{tab:alf-second-pillow}\\
\toprule
$t$ & DARS & GiGPO \\
\midrule
\endfirsthead
\multicolumn{3}{l}{\small Table~\thetable\ (continued)}\\
\toprule
$t$ & DARS & GiGPO \\
\midrule
\endhead
\bottomrule
\endfoot
0 & \texttt{go to sidetable 1} & \texttt{go to sidetable 1} \\
1 & \texttt{go to shelf 1} & \texttt{go to dresser 1} \\
2 & \texttt{go to dresser 1} & \texttt{go to sofa 1} \\
3 & \texttt{go to sofa 1} & \texttt{go to sidetable 1} \\
4 & \texttt{go to sidetable 1} & \texttt{go to armchair 1} \\
5 & \texttt{go to armchair 1} & \texttt{take pillow 1 from armchair 1} \\
6 & \texttt{take pillow 1 from armchair 1} & \texttt{go to sofa 1} \\
7 & \texttt{go to sofa 1} & \texttt{move pillow 1 to sofa 1} \\
8 & \texttt{move pillow 1 to sofa 1} & \texttt{go to sidetable 1} \\
9 & \texttt{go to armchair 1} & \texttt{go to dresser 1} \\
10 & \texttt{take pillow 2 from armchair 1} & \texttt{go to sofa 1} \\
11 & \texttt{go to sofa 1} & \texttt{take pillow 1 from sidetable 1} (no effect) \\
12 & \texttt{move pillow 2 to sofa 1} & \texttt{go to sidetable 1} \\
13 & \textemdash{} & \texttt{go to dresser 1} \\
14 & \textemdash{} & \texttt{go to sofa 1} \\
15 & \textemdash{} & \texttt{take pillow 1 from sidetable 1} (no effect) \\
16 & \textemdash{} & \texttt{go to sidetable 1} \\
17 & \textemdash{} & \texttt{go to dresser 1} \\
18 & \textemdash{} & \texttt{go to shelf 1} \\
19 & \textemdash{} & \texttt{go to sidetable 1} \\
20 & \textemdash{} & \texttt{go to sofa 1} \\
21 & \textemdash{} & \texttt{take pillow 1 from sidetable 1} (no effect) \\
22 & \textemdash{} & \texttt{go to dresser 1} \\
23 & \textemdash{} & \texttt{go to sidetable 1} \\
24 & \textemdash{} & \texttt{go to sofa 1} \\
25 & \textemdash{} & \texttt{take pillow 1 from sidetable 1} (no effect) \\
26 & \textemdash{} & \texttt{go to dresser 1} \\
27 & \textemdash{} & \texttt{go to sidetable 1} \\
28 & \textemdash{} & \texttt{go to sofa 1} \\
29 & \textemdash{} & \texttt{take pillow 1 from sidetable 1} (no effect) \\
30 & \textemdash{} & \texttt{go to dresser 1} \\
31 & \textemdash{} & \texttt{go to sidetable 1} \\
32 & \textemdash{} & \texttt{go to sofa 1} \\
33 & \textemdash{} & \texttt{take pillow 1 from sidetable 1} (no effect) \\
34 & \textemdash{} & \texttt{go to dresser 1} \\
35 & \textemdash{} & \texttt{go to sidetable 1} \\
36 & \textemdash{} & \texttt{go to sofa 1} \\
37 & \textemdash{} & \texttt{take pillow 1 from sidetable 1} (no effect) \\
38 & \textemdash{} & \texttt{go to dresser 1} \\
39 & \textemdash{} & \texttt{go to sidetable 1} \\
40 & \textemdash{} & \texttt{go to sofa 1} \\
41 & \textemdash{} & \texttt{take pillow 1 from sidetable 1} (no effect) \\
42 & \textemdash{} & \texttt{go to dresser 1} \\
43 & \textemdash{} & \texttt{go to sidetable 1} \\
44 & \textemdash{} & \texttt{go to sofa 1} \\
45 & \textemdash{} & \texttt{take pillow 1 from sidetable 1} (no effect) \\
46 & \textemdash{} & \texttt{go to dresser 1} \\
47 & \textemdash{} & \texttt{go to sidetable 1} \\
48 & \textemdash{} & \texttt{go to sofa 1} \\
49 & \textemdash{} & \texttt{take pillow 1 from sidetable 1} (no effect) \\
\end{longtable}
\endgroup

\subsection{Checking the condition that makes a candidate eligible}
\label{app:qual-exclusion}

Question~181 asks for a starting pitcher on the 2000 Olympic Team USA
who was \emph{not} born on 29~August~1977. DARS retrieves the team and
then searches Ben Sheets's birthdate. The result states 18~July~1978,
which establishes the exclusion; DARS answers Ben Sheets correctly.
GiGPO answers Juan Pe\~na after one search, although that name is absent
from its retrieved text (Table~\ref{tab:search-q181}). Finding a candidate
and verifying the condition that makes it eligible are distinct
contributions that the hop annotations can identify. This example shows
a purposeful additional lookup, not a reduction in search count or an
exhaustive proof that no other pitcher qualifies.

\begingroup\small
\setlength{\tabcolsep}{4pt}
\begin{longtable}{@{}r >{\raggedright\arraybackslash}p{0.44\linewidth} >{\raggedright\arraybackslash}p{0.44\linewidth}@{}}
\caption{Search: checking the birthdate exclusion. Complete action sequences; no shaping credits were recorded at evaluation.}\label{tab:search-q181}\\
\toprule
$t$ & DARS & GiGPO \\
\midrule
\endfirsthead
\multicolumn{3}{l}{\small Table~\thetable\ (continued)}\\
\toprule
$t$ & DARS & GiGPO \\
\midrule
\endhead
\bottomrule
\endfoot
0 & \textbf{Search:} \texttt{Team USA baseball at 2000 Summer Olympics} & \textbf{Search:} \texttt{Which of the starting pitchers for Team USA in the Baseball at the 2000 Summer Olympics was not born August 29, 1977} \\*
1 & \textbf{Search:} \texttt{Ben Sheets born date 1977 August 29} & \textbf{Answer:} \texttt{Juan Peña} \\*
2 & \textbf{Answer:} \texttt{Ben Sheets} & \textemdash{} \\
\end{longtable}
\endgroup

\subsection{Selection, archive counts, and interpretation}
\label{app:qual-scope}

The examples were selected retrospectively for a visible local decision
and its consequence. They illustrate how the credit rule can encourage
useful operations; they do not isolate the training updates that produced
each action. The WebShop records come from a separate, later 7B training
run with the corrected annotator rendering, not the checkpoint in
Table~\ref{tab:main-gigpo}; they establish the signal available during
training, not an additional WebShop performance gain.

The paired Search cases come from a separate 350-question archive, not
the 700-question shared-pool diagnostic in Section~\ref{sec:exp-main}.
The archived DARS and GiGPO checkpoints are selected peaks (steps 180
and 600) of a later run with the corrected annotator rendering, with
comparable aggregate accuracy in this archive; DARS additionally uses the
answer-commitment term described in Appendix~\ref{app:protocol}. The
selected contrasts therefore illustrate decision-level differences and do
not by themselves establish an aggregate accuracy gain or isolate the dense
channel from the full training recipe.

For ALFWorld, we pair by environment index and decoding repeat and verify
identical game files. Each 1.5B arm contains 256 trajectories covering 92 distinct games;
some environment indices and decoding draws repeat a game. A wrong-object pickup requires an executed \texttt{take}
of a non-target type and an observation confirming the pickup.
Among the 209 DARS--GiGPO pairs that both succeed (81 distinct games),
the counts are 1 versus 10; the events occur in 1 versus 5 distinct
games. Against GRPO, the shared-success counts are 1 versus 12 over
156 pairs; against RLOO they are 1 versus 0 over 198 pairs. The favorable
contrast is thus specific to GiGPO and GRPO.

For second-object acquisition, we require a new target ID after a
successful first placement at the requested receptacle class. Among
23 DARS--GiGPO pairs from nine games where both policies observe at least
two target IDs immediately before their first target pickup and then
place the first, the counts are 22 versus 20. Across all 36 two-object
episodes, they are 33 versus 31. These are descriptive archive counts,
not independent trials or estimates of a causal effect.

\end{document}